\documentclass[journal]{IEEEtran}

\usepackage[normalem]{ulem}
\usepackage{amsmath,graphicx}
\usepackage{colortbl}
\usepackage{enumerate}
\usepackage{color}
\usepackage{url}
\usepackage{amsfonts}
\usepackage{multirow}
\usepackage{siunitx}
\usepackage{adjustbox}
\usepackage{balance}
\usepackage{makecell}

\usepackage{placeins}
\usepackage{etoolbox}
\newrobustcmd{\B}{\bfseries}
\DeclareMathOperator{\sign}{sign}

\DeclareMathOperator*{\argmax}{arg\,max}

\usepackage{pifont}
\newcommand{\cmark}{\ding{51}}%
\usepackage{hyperref}

\usepackage{tikz}
\usepackage{pgfplots}
\usepackage{pgfplotstable, filecontents}
\usetikzlibrary{positioning,chains,fit,shapes,calc}
\usepgfplotslibrary{statistics}
\usetikzlibrary{pgfplots.groupplots}
\pgfplotsset{compat=newest}
\pgfplotsset{myerrydown/.append style={solid, mark=triangle*, mark size=4pt, error bars/.cd, x dir=plus, x explicit} }
\pgfplotsset{myerryup/.append style={solid, mark=triangle*, mark size=4pt, error bars/.cd, x dir=minus, x explicit} }

\definecolor{cR18}{RGB}{230,25,75}
\definecolor{cR50}{RGB}{60,180,75}
\definecolor{cA}{RGB}{0,130,200}
\definecolor{c3m}{RGB}{206,66,244}

\definecolor{c0}{RGB}{0,76,153}    %plot blue
\definecolor{L}{RGB}{204,0,0}    %L-FGSM
\definecolor{N}{RGB}{0,204,0}    %N-FGSM
\definecolor{R}{RGB}{96,96,96}    %R-FGSM
\definecolor{P}{RGB}{0,0,204}    %P-FGSM

\definecolor{c0}{RGB}{0,76,153}    %plot blue

\definecolor{c1}{RGB}{255,0,0}    %R-FGSM
\definecolor{c2}{RGB}{0,0,0}    %sigma=0.5
\definecolor{c3}{RGB}{96,96,96}   %sigma=0.9
\definecolor{c4}{RGB}{127,127,127}    %sigma=0.95
\definecolor{c5}{RGB}{160,160,160}   %sigma=0.95u\definecolor{c6}{RGB}{0,0,255}   %L-FGSM

\definecolor{c7}{RGB}{255,0,255}   %Uniform
\definecolor{c8}{RGB}{0,204,0}   %U-FGSM

\definecolor{americanrose}{rgb}{0.4, 0.6, 0.8}
\definecolor{rosso}{rgb}{0.64, 0.76, 0.68}
\definecolor{eggplant}{rgb}{0.38, 0.25, 0.32}
\definecolor{darkpowderblue}{rgb}{0.0, 0.2, 0.6}
\definecolor{blue(ryb)}{rgb}{0.01, 0.28, 1.0}
\definecolor{capri}{rgb}{0.0, 0.75, 1.0}
\definecolor{blue(ncs)}{rgb}{0.0, 0.53, 0.74}
\definecolor{blue-violet}{rgb}{0.54, 0.17, 0.89}
\definecolor{ferrarired}{rgb}{1.0, 0.11, 0.0}
\definecolor{falured}{rgb}{0.5, 0.09, 0.09}
\definecolor{debianred}{rgb}{0.84, 0.04, 0.33}
\definecolor{cadmiumgreen}{rgb}{0.0, 0.42, 0.24}

\usepackage{booktabs}
\newcolumntype{N}{>{\centering\arraybackslash}m{.5in}}
\newcolumntype{G}{>{\centering\arraybackslash}m{2in}}

\usepackage{xfrac}

\usepackage{tkz-kiviat} 
\usetikzlibrary{arrows}

\begin{document}

\title{Exploiting vulnerabilities of deep neural networks for privacy protection}


\author{Ricardo Sanchez-Matilla, Chau Yi Li, Ali Shahin Shamsabadi, Riccardo Mazzon, Andrea Cavallaro
\thanks{%Manuscript received May 5, 2019; revised December 14, 2019 and March 24, 2020; accepted March 26, 2020.
{Chau Yi Li and Ali Shahin Shamsabadi equally contributed.}
%The guest editor coordinating the review of this manuscript and approving it for publication was Dr. Jingdong Wang. 
The authors are with the Centre for Intelligent Sensing (CIS), Queen Mary University of London (QMUL), London E1 4NS, U.K.}
%\thanks{Digital Object Identifier 10.1109/TMM.2020.2987694}
}

%\markboth{IEEE TRANSACTIONS ON MULTIMEDIA, 2020}%
%{Shell \MakeLowercase{\textit{et al.}}: Bare Demo of IEEEtran.cls for Journals}

\maketitle

\begin{abstract}
Adversarial perturbations can be added to images to protect their content from unwanted inferences. These perturbations may, however, be ineffective against classifiers that were not {seen}  during the generation of the perturbation, or against defenses {based  on re-quantization, median filtering or JPEG compression. 
To address these limitations, we present an adversarial attack {that is} specifically designed to protect visual content against { unseen} classifiers and known defenses. We craft perturbations using an iterative process that is based on the Fast Gradient Signed Method and {that} randomly selects a classifier and a defense, at each iteration}. This randomization prevents an undesirable overfitting to a specific classifier or defense. We validate the proposed attack in both targeted and untargeted settings on the private classes of the Places365-Standard dataset. Using ResNet18, ResNet50, AlexNet and DenseNet161 {as classifiers}, the performance of the proposed attack exceeds that of eleven state-of-the-art attacks. The implementation is available at \href{https://github.com/smartcameras/RP-FGSM/}{https://github.com/smartcameras/RP-FGSM/}.
\end{abstract}


\IEEEpeerreviewmaketitle

%%%%%%%%%%%%%%%%%%%%%%%%%%%%%
\section{Introduction}
\label{sec:intro}

Images shared online capture people and scenes that reveal personal information, {as well as} information about personal choices and preferences. This information can be automatically inferred by classifiers. To prevent this potential privacy violation and to protect the visual content from unwanted automatic inferences~\cite{Li2019}, we aim to exploit the vulnerability of classifiers to adversarial attacks~\cite{goodfellow2014,papernot2016limitations, carlini2017towards}. 

An adversarial attack should mislead classifiers {that} the attacker has access to (\emph{seen} classifiers), as well as classifiers {that} the attacker has no information about, not even the prediction output (\emph{unseen} classifiers). However, adversarial attacks often fail to mislead unseen classifiers as  {the generated} adversarial perturbations overfit {to} a specific classifier~\cite{Athalye2017} or {to} an ensemble of classifiers~\cite{tramer2017ensemble}. Adversarial {images} should also be robust to {\em defenses}. Defenses can be  based on re-quantization~\cite{Xu2018FeatureSqueezing}, median filtering~\cite{Xu2018FeatureSqueezing} and JPEG compression~\cite{liu2018feature}. Moreover, adversarial perturbations should not degrade the image quality{,} {especially when added for protecting privacy}~\cite{Li2019,Mediaeval2018}.

In this paper, we propose an adversarial attack that aims to prevent both seen and unseen classifiers from inferring private information present in {an} image, even when {the classifiers} are equipped with {known} defenses. At the core of the proposed attack there is an iterative combination of random selections of a classifier and a defense within a Fast Gradient Signed Method (FGSM) framework\footnote{FGSM is the iterative attack proposed by Kurakin {\em et al.}~\cite{kurakin2016adversarialscale}. Note that this attack is also known as Basic Iterative Method (BIM)~\cite{kurakin2016adversarialscale} and as Projected Gradient Descent (PGD) attack with $L_\infty$ norm~\cite{madry2017}.}~\cite{kurakin2016adversarial}. This random selection avoids the creation of perturbations that overfit to a specific classifier or defense, and improves the misleading rate for seen and unseen classifiers. The proposed attack, which can work in both targeted and untargeted settings, is related to methods that are based on ensemble of classifiers~\cite{tramer2017ensemble} and on defense transformations~\cite{Athalye2017,xie2018improving,dong2019evading}, but differs in the fact that {both} classifiers {and transformations} are randomly chosen at each iteration. Moreover, the proposed attack enables the use of complex defense transformations with null derivative.
We validate the proposed attack for the protection of scene content on the privacy subset of the Places365-Standard dataset~\cite{Mediaeval2018}, {which consists} of scenes such as places of {worship} and hospitals. We evaluate the attack on state-of-the-art classifiers, namely ResNet18, ResNet50, AlexNet, and DenseNet161, based on {the} misleading rate, detectability and image quality.

%%%%%%%%%%%%%%%%%%%%%%%%%%%%%%
\section{Problem definition}
\label{sec:problem_formulation}

Let~$\mathbf{x} \in \mathbb{R}^{W \times H \times {C}}$ be an image of width~$W$ and height~$H$ pixels, and {$C$} color channels whose dynamic ranges {are} $[0,255]$. Let~$M(\cdot)$ be a~$D$-class {deep neural network} classifier with parameters~$\mathbf{\theta}$ and trained using the cost function~$J_M(\cdot)$.

Let $\hat{y}_{\mathbf{x}}$ be the {\em true} class associated with~$\mathbf{x}$. The classifier outputs a prediction vector for $\mathbf{x}$,~$ \mathbf{p}_{\mathbf{x}} = (p_i)_{i=1}^D = M(\mathbf{x})$, where $p_i$ represents the probability of~$\mathbf{x}$ being associated with class $i \in \{1, ..., D\}$.
The {\em predicted} class for~$\mathbf{x}$,~$y_{\mathbf{x}}$, is the most likely {of}~$D$ classes:
\begin{equation}
\label{eq:predicted_class}
y_{\mathbf{x}} = \argmax_{i=1, ..., D}~p_i.
\end{equation}
Note that the predicted class might differ from the true class, which is unknown during the execution of the adversarial attack.
Adversarial attacks for privacy protection should aim to hide the {true} class, $\hat{y}_{\mathbf{x}}$, from~$M(\cdot)$, even when the predicted class is incorrect. The adversarial perturbation, $\boldsymbol{\delta}_\mathbf{x} \in \mathbb{R}^{W \times H \times C}$, added to the original image, $\mathbf{x}$, generates an adversarial image{,} as~{$\dot{\mathbf{x}} = \mathbf{x} + \boldsymbol{\delta}_\mathbf{x}$}. This perturbation causes the classifier to predict an {\em adversarial} class, $y_{\dot{\mathbf{x}}}$, by decreasing the probability of the predicted class, $y_{\mathbf{x}}$, (untargeted attack~\cite{kurakin2016adversarial,xie2018improving,MoosaviDezfooli16,modas2018sparsefool}) until
\begin{equation}
\label{eq:untargeted}
    y_{\dot{\mathbf{x}}} \neq y_{\mathbf{x}},
\end{equation}
or by increasing the probability of a specific target class, $y_{t}${,} (targeted attack~\cite{Li2019,papernot2016limitations,carlini2017towards,tramer2017ensemble,kurakin2016adversarialscale,kurakin2016adversarial,xie2018improving}), such that
\begin{equation}\label{eq:targeted}
y_{\dot{\mathbf{x}}} = y_{t} \neq y_{\mathbf{x}}.
\end{equation}
The target class can be determined randomly~\cite{Li2019, kurakin2016adversarial}{,} systematically as the least-likely class~\cite{kurakin2016adversarial}, or adaptively from the prediction vector~\cite{Li2019}. 

Defenses against adversarial attacks aim to eliminate, prior to inputting images to the classifier, {the effect of possible} adversarial perturbations using {a transformation, $\phi(\cdot)$, namely} median filtering~\cite{Xu2018FeatureSqueezing}, re-quantization~\cite{Xu2018FeatureSqueezing} or JPEG compression~\cite{liu2018feature}. Moreover, to {detect} an image as adversarial, the probability vector from a classifier for an image $\dot{\mathbf{x}}$ and its transformed version, $\phi(\dot{\mathbf{x}})$, can be compared with the $L_1$ {norm as} 
\begin{equation}
    \label{eq:detection}
    \| M\left(\dot{\mathbf{x}}\right) - M\left(\phi\left(\dot{\mathbf{x}}\right)\right) \|_1   > \tau,
\end{equation}
where $\tau$ is learned to accept a specific false-positive rate~\cite{Xu2018FeatureSqueezing}.

%%%%%%%%%%%%%%%%%%%%%%%%%%%%%%%%
\section{Background}\label{sec:background}

%=============================================
\begin{table}[t!]
\centering
\setlength\tabcolsep{2pt}
\caption{Comparison of adversarial attacks: choice of {Norm} and corresponding density of the {Perturbation}, which can be {sparse} (S) or {dense}~(D). The class selection ({Type}) can be untargeted (U) or targeted (T). An attack can be designed for multiple classifiers (MC) and to withstand image transformations (Trans.).}
\begin{tabular}{llccccc}
\Xhline{3\arrayrulewidth}
\multicolumn{1}{l}{Reference}  &    
\multicolumn{1}{l}{Attack}  &    
\multicolumn{1}{c}{Norm}    & 
\multicolumn{1}{c}{Perturbation}   &
\multicolumn{1}{c}{Type} &
\multicolumn{1}{c}{MC} &
\multicolumn{1}{c}{Trans.}\\ \hline
\cite{papernot2016limitations} & \multicolumn{1}{l}{JSMA} &$L_0$&S&T&&
\\
\cite{carlini2017towards} & \multicolumn{1}{l}{CW} &$L_0$,~$L_2$,~$L_\infty$&S, D &T&&
\\ %\hline
\cite{MoosaviDezfooli16} & \multicolumn{1}{l}{DeepFool} &$L_2$&D&U&&
\\
\cite{modas2018sparsefool} & \multicolumn{1}{l}{SparseFool} &$L_1$&S&U&&
\\
\cite{kurakin2016adversarial} & \multicolumn{1}{l}{U-FGSM} &$L_\infty$&D&U&&
\\
\cite{kurakin2016adversarialscale} & \multicolumn{1}{l}{R-FGSM}&$L_\infty$&D&T&&
\\
\cite{kurakin2016adversarial} & \multicolumn{1}{l}{L-FGSM}&$L_\infty$&D&T&&
\\
\cite{Li2019} & \multicolumn{1}{l}{P-FGSM}&$L_\infty$&D&T&&
\\
\cite{Athalye2017} & \multicolumn{1}{l}{EOT} &$L_\infty$&D&T&&\cmark
\\
\cite{xie2018improving} & \multicolumn{1}{l}{DI-FGSM}&$L_\infty$&D&U, T&\cmark&\cmark
\\
\cite{tramer2017ensemble} & \multicolumn{1}{l}{E-FGSM}&$L_\infty$&D&T&\cmark&
\\
Proposed & \multicolumn{1}{l}{RP-FGSM}&$L_\infty$&D&U, T&\cmark&\cmark
\\
\Xhline{3\arrayrulewidth}
\end{tabular}
%}
\label{tab:RelatedWork}
\end{table}
%=============================================

\begin{figure}[t!]
    \centering
    \setlength\tabcolsep{1pt}
   \begin{tabular}{ccccc}
   %\begin{tabular}{p{0.2\columnwidth}p{0.2\columnwidth}p{0.2\columnwidth}p{0.2\columnwidth}p{0.2\columnwidth}}
    \scriptsize Original & \scriptsize JSMA & \scriptsize CW & \scriptsize \hspace*{-5.8pt}{DeepFool} & \scriptsize \hspace*{-5.8pt}{SparseFool} \\
    \includegraphics[width=0.19\columnwidth]{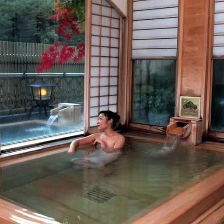} &
    \includegraphics[width=0.19\columnwidth]{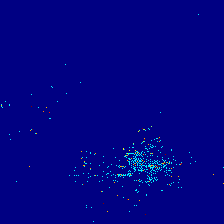} & \includegraphics[width=0.19\columnwidth]{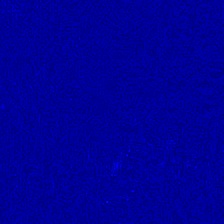} &
    \hspace*{-5.8pt}{\includegraphics[width=0.19\columnwidth]{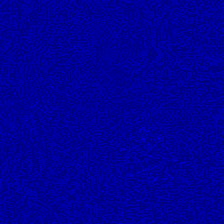}} &
    \hspace*{-5.8pt}{\includegraphics[width=0.19\columnwidth]{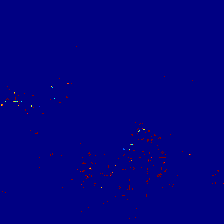}}\\
    & \scriptsize EOT & \scriptsize DI-FGSM & \scriptsize \hspace*{-5.8pt}{E-FGSM} & \scriptsize \hspace*{-5.8pt}{RP-FGSM} \\
    & \includegraphics[width=0.19\columnwidth]{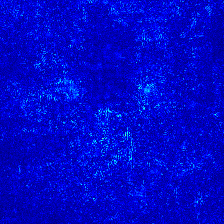} &
    \includegraphics[width=0.19\columnwidth]{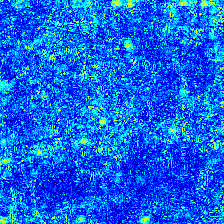} &
    \hspace*{-5.8pt}{\includegraphics[width=0.19\columnwidth]{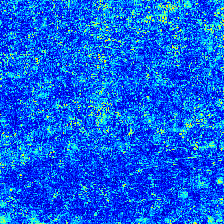}} &
    \hspace*{-5.8pt}{\includegraphics[width=0.19\columnwidth]{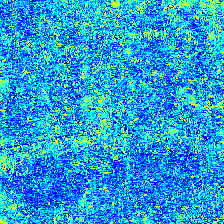}} \\
    & \multicolumn{3}{l}{
        \hspace*{-8.5pt}
        {
        \begin{tikzpicture}
        \pgfplotscolorbardrawstandalone[ 
            colormap/jet,
            colorbar horizontal,
            point meta min=0,
            point meta max=100,
            colorbar style={
                height=0.3cm,
                width=0.58\columnwidth,
                xtick style={draw=none},
                label style={font=\small},
                tick label style={font=\scriptsize},
                xtick={0,30,60,95},
                xticklabels={0,20,40,{72+}}, 
                tick label style={font=\small},
                x tick label style={xshift={(\ticknum==0)*0.25em},xshift={-(\ticknum==4)*2.5em}}
        }]
        \end{tikzpicture}
    }
    }
    \end{tabular}
    \caption{Cumulative adversarial perturbations (sum of the absolute difference between original and adversarial images across the three color channels) generated by 
    JSMA, CW, DeepFool, SparseFool, EOT, DI-FGSM (untargeted), E-FGSM, and {the proposed} RP-FGSM (untargeted).
    }
    \label{fig:noises}
\end{figure}

Adversarial images are crafted by constraining the added perturbation with, typically, an $L_p$ norm between $\mathbf{x}$ and $\dot{\mathbf{x}}$. Attacks may {constrain} the total number of perturbed pixels ($L_0$)~\cite{papernot2016limitations}, the sum of magnitudes ($L_1$)~\cite{modas2018sparsefool}, the Euclidean distance ($L_2$)~\cite{carlini2017towards,MoosaviDezfooli16}, or the maximum per-pixel variation ($L_\infty$)~\cite{Li2019,tramer2017ensemble,kurakin2016adversarial}. As the optimization of the perturbation, regularized by the norm,  has no closed-form solution due to non-linear operations with the parameters{,} $\theta${,} and non-convex cost functions {used} to train the {deep neural network} classifiers, several adversarial attacks iteratively generate the adversarial image,  $\dot{\mathbf{x}}$, as~\cite{kurakin2016adversarial}:
\begin{equation}
    \dot{\mathbf{x}}_{n+1} = \dot{\mathbf{x}}_{n} + \boldsymbol{\delta}_{\dot{\mathbf{x}}_{n}},
\end{equation}
from $\dot{\mathbf{x}}_0 = \mathbf{x}$. The iteration process stops when a specific number of iterations, $N$, is reached, i.e. $\dot{\mathbf{x}} = \dot{\mathbf{x}}_{N+1}$, or when the misleading objective is achieved \cite{papernot2016limitations,MoosaviDezfooli16}. $N$ is typically chosen as a function of parameters linked to the preservation of image quality~\cite{papernot2016limitations, kurakin2016adversarialscale}.
Table~\ref{tab:RelatedWork} provides a comparative summary of adversarial attacks. Moreover, Figure~\ref{fig:noises} shows examples of the magnitude of perturbations generated by representative adversarial attacks, which are described next.

%=============================================

JSMA~\cite{papernot2016limitations}, a targeted attack, uses the~$L_0$ norm and aims to identify and perturb by one {intensity} unit, at each iteration, the {\em two} most effective pixels. JSMA increases the probability of the target class, $y_{t}$, compared with that of any other classes, using a saliency score, $S(\cdot)$, {which} determines the effect of perturbing {a} pixel{,} $x_i \in \mathbf{x}$:
\begin{align}
    S(x_i,y_{t}) &= \\ \nonumber
    &\begin{cases}
    0 \quad \text{if~${\nabla_{x_i} M_t(\mathbf{x})}<0$ or~$\sum_{j \neq t}\nabla_{x_i} M_j(\mathbf{x})>0$}\\ 
    \nabla_{x_i} M_t(\mathbf{x})\Big\|\sum_{j \neq t}\nabla_{x_i} M_j(\mathbf{x}) \Big\|_1 \quad \text{otherwise},
  \end{cases}
\end{align}
where~$\nabla_{x_i} M_t(\mathbf{x})$ is the gradient of the target class, $y_{t}$, with respect to pixel~$x_i$. The attack iteratively generates the perturbation, until the adversarial image is classified as the target class or a specified number of pixels have been perturbed.

Carlini-Wagner (CW)~\cite{carlini2017towards}, a targeted attack, maximizes the difference between the logarithmic probabilities of the target class and other classes using the~$L_2$ norm.
Three forms of selecting the target classes were used, namely uniform random selection, second- and least-likely cases.

DeepFool~\cite{MoosaviDezfooli16}, an untargeted attack, crafts adversarial perturbations controlled by the $L_2$ norm. At each iteration, the adversarial perturbation is the orthogonal projection of the adversarial image {from} the previous iteration onto the closest linearized class boundary of $M(\cdot)$. The final adversarial image,~$\dot{\mathbf{x}}=\dot{\mathbf{x}}_{N+1}$, is the one that exceeds the closest class boundary as 
\begin{equation}
    \dot{\mathbf{x}}_{N+1} = \dot{\mathbf{x}}_{N} + (1+\eta)\boldsymbol{\delta}_{\dot{\mathbf{x}}_{N}},
\end{equation}
where~$\eta \ll 1$ and~$\boldsymbol{\delta}_{\dot{\mathbf{x}}_{N}}$ is the final adversarial perturbation. As DeepFool might generate adversarial images whose pixels exceed the allowed dynamic range, 
SparseFool~\cite{modas2018sparsefool}, an untargeted attack similar to DeepFool, clips the pixel values within $[\mathbf{l},\mathbf{u}]$, as 
\begin{equation}\label{eq:sparsefool}
    \mbox{min} \; \|\dot{\mathbf{x}}-\mathbf{x}\|_1 \quad\mbox{s.t.}\quad  y_{\dot{\mathbf{x}}} \neq y_{\mathbf{x}} \quad \text{and} \quad \mathbf{l} \leq \dot{\mathbf{x}} \leq \mathbf{u}.
\end{equation}
This minimization is solved based on the low-mean curvature properties of the decision boundary of each image~\cite{fawzi2018empirical}. Note that, unlike DeepFool, SparseFool uses the $L_1$ norm{, thus generating sparse perturbations}.

The family of attacks based on FGSM generates the perturbation in the direction of the cost function,~$J_M(\cdot)$, in order to maximize the cost of remaining in the predicted class $y_\mathbf{x}$ (untargeted) or to minimize the cost of predicting the targeted class $y_t$ (targeted):
\begin{equation}
\label{eq:optimization_adversarial}
    \boldsymbol{\delta}_\mathbf{x} = 
\begin{cases}
 \ \ \delta \sign\left(\nabla_{\mathbf{x}}J_M\left(\mathbf{\theta},\mathbf{x},y_{{\mathbf{x}}}\right)\right) &\mbox{untargeted attack},\\
- \delta \sign\left( 
\nabla_{\mathbf{x}}J_M
\left ( \mathbf{\theta},\mathbf{x},y_{t} \right)\right) & \mbox{targeted attack},
\end{cases}
\end{equation}
where~$\delta$ controls the magnitude of the perturbation,~$\sign(\cdot)$ is the sign function and~$\nabla_{\mathbf{x}}J_M(\cdot)$ is the gradient of {the cost function}, $J_M(\cdot)$, showing the updated direction with respect to~$\mathbf{x}$. FGSM-based attacks use {the}~$L_\infty$ {norm} and limit $\dot{\mathbf{x}}$ to the~{$\epsilon$-neighborhood} of~$\mathbf{x}$ and within the dynamic range of an image at each iteration{,} as
\begin{equation}
\dot{\mathbf{x}}_{n+1} = \mathcal{C}_{\mathbf{x},\epsilon} \left(\dot{\mathbf{x}}_{n} + \boldsymbol{\delta}_{\dot{\mathbf{x}}_{n}}\right),
\end{equation}
where~$\mathcal{C}_{\mathbf{x},\epsilon}(\cdot)$ is a clipping function, defined as
\begin{equation}
\label{eq:clipping}
    \mathcal{C}_{\mathbf{x},\epsilon} (\dot{\mathbf{x}}) = \text{min} \big\{255, \mathbf{x} + E, \text{max}\left\{ 0, \mathbf{x}-E,\dot{\mathbf{x}}\right\} \big\},
\end{equation}
where $E = \left \{ \epsilon \right \}^{W \times H \times C}$.
The value of~$\epsilon$ is a trade-off between {the} misleading rate and the quality of the adversarial image: the larger~$\epsilon$, the higher the potential misleading rate but, also, the stronger the image degradation.

To evade defenses based on transformations and to preserve visual quality, Expectation Over Transformation (EOT)~\cite{Athalye2017} optimizes the loss on the target class over a set of {pre-defined} 2D transformations, $\mathbf{\Phi}_{2D}$, while minimizing the distance between the transformed original image and the transformed adversarial image in the $Lab$ color space~\cite{ruderman1998statistics}: 
\begin{equation}
\label{eq:eot}
%\small
\begin{split}
    \dot{\mathbf{x}}_{n+1} = \, &\mathcal{C}_{\mathbf{x},\epsilon}(\dot{\mathbf{x}}_{n} - \delta \sign( \nabla_{  {\mathbf{x}}}J_{ M} (\mathbb{\theta} ,\phi_{2D} (\dot{\mathbf{x}}_{n}),y_{t}) \, + \\ 
    &  + \, \lambda \| {\mathcal{L}}(\phi_{2D} ({\dot{\mathbf{x}}_{n}}))  -  {\mathcal{L}}(\phi_{2D}(\dot{\mathbf{x}}_{n}))\|_{2} ),
    \end{split}
\end{equation}
where $\mathcal{L}(\cdot)$ {is a function that} converts images from the $RGB$ to the $Lab$ color space, $\lambda$ controls the visual similarity, and $\phi_{2D}(\cdot) \in  \mathbf{\Phi}_{2D}$ {is an image transformation chosen with uniform random probability at each iteration. }

{To increase the misleading rate with unseen classifiers}, similarly to EOT, Diverse Input Fast Gradient Sign Method (\mbox{DI-FGSM})~\cite{xie2018improving} applies a random resizing followed by a padding transformation{, with a pre-defined probability, on the adversarial image at each iteration.}

To improve {misleading} performance with unseen classifiers, Ensemble FGSM (E-FGSM)~\cite{tramer2017ensemble}  employs multiple classifiers, $M_k (\cdot)$, simultaneously when creating the perturbation:
\begin{equation}
\label{eq:tfgsm_iter}
\dot{\mathbf{x}}_{n+1} = \mathcal{C}_{\mathbf{x},\epsilon}\left(\dot{\mathbf{x}}_{n} -\delta \sign\left(\sum_{k=1}^{K}\nabla_{\mathbf{x}}J_{M_k}\left(\mathbb{\theta},\dot{\mathbf{x}}_{n},y_{t}\right)\right)\right), 
\end{equation}
where $K \ge 1$.
However, the {use of all available classifiers at each iteration results in an} overfitting {that} limits the ability of the adversarial image to mislead unseen classifiers~\cite{xie2018improving}.

{Regarding the approach for selecting the target class, } Least-likely FGSM (L-FGSM)~\cite{kurakin2016adversarial} selects the least-likely predicted class. However, this systematic target class selection can compromise the protection of the image, as the selection process can be reversed~\cite{Li2019}. Random-FGSM (R-FGSM)~\cite{kurakin2016adversarialscale}, E-FGSM~\cite{tramer2017ensemble} and EOT~\cite{Athalye2017} randomly select the target class {from} all possible classes except the predicted class and the risk of reversibility is negligible. However, as the goal of the adversarial attacks is to hide the true class, which may not be the same as the predicted class (see Table~\ref{tab:accuracy}), \emph{this strategy can result in the selection of the true class as the target}. Private-FGSM (P-FGSM)~\cite{Li2019} avoids targeting the true class by discarding from the selection process the top predicted classes, which are more likely to contain the true class. The selection from the remaining classes is still random, thus limiting the risk of reversibility. Instead, untargeted approaches~\cite{kurakin2016adversarial,xie2018improving} do not need the selection of a target class. However, knowledge of the true class is required for privacy protection.
%
%===============================
\begin{table}[t!]
    \centering
    \caption{Classification accuracy in the test set of the Private Places365 dataset for ResNet18, ResNet50, AlexNet and DenseNet161. KEY - T1: top-1 classification accuracy; T5: top-5 classification accuracy.}
    \label{tab:accuracy}
    \begin{tabular}{cccccccc}
        \Xhline{3\arrayrulewidth}
        \multicolumn{2}{c}{ResNet18} & \multicolumn{2}{c}{ResNet50} & \multicolumn{2}{c}{AlexNet} & \multicolumn{2}{c}{DenseNet161} \\ %\hline
        \multicolumn{1}{c}{T1} & \multicolumn{1}{c}{T5} & \multicolumn{1}{c}{T1} & \multicolumn{1}{c}{T5} & \multicolumn{1}{c}{T1} & \multicolumn{1}{c}{T5} & \multicolumn{1}{c}{T1} & T5\\ \hline
        \multicolumn{1}{c}{54.6} & \multicolumn{1}{c}{84.4} & \multicolumn{1}{c}{56.4} & \multicolumn{1}{c}{86.5} & \multicolumn{1}{c}{47.7} & \multicolumn{1}{c}{79.0} & \multicolumn{1}{c}{{58.4}} & {86.6}\\
    \Xhline{3\arrayrulewidth}
    \end{tabular}
\end{table}
%===============================

%%%%%%%%%%%%%%%%%%%%%%%%%%%%%%%%%%%%%
\section{Robust and Private FGSM}
\label{sec:proposed_method}

We aim to generate adversarial perturbations that protect the {\em true} class of images and thus protect {the} private information they contain from unwanted automatic inferences. These adversarial perturbations should {be robust to} defenses, be undetectable, preserve image quality, and mislead both seen and unseen classifiers.

\begin{figure*}[!t]
    \centering
    \includegraphics[width=0.95\textwidth]{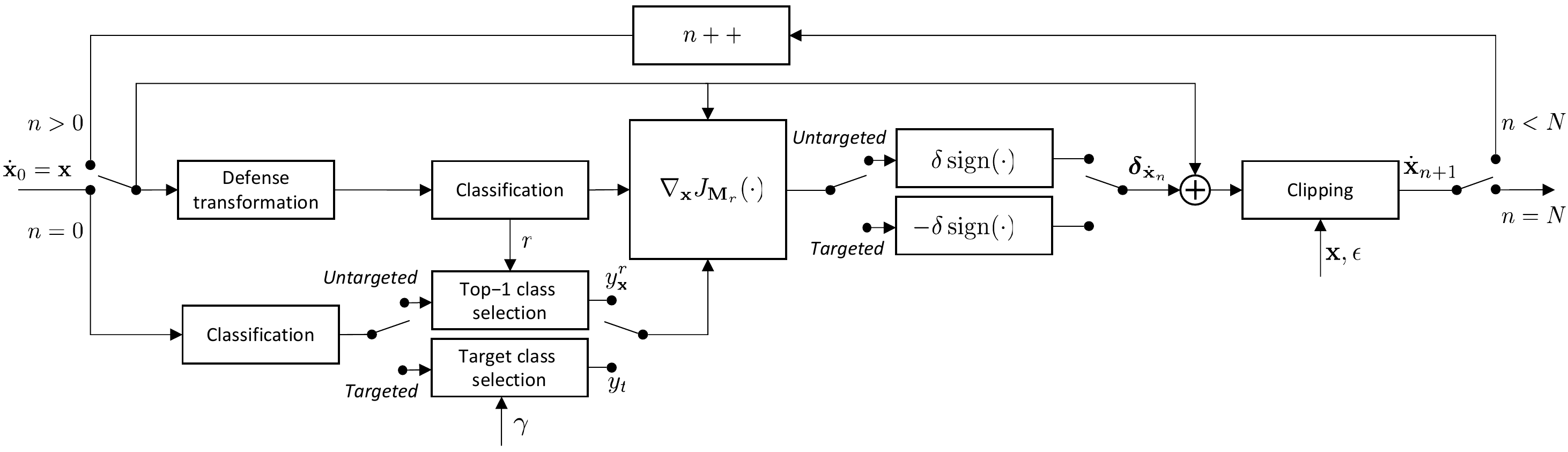}
    \caption{Block diagram for the proposed adversarial attack, RP-FGSM. KEY -- 
    $\mathbf{x}$: original image;
    $r$: random index for selection of transformation and classifier;
    $\phi_r(\cdot)$: {randomly} chosen defense transformation;
    $M_r(\cdot)$: {randomly} chosen classifier;
    $\nabla_{\mathbf{x}}J_{M_{r}}$: gradient of the cost function $J(\cdot)$ for the classifier $M_r(\cdot)$ when predicting class~$y_{t}$ (targeted) or~$y_{\mathbf{x}}^r$ (untargeted) with respect to $\mathbf{x}$;
    $\delta$: magnitude of the perturbation added at each iteration;
    $\epsilon$: parameter that controls the clipping function to maintain intensity values of each pixel within the $\epsilon$-neighborhood of the original intensity value;
    $\gamma$: parameter for target class selection;
    $N$: number of iterations;
    $\mathbf{\dot{x}}_{n+1}$: adversarial image at iteration $n+1$.
    Note that when multiple classifiers are employed, the target class selection uses the intersection of the prediction vectors, as indicated in Eq.~\ref{eq:mu_selection}.}
    \label{fig:diagram}
\end{figure*}

We propose an iterative approach, \emph{Robust Private FGSM} (RP-FGSM), that avoids overfitting by randomly selecting, at each iteration, a classifier to attack and a defense to evade. The proposed approach differs from E-FGSM~\cite{tramer2017ensemble} and DI-FGSM~\cite{xie2018improving}, which consider an ensemble of classifiers, and from EOT~\cite{Athalye2017} and DI-FGSM~\cite{xie2018improving}, which use only non-defense transformations (e.g.~rotations).  Moreover, as {in} other FGSM-based attacks~\cite{kurakin2016adversarial}, RP-FGSM maintains image quality inherently by controlling the magnitude of the perturbation with the parameter~$\epsilon$. The block diagram of RP-FGSM is shown in Figure~\ref{fig:diagram}.
%===============================
%===============================

Let~$\mathbf{M}=\{ M_k(\cdot) \}_{k=1}^{K}$ be a set of~$K \ge 1$ classifiers, and~{$\mathbf{\Phi} = \{ \phi_f(\cdot) \}_{f=0}^{F}$} be a set of~$F$ defense {\em transformations}, where $\phi_0(\cdot)$ is the identity function (i.e.~no transformation is applied to the input). For a classifier~$M_k(\cdot)$ and transformation~$\phi_{f}(\cdot)$, let $y^k_{\dot{\mathbf{x}}}$ and $y^k_{\phi_{f}(\dot{\mathbf{x}})}$ be the predicted classes (Eq.~\ref{eq:predicted_class}) of the adversarial image,~$\dot{\mathbf{x}}$, and the transformed adversarial image,~${\phi_{f}(\dot{\mathbf{x}})}$, respectively.
We generate the adversarial image,~$\dot{\mathbf{x}}$, for $\mathbf{x}$, whose true class is $\hat{y}_{\mathbf{x}}$, as
\begin{equation}
    \hat{y}_{\mathbf{x}} \neq y^k_{\dot{\mathbf{x}}} \quad \text{and} \quad 
    \hat{y}_{\mathbf{x}} \neq y^k_{\phi_{f}(\dot{\mathbf{x}})} \quad \forall f, k.
\end{equation}
The process is initialized with~{$\dot{\mathbf{x}}_0 = \mathbf{x}$} and ends with {$\dot{\mathbf{x}} = \dot{\mathbf{x}}_{N+1}$}. At each iteration, $n$, we randomly select a transformation $R(\mathbf{\Phi}) \in \mathbf{\Phi}$ and a classifier $R(\mathbf{M}) \in \mathbf{M}$, 
where $R(\cdot)$ is a function that randomly selects an element from a set.

We use the most effective defense transformations,  namely re-quantization~\cite{Xu2018FeatureSqueezing}, median filtering~\cite{Xu2018FeatureSqueezing} and JPEG compression~\cite{liu2018feature}. 
Both re-quantization and JPEG compression have a quantization and rounding step, where the derivative, needed to compute the perturbation, is zero for almost every pixel, thus resulting in a null adversarial perturbation. We prevent the generation of a null perturbation when applying these transformations by approximating each pixel of the image $\dot{\mathbf{x}}_n$ as $\lfloor \dot{\mathbf{x}}_{n}^i\rceil+(\dot{\mathbf{x}}_{n}^i-\lfloor \dot{\mathbf{x}}_n^i \rceil)^3$, where~{$\lfloor\cdot\rceil$} represents the rounding operator to the nearest integer, and $i$ is the pixel index, whose derivative is unlikely to be zero~\cite{Shin2017}.  
To avoid overfitting to certain transformations and parameters, the parameters of each transformation (i.e.~the number of bits for re-quantization, the kernel size for median filter and the compression parameters for JPEG compression) are also chosen randomly from a pre-defined set of values~\cite{xie2018improving} (more details in Sec.~\ref{sec:implementation_details}). 

RP-FGSM can craft an adversarial image using targeted or untargeted attack by adding, at each iteration, the perturbation 
\begin{equation}
\label{eq:rm-fgsm}
\begin{split}
 \!\!\boldsymbol{\delta}_{\dot{\mathbf{x}}_{n}} = 
\begin{cases}
    \begin{aligned}
    \!&\delta\,\sign(\nabla_{\!{\mathbf{x}}}J_{{M}_r}(\mathbb{\theta},\phi_r(\dot{\mathbf{x}}_{n}),y_\mathbf{x}^r))&\!\!\text{untargeted},\!\!\!\\[2pt]
    \!-&\delta\,\sign(\nabla_{\!{\mathbf{x}}}J_{{M}_r}(\mathbb{\theta},\phi_r(\dot{\mathbf{x}}_{n}),y_t))&\!\!\text{targeted},\!\!\!
    \end{aligned}
\end{cases}
\end{split}
\end{equation}
where $y_\mathbf{x}^r$ is the class predicted by the classifier $M_r(\cdot)$. Moreover, RP-FGSM maintains the perturbation within the $\epsilon$-neighborhood of $\dot{\mathbf{x}}$ as 
\begin{equation}
%\label{eq:fgsm_iter}
\dot{\mathbf{x}}_{n+1} = \mathcal{C}_{\mathbf{x},\epsilon} \left(\dot{\mathbf{x}}_{n} + \boldsymbol{\delta}_{\dot{\mathbf{x}}_{n}}\right),
\end{equation}
where $\mathcal{C}_{\mathbf{x},\epsilon}(\cdot)$ is the clipping function {(Eq.~\ref{eq:clipping})}. 

Untargeted RP-FGSM crafts the adversarial perturbation {in such a way} that the predicted adversarial class moves farther away, in the decision space, from the most-likely predicted class {for each classifier}, which we assume to be in the neighborhood of the true class.

\begin{figure}[!t]
\centering
\includegraphics[width=\columnwidth]{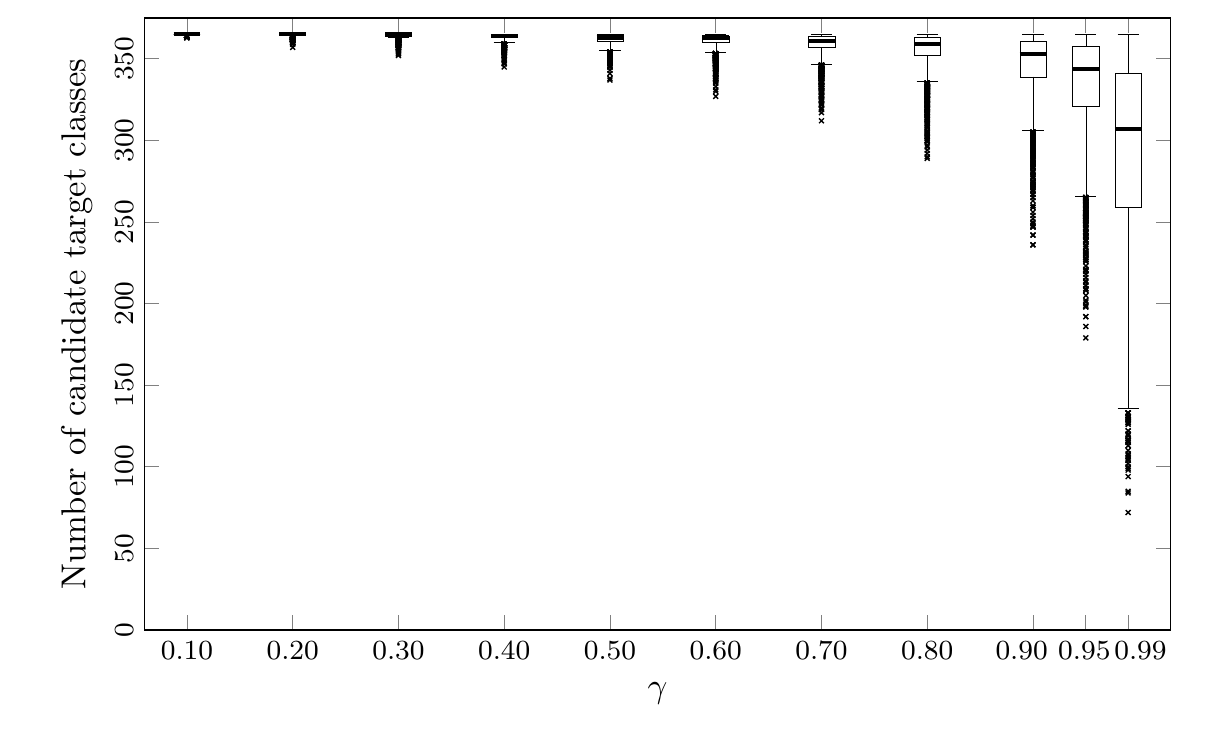}
\caption{Influence of $\gamma$ on the number of candidate target classes, obtained as intersection of the subsets of classes whose cumulative probability exceeds the threshold $\gamma$. The plot is generated for 3000 images from the scene privacy subset defined in~\cite{Mediaeval2018} using ResNet18, ResNet50 and AlexNet. 
The horizontal line within the box is the median; the lower and upper edges are the 25-percentile and 75-percentile, respectively. Each dot represents an outlier beyond 2.7 standard deviations from the average.
When $\gamma=0.1$, the median of the number of candidate classes is 364 (i.e. the target class is chosen from all but the predicted class for most images). 
As $\gamma$ increases, the number of target-class candidates decreases. When $\gamma = 0.99$, the median decreases to $307$ with a distribution that is more likely to exclude the true class from the target-class selection.
}
\label{fig::effect_gamma_intersection}
\end{figure}

Targeted RP-FGSM selects the target class from a set that is more likely to exclude the true class, by leveraging the fact that the true class is often among the top classes predicted by a classifier~\cite{Li2019}. Moreover, we reduce the risk that an attack is reversed by randomly selecting  the target class. Specifically, we exploit the prediction probability vectors~$\{\mathbf{p}^{k}_\mathbf{x}\}$ of the~$K$ classifiers.
Let~$\mathbf{p}'^{k}_\mathbf{x} = ({p_i'^{k}})_{i=1}^{D}$ contain the elements of~$\mathbf{p}^{k}_\mathbf{x}$ sorted in descending order. The target class is selected as 
\begin{equation}
    \label{eq:mu_selection}
    y_{t}= R\left(\bigcap^{K}_{k=1} \left\{ y_{j+1} : \sum_{i=1}^{j} {p'}_i^k > \gamma, j \in \{1, ..., D-1\} \right\}\right),
\end{equation}
where $R(\cdot)$ randomly selects an element from the set {of candidate target classes,} obtained as the intersection of the subset of classes whose cumulative probability exceeds a threshold~{$\gamma \in [0,1]$}. The larger~$\gamma$, the fewer the classes. Figure~\ref{fig::effect_gamma_intersection} shows the influence of~$\gamma$ on the number of classes from which the target class~$y_t$ is selected.

After $N$ iterations{,} the attack uses, on average, each classifier $\sfrac{N}{K}$ times and each transformation $\sfrac{N}{F}$ times. The number of iterations $N$ {might} affect the quality of the image and therefore we choose it as a function of $\epsilon$~\cite{kurakin2016adversarialscale} (see Sec.~\ref{sec:implementation_details}).

%================
%================

%%%%%%%%%%%%%%%%%%%%%%%%%%%%%%%%%
\section{Validation}
\label{sec:validation}

%%%%%%%%%%%%%%%%%%%%%%%%%%%%%%%%%%
\subsection{Experimental setup}
\label{sec:expSetup}

We compare the targeted and untargeted versions of the proposed attack, {Robust Private FGSM} (RP-FGSM), with eleven state-of-the-art methods, namely
Jacobian-based Saliency Map Approach (JSMA)~\cite{papernot2016limitations}, Carlini-Wagner (CW) with~$L_2$ norm~\cite{carlini2017towards}, DeepFool~\cite{MoosaviDezfooli16}, SparseFool{~\cite{modas2018sparsefool}},
iterative untargeted FGSM (U-FGSM)~\cite{kurakin2016adversarial}, Random FGSM (R-FGSM)~\cite{kurakin2016adversarialscale}, Least-likely FGSM (L-FGSM)~\cite{kurakin2016adversarial}, Private FGSM (P-FGSM)~\cite{Li2019}, Expectation Over Transformation (EOT)~\cite{Athalye2017}, Diverse Input FGSM (DI-FGSM, targeted and untargeted versions)~\cite{xie2018improving} and Ensemble FGSM (E-FGSM)~\cite{tramer2017ensemble}. 
{JSMA, CW, DeepFool and SparseFool} are run with the parameters proposed by their authors.  The target class for JSMA is the one that introduces the smallest perturbation, whereas for CW {it} is the second most-likely predicted class. 
{The target class is the least-likely predicted class for L-FGSM, chosen as described in Eq. 17 for P-FGSM and RP-FGSM, and, for the other targeted FGSM-based attacks, is selected randomly from the set of possible classes, excluding the predicted class of the original image.}
For untargeted FGSM-based attacks, we use the predicted class to craft the adversarial perturbation. 

The Places365-Standard dataset~\cite{zhou2017places}, which has over 1.8 million images of 365 scene classes, is used for training the classifiers (see Table~\ref{tab:accuracy}).
We use{,} as test set{,} the {\em scene privacy} subset of the validation set defined for the MediaEval 2018 Pixel Privacy Challenge~\cite{Mediaeval2018}, which comprises 60 private classes\footnote{The subset of private classes includes scenes that may require, for various reasons, privacy protection, such as
{\em army-base, bathroom, bedchamb, bedroom, church (indoor and outdoor), hospital and hospital-room, nursing-home, pharmacy, sauna, shower, swimming pool (indoor and outdoor), jacuzzi (indoor), temple (Asia)}; as well as scenes that would disclose private personal information, such as {\em airplane-cabin, airport terminal, amusement-park, aqueduct, bank-vault, bar, beach, beach-house, beer-garden, beer-hall, berth, bullring, bus-interior, bus station/indoor, campsite, car-interior, castle, catacomb, chalet, child-room, classroom, closet, coast, discotheque, dorm-room, drugstore, gymnasium (indoor), home-office, kindergarten-classroom, locker-room, mosque (outdoor), playground, playroom, pub (indoor), sandbox, schoolhouse, ski-resort, ski-slope, slum, swimming-hole, train-interior, train station/platform, tree-house, and waiting-room}. Note that we do not force the target class to be out of the 60 private classes, as this would restrict the diversity of target class and would disclose the private classes.}{, {that are a} subset of the classes of the Places365-Standard dataset~\cite{zhou2017places}}. 
{The test set is composed of 3,000 images, with 50 images from each of the (60) private classes.}
We consider four state-of-the-art classifiers, namely ResNet~\cite{He16} with 18 and 50 layers, AlexNet~\cite{krizhevsky2012imagenet} and DenseNet161~\cite{Huang2016densenet}, trained for scene classification~\cite{zhou2017places}\footnote{\url{https://github.com/CSAILVision/places365}}.
Table~\ref{tab:accuracy} reports their classification accuracy on the test set. Adversarial attacks are performed on ResNet18, ResNet50 and AlexNet, whereas DenseNet161, the most accurate classifier, is only used as an unseen classifier for testing. This represents the most challenging setup, given that the {largest and most} accurate classifier was never seen by the adversarial attacks~\cite{tramer2017ensemble}.

\subsection{Implementation details}
\label{sec:implementation_details}
We compare all FGSM-based attacks with the same parameters proposed in Kurakin {\em et al.}~\cite{kurakin2016adversarial}. 
The adversarial noise per iteration is~$\delta=1$, which corresponds to the smallest variation in an 8-bit image. {We} constrain the maximum {perturbation magnitude} {by setting}~$\epsilon=16${,} to provide a trade-off between {the} misleading rate and image quality degradation~\cite{xie2018improving}. 
{
For a fair comparison, we ensure that all {FGSM-based} attacks, either employing one classifier or an ensemble of classifiers (E-FGSM and DI-FGSM), perform the same number of forward/backward passes on the classifiers, with $\overline{N}=\text{min}(1.25 \, \epsilon, \epsilon+4)$~\cite{kurakin2016adversarial} iterations.
}
The number of iterations for {RP-FGSM} is~$N= \overline{N} \cdot K$, as only one classifier is used at each iteration {(ensembles use all classifiers at each iteration)}. This ensures that all attacks that employ $K$ classifiers perform $K \cdot${$\overline{N}$} forward/backward passes in total.

For EOT, we use as 2D transformations{, $\mathbf{\Phi}_{2D}$, scaling, translation, rotation, lightening, darkening and additive Gaussian noise.
The specific parameters of the transformations are chosen randomly from pre-defined intervals as follows: scaling with a factor between 0.8 and 1.2; translation between -0.2 and 0.2 times $W$; rotation between -60 and 60 degrees; lightening/darkening between 0 and 13 intensity points; and Gaussian noise with zero mean and 25 variance.} {We set $\lambda=0.5$ to achieve a misleading rate of approximately 90\%, as suggested in~\cite{Athalye2017}.}
For DI-FGSM, we employ random resizing followed by random padding with {a} probability {of} 50\% at each iteration. Original images {$\mathbf{x} \in\mathbb{R}^{224\times 224\times {C}}$} are resized and padded to $r \times r \times {C}$ with $r$ randomly chosen within the range $[200,224)$.

The parameters for the defense transformations are: for re-quantization, 1 to 7 bits per color channel in steps of 1~\cite{Xu2018FeatureSqueezing}; for median filtering, squared kernel of dimensions 2, 3 and 5~\cite{Xu2018FeatureSqueezing}; and for lossy JPEG compression, quality parameters 25, 50, 75 and 100~\cite{dziugaite2016study,das2017keeping}.
For RP-FGSM, we randomly and uniformly select the transformation to apply to the image at each iteration, by selecting the parameters (i.e. number of bits for re-quantization, kernel dimension for median filtering and compression factor for JPEG compression) from the same sets used for the defense.
As Li {\em et al.}~\cite{Li2019}, in targeted attacks, we use~{$\gamma=0.99$} for {target} class selection.

%%%%%%%%%%%%%%%%%%%%%%%%%%%%%%%%%%
\subsection{Performance measures}\label{sec:perfMeasures}

%Convolutional Neural Networks are state-of-the-art in a number of applications that are commonly used every day. However, there exists a lack of study in their interpretability, thus preventing to extend their applicability to tasks where human risk might be involved such as robotics [REF].
%In this paper, we have seen how Convolutional Neural Network are vulnerable to adversarial perturbations. 

We consider as performance measures the misleading rate, detectability and image quality.

{\em Misleading rate} is the ratio between the number of adversarial images that mislead the classifier with respect to the {\em true} class and the total number of images, expressed as a percentage. The higher the misleading rate in top ranks, the better the privacy protection. We consider as ranks top-1 and top-5.
The misleading rate is calculated for seen and unseen classifiers, and with and without defenses.

{\em Detectability} is the percentage of correctly detected adversarial images {with respect to the total number of images}. We distinguish between adversarial images and original images by learning a threshold, $\tau$, for each defense (re-quantization, median {and} JPEG compression) by accepting 5\% of the original (training) images as adversarial (i.e.~5\% false positive detection rate) {(Eq.~\ref{eq:detection})}~\cite{Xu2018FeatureSqueezing}.

{\em Image quality} is evaluated with Blind Referenceless Image Spatial Quality Evaluator (BRISQUE)~\cite{Mittal2012}, a no-reference measure; and Peak-Signal-to-Noise Ratio (PSNR) and Most Apparent Distortion (MAD)~\cite{MAD:ImageQuality:Larson}, two full-reference measures.
BRISQUE~\cite{Mittal2012} quantifies distortions and unnaturalness. The lower {the} BRISQUE {score}, the better the image quality. 
{PSNR quantifies the pixel-by-pixel difference between two images}. The higher the PSNR, the smaller the added perturbation (and therefore, the better the image quality).
MAD aims at measuring the  perceived quality under different levels of distortion~\cite{MAD:ImageQuality:Larson}. The lower {the} MAD {score}, the higher the image quality. 

%%%%%%%%%%%%%%%%%%%%%%%%%%%%%%%%%%
\subsection{Evaluation of privacy protection}
\label{sec:eval:privacy_protection}

Figure~\ref{fig:properties} compares the results of RP-FGSM with those of the eleven considered state-of-the-art attacks. For visualization purposes, all metrics are normalized using their known ranges {to 0\% and 100\%, where a higher value is more desirable}. We report undetectability, the inverse of detectability; the top-1 misleading rate with seen (unseen) classifiers obtained with ResNet50 (DenseNet161); and image quality in terms of naturalness, BRISQUE.
We report the results of the attacks when using the classifier (or combination of classifiers) with the highest misleading rate: ResNet50 for attacks that use a single classifier, and the combination of ResNet18, ResNet50 and AlexNet for attacks that use more than one classifier.

%===================================================
\usetikzlibrary{positioning, decorations.pathreplacing, calc, intersections, pgfplots.groupplots}
\pgfplotstableread{tikz/plot_attacks_R50.txt}\attack
\begin{figure}[!t]
\centering
\setlength\tabcolsep{-2pt}
\begin{tabular}{cc}
     \begin{tikzpicture}[discontinuity/.style={decoration={show path construction,lineto code={%
        \path (\tikzinputsegmentfirst) -- (\tikzinputsegmentlast) coordinate[pos=.1] (mid);%
        \draw (\tikzinputsegmentfirst) -- ([yshift=-6pt]mid) -- ++(-3pt,2pt) -- ++(6pt,2pt) -- ++(-3pt,2pt) -- (\tikzinputsegmentlast);%
      }}},discontinuity_x/.style={decoration={show path construction,lineto code={%
        \path (\tikzinputsegmentfirst) -- (\tikzinputsegmentlast) coordinate[pos=.1] (mid);%
        \draw (\tikzinputsegmentfirst) -- ([xshift=-5pt]mid) -- ++(1.5pt,-3pt) -- ++(1.5pt,6pt) -- ++(1.5pt,-3pt) -- (\tikzinputsegmentlast);%
      }}}] 
\begin{groupplot}[
    group style={
        group size=2 by 1,
    },
    xlabel={MSwD},
	ylabel={MS},
	label style={font=\footnotesize},
    tick label style={font=\footnotesize},
    ]
    \nextgroupplot[
        xmin=20, xmax=100,
	    ymin=55, ymax=100,
        xtick={40,60,80,100},
        ytick={60,80,100},
        axis x discontinuity=crunch,
        %axis y discontinuity=crunch,
        every outer y axis line/.append style={discontinuity},
        every outer x axis line/.append style={discontinuity_x},
        axis equal,
        axis y line*=left,
        axis x line*=bottom,
        width=0.5\columnwidth,
    ]
    \addplot+[
        scatter,
        mark size=1.5pt,
        only marks,
        scatter src=explicit symbolic,
        scatter/classes={
        DF={mark=square*, brown},
        SF={mark=square*,darkpowderblue},
        JSMA={mark=*, americanrose},
        CW={mark=*,blue-violet},
        N-FG={mark=square*,eggplant},
        R-FG={mark=*, blue(ryb)},
        L-FG={mark=*, rosso},
        P-FG={mark=*,falured},
        E-FG={mark=*,cadmiumgreen},
        RM-FG_T={mark=*,debianred},
        RM-FG_U={mark=square*,debianred},
        DI-FG_T={mark=*,capri},
        DI-FG_U={mark=square*,capri},
        EOT={mark=*,orange}
        }
    ]
    table[x=MS_with_defense,y=MS_no_defense,meta=Attack]{\attack};
    \end{groupplot}
    \begin{groupplot}[
    group style={
        group size=2 by 1,
    },
	label style={font=\footnotesize},
    tick label style={font=\footnotesize},
    ]
    \nextgroupplot[
        xmin=30, xmax=100,
	    ymin=55, ymax=100,
        xtick={40,60,80,100},
        xticklabel=\empty,
    yticklabel=\empty,
        ytick={60,80,100},
        axis equal,
        axis y line*=right,
        axis x line*=top,
        width=0.5\columnwidth,
    ]
    \end{groupplot}
\end{tikzpicture}
&  
\begin{tikzpicture}
[discontinuity/.style={decoration={show path construction,lineto code={%
        \path (\tikzinputsegmentfirst) -- (\tikzinputsegmentlast) coordinate[pos=.1] (mid);%
        \draw (\tikzinputsegmentfirst) -- ([yshift=-6pt]mid) -- ++(-3pt,2pt) -- ++(6pt,2pt) -- ++(-3pt,2pt) -- (\tikzinputsegmentlast);%
      }}},discontinuity_x/.style={decoration={show path construction,lineto code={%
        \path (\tikzinputsegmentfirst) -- (\tikzinputsegmentlast) coordinate[pos=.1] (mid);%
        \draw (\tikzinputsegmentfirst) -- ([xshift=-6pt]mid) -- ++(2pt,-3pt) -- ++(2pt,6pt) -- ++(2pt,-3pt) -- (\tikzinputsegmentlast);%
      }}}] 
\begin{groupplot}[
    group style={
        group size=2 by 1,
    },
    xlabel={MUwD},
	ylabel={MU},
	label style={font=\footnotesize},
    tick label style={font=\footnotesize},
    ]
    \nextgroupplot[
        xmin=30, xmax=100,
	    ymin=30, ymax=100,
        xtick={40,60,80,100},
        ytick={40,60,80,100},
        yticklabels={40,60,80,100},
        axis x discontinuity=crunch,
        %axis y discontinuity=crunch,
        every outer y axis line/.append style={discontinuity},
        every outer x axis line/.append style={discontinuity_x},
        axis equal,
        axis y line*=left,
        axis x line*=bottom,
        width=0.5\columnwidth,
    ]
    \addplot+[
    scatter,
    mark size=1.5pt,
    only marks,
    scatter src=explicit symbolic,
    scatter/classes={
    DF={mark=square*, brown},
    SF={mark=square*,darkpowderblue},
    JSMA={mark=*, americanrose},
    CW={mark=*,blue-violet},
    N-FG={mark=square*,eggplant},
    R-FG={mark=*, blue(ryb)},
    L-FG={mark=*, rosso},
    P-FG={mark=*,falured},
    E-FG={mark=*,cadmiumgreen},
    RM-FG_T={mark=*,debianred},
    RM-FG_U={mark=square*,debianred},
    DI-FG_T={mark=*,capri},
    DI-FG_U={mark=square*,capri},
    EOT={mark=*,orange}
    }
    ]
    table[x=MU_with_defense,y=MU_no_defense,meta=Attack]{\attack};
    \end{groupplot}
    \begin{groupplot}[
    group style={
        group size=2 by 1,
    },
	label style={font=\footnotesize},
    tick label style={font=\footnotesize},
    ]
    \nextgroupplot[
        xmin=30, xmax=100,
	    ymin=30, ymax=100,
        xticklabel=\empty,
        ytick={40,60,80,100},
        yticklabel=\empty,
        %axis y discontinuity=parallel,
        axis equal,
        axis y line* = right,
        axis x line* = top,
        width=0.5\columnwidth,
    ]
    \end{groupplot}
\end{tikzpicture}
\end{tabular}
%\vspace{0.5cm}
%%
\begin{tikzpicture}
[discontinuity/.style={decoration={show path construction,lineto code={%
        \path (\tikzinputsegmentfirst) -- (\tikzinputsegmentlast) coordinate[pos=.1] (mid);%
        \draw (\tikzinputsegmentfirst) -- ([yshift=-6pt]mid) -- ++(-3pt,2pt) -- ++(6pt,2pt) -- ++(-3pt,2pt) -- (\tikzinputsegmentlast);%
      }}}]  
\begin{groupplot}[
    group style={
        group size=2 by 1,
    },
    xlabel={Undetectability},
	ylabel={Image quality},
	label style={font=\footnotesize},
    tick label style={font=\footnotesize},
    ]
    \nextgroupplot[
        xmin=0, xmax=80,
    	ymin=20, ymax=60,
     	xtick={0,20,40,60,80},
        ytick={20,40,60},
        %axis y discontinuity=crunch,
        every outer y axis line/.append style={discontinuity},
        axis equal,
        axis y line*=left,
        axis x line*=bottom,
        width=0.5\columnwidth,
    ]
    \addplot+[
        scatter,
        mark size=1.5pt,
        only marks,
        scatter src=explicit symbolic,
        scatter/classes={
        DF={mark=square*, brown},
        SF={mark=square*,darkpowderblue},
        JSMA={mark=*, americanrose},
        CW={mark=*,blue-violet},
        N-FG={mark=square*,eggplant},
        R-FG={mark=*, blue(ryb)},
        L-FG={mark=*, rosso},
        P-FG={mark=*,falured},
        E-FG={mark=*,cadmiumgreen},
        RM-FG_T={mark=*,debianred},
        RM-FG_U={mark=square*,debianred},
        DI-FG_T={mark=*,capri},
        DI-FG_U={mark=square*,capri},
        EOT={mark=*,orange}
    }
    ]
    table[x=Undetectability,y=brisque,meta=Attack]{\attack};
\end{groupplot}

 \begin{groupplot}[
    group style={
        group size=2 by 1,
    },
	label style={font=\footnotesize},
    tick label style={font=\footnotesize},
    ]
    \nextgroupplot[
        xmin=30, xmax=100,
	    ymin=30, ymax=100,
        xticklabel=\empty,
        ytick={40,60,80,100},
        yticklabel=\empty,
        %axis y discontinuity=parallel,
        axis equal,
        axis y line* = right,
        axis x line* = top,
        width=0.5\columnwidth,
    ]
    \end{groupplot}
    
\end{tikzpicture}
\caption{Comparison of the privacy properties for: {JSMA~\protect\tikz \protect\draw[americanrose,fill=americanrose] (0,0) circle (.5ex);,} 
{CW~\protect\tikz \protect\draw[blue-violet,fill=blue-violet] (0,0) circle (.5ex);,}
{DeepFool~\protect\tikz \protect\draw[brown,fill=brown] (0,0) rectangle (1.ex,1.ex);,} 
{SparseFool~\protect\tikz \protect\draw[darkpowderblue,fill=darkpowderblue] (0,0) rectangle (1.ex,1.ex);,} 
{U-FGSM~\protect\tikz \protect\draw[eggplant,fill=eggplant] (0,0) rectangle (1.ex,1.ex);,} {
R-FGSM~\protect\tikz \protect\draw[blue(ryb),fill=blue(ryb)] (0,0) circle (.5ex);,} 
{L-FGSM~\protect\tikz \protect\draw[rosso,fill=rosso] (0,0) circle (.5ex);,} 
{P-FGSM~\protect\tikz \protect\draw[falured,fill=falured] (0,0) circle (.5ex);,} 
{EOT~\protect\tikz \protect\draw[orange,fill=orange] (0,0) circle (.5ex);,} 
{DI-FGSM (targeted)~\protect\tikz \protect\draw[capri,fill=capri] (0,0) circle (.5ex);,} {
DI-FGSM (untargeted)~\protect\tikz \protect\draw[capri,fill=capri] (0,0) rectangle (1.ex,1.ex);,} 
{E-FGSM~\protect\tikz \protect\draw[cadmiumgreen,fill=cadmiumgreen] (0,0) circle (.5ex);,}
{RP-FGSM (targeted)~\protect\tikz \protect\draw[debianred,fill=debianred] (0,0) circle (.5ex);} and 
{
RP-FGSM (untargeted)~\protect\tikz \protect\draw[debianred,fill=debianred] (0,0) rectangle (1.ex,1.ex);.}
For visualization purposes, all measures are normalized in the range 0-100\%. 
KEY: MS, misleading a seen classifier; MU, misleading an unseen classifier; MSwD, MS with defense; MUwD, MU with defense; 
Image quality measured by BRISQUE~\cite{Mittal2012}.
}
\label{fig:properties}
\end{figure}
%=============================

%===================================================
\begin{table*}[!t]
    \centering
    \setlength\tabcolsep{1.1pt}
    \label{tab:All-in-one}
    \caption{Privacy protection results in the test split of Private Places365 dataset for misleading rate, with and without defenses, detectability and image quality. Measures are reported as average of the dataset. Standard deviation for image quality is shown in brackets. 
    Gray shading indicates unseen classifier.
    Bold font indicates the best performing attack for a given performance measure.
    KEY --
    Att.: attack;
    R18: ResNet18; R50: ResNet50; A: AlexNet; DN: DenseNet161; Det.: Detectability;
    BRISQUE: Blind Referenceless Image Spatial Quality Evaluator; PSNR: Peak-Signal-to-Noise Ratio; MAD: Most Apparent Distortion; T1: top-1 misleading rate; T5: top-5 misleading rate;
    JSMA: Jacobian-based Saliency Map Attack;
    CW: CarliniWagner;
    DF: DeepFool; 
    SF: SparseFool;
    U-FG: Untargeted FGSM;
    R-FG: Random FGSM;
    L-FG: Least-Likely FGSM; P-FG: Private FGSM; EOT: Expectation Over Transformation;  
    DI-FG: Diverse Input FGSM;
    E-FG: Ensemble FGSM;  RP-FGSM: Robust Private FGSM, the proposed attack; T: targeted attack; U: untargeted attack.}
    \resizebox{0.99\textwidth}{!}{%
    \begin{tabular}{ll|rr|rr|rr|rr||rr|rr|rr|rr||r||rr|rr|rr}
    \Xhline{3\arrayrulewidth}
     & & \multicolumn{8}{c||}{Misleading $\uparrow$} & \multicolumn{8}{c||}{Misleading with defenses $\uparrow$} & \multicolumn{1}{c||}{\multirow{3}{*}{Det. $\downarrow$}}  & \multicolumn{6}{c}{Image quality}  \\ \cline{3-18}\cline{20-25} 
    Att. & Classifier~ & \multicolumn{2}{c|}{R18} & \multicolumn{2}{c|}{R50} & \multicolumn{2}{c|}{A} & \multicolumn{2}{c||}{DN} &  
    \multicolumn{2}{c|}{R18} & \multicolumn{2}{c|}{R50} & \multicolumn{2}{c|}{A} & \multicolumn{2}{c||}{DN} &
  & \multicolumn{2}{c|}{\multirow{2}{*}{BRISQUE~$\downarrow$}} & \multicolumn{2}{c|}{\multirow{2}{*}{PSNR~$\uparrow$}} & \multicolumn{2}{c}{\multirow{2}{*}{MAD~$\downarrow$}}   \\
  &  & T1 & T5  & T1 & T5 & T1 & T5 & T1 & T5 & T1 & T5 & T1 & T5 & T1 & T5 & T1 & T5 & & & & & & & \\ \hline
 \multirow{3}{*}{\rotatebox[origin=c]{90}{JSMA}} & R18 & 85.3 & 16.4 & \cellcolor[gray]{.9} 44.5 & \cellcolor[gray]{.9} 13.8 & \cellcolor[gray]{.9} 52.6 & \cellcolor[gray]{.9} 21.3 & \cellcolor[gray]{.9} 42.1 & \cellcolor[gray]{.9} 13.4 & 45.4 & 14.9 & \cellcolor[gray]{.9} 44.0 & \cellcolor[gray]{.9} 13.3 & \cellcolor[gray]{.9} 51.7 & \cellcolor[gray]{.9} 19.9 & \cellcolor[gray]{.9} 42.1 & \cellcolor[gray]{.9} 13.6 & 25.3 & 51.2 & (13.9) & 50.6 & (7.4) & 16.2 & (8.1)\\
&R50 & \cellcolor[gray]{.9} 46.4 & \cellcolor[gray]{.9} 16.2 & 84.0 & 14.6 & \cellcolor[gray]{.9} 52.7 & \cellcolor[gray]{.9} 21.3 & \cellcolor[gray]{.9} 42.4 & \cellcolor[gray]{.9} 13.5 & \cellcolor[gray]{.9} 44.6 & \cellcolor[gray]{.9} 15.1 & 44.3 & 13.8 & \cellcolor[gray]{.9} 51.5 & \cellcolor[gray]{.9} 19.7 & \cellcolor[gray]{.9} 42.0 & \cellcolor[gray]{.9} 13.0 & 33.9 & 51.0 & (14.0) & 50.4 & (7.4) & 12.6 & (7.3)\\
&A & \cellcolor[gray]{.9} 45.6 & \cellcolor[gray]{.9} 15.6 & \cellcolor[gray]{.9} 43.8 & \cellcolor[gray]{.9} 13.8 & 85.6 & 22.3 & \cellcolor[gray]{.9} 41.5 & \cellcolor[gray]{.9} 13.3 & \cellcolor[gray]{.9} 44.7 & \cellcolor[gray]{.9} 15.0 & \cellcolor[gray]{.9} 43.5 & \cellcolor[gray]{.9} 13.3 & 52.2 & 19.9 & \cellcolor[gray]{.9} 41.5 & \cellcolor[gray]{.9} 13.3 & 26.8 & 51.0 & (14.1) & 48.8 & (7.1) & 13.0 & (7.4)\\\hline
 \multirow{3}{*}{\rotatebox[origin=c]{90}{CW}} & R18 & 83.8 & 25.6 & \cellcolor[gray]{.9} 44.7 & \cellcolor[gray]{.9} 13.8 & \cellcolor[gray]{.9} 52.3 & \cellcolor[gray]{.9} 21.3 & \cellcolor[gray]{.9} 42.5 & \cellcolor[gray]{.9} 13.6 & 50.6 & 16.4 & \cellcolor[gray]{.9} 44.0 & \cellcolor[gray]{.9} 13.3 & \cellcolor[gray]{.9} 51.5 & \cellcolor[gray]{.9} 19.9 & \cellcolor[gray]{.9} 42.4 & \cellcolor[gray]{.9} 13.5 & 72.0 & 48.5 & (14.0) & \textbf{51.1} & (0.4) & \textbf{0.0} & (0.1)\\
&R50 & \cellcolor[gray]{.9} 46.9 & \cellcolor[gray]{.9} 16.3 & 83.8 & 22.7 & \cellcolor[gray]{.9} 52.5 & \cellcolor[gray]{.9} 21.1 & \cellcolor[gray]{.9} 43.1 & \cellcolor[gray]{.9} 13.8 & \cellcolor[gray]{.9} 45.1 & \cellcolor[gray]{.9} 15.2 & 47.8 & 14.7 & \cellcolor[gray]{.9} 51.7 & \cellcolor[gray]{.9} 19.7 & \cellcolor[gray]{.9} 42.8 & \cellcolor[gray]{.9} 12.8 & 74.4 & 48.6 & (14.0) & 51.0 & (0.4) & \textbf{0.0} & (0.1)\\
&A & \cellcolor[gray]{.9} 45.2 & \cellcolor[gray]{.9} 15.8 & \cellcolor[gray]{.9} 43.6 & \cellcolor[gray]{.9} 13.7 & 84.4 & 30.0 & \cellcolor[gray]{.9} 41.7 & \cellcolor[gray]{.9} 13.5 & \cellcolor[gray]{.9} 44.6 & \cellcolor[gray]{.9} 15.0 & \cellcolor[gray]{.9} 43.3 & \cellcolor[gray]{.9} 13.4 & 56.1 & 20.9 & \cellcolor[gray]{.9} 41.8 & \cellcolor[gray]{.9} 13.4 & 71.9 & 48.8 & (14.1) & 51.0 & (0.5) & \textbf{0.0} & (0.1)\\ \hline
\multirow{3}{*}{\rotatebox[origin=c]{90}{DF}}
&R18 & 82.6 & 28.5 & \cellcolor[gray]{.9} 47.0 & \cellcolor[gray]{.9} 16.6 & \cellcolor[gray]{.9} 45.2 & \cellcolor[gray]{.9} 15.9 & \cellcolor[gray]{.9} 42.8 & \cellcolor[gray]{.9} 13.6 & 50.5 & 16.7 & \cellcolor[gray]{.9} 43.7 & \cellcolor[gray]{.9} 13.3 & \cellcolor[gray]{.9} 51.6 & \cellcolor[gray]{.9} 20.0 & \cellcolor[gray]{.9} 42.6 & \cellcolor[gray]{.9} 13.4 & 74.6 & 48.1 & (13.7) & 50.1 & (3.7) & 1.4 & (7.4)\\
&R50 & \cellcolor[gray]{.9} 44.9 & \cellcolor[gray]{.9} 13.6 & 80.8 & 24.5 & \cellcolor[gray]{.9} 43.3 & \cellcolor[gray]{.9} 13.8 & \cellcolor[gray]{.9} 43.7 & \cellcolor[gray]{.9} 13.9 & \cellcolor[gray]{.9} 45.6 & \cellcolor[gray]{.9} 15.2 & 48.4 & 15.4 & \cellcolor[gray]{.9} 51.8 & \cellcolor[gray]{.9} 19.8 & \cellcolor[gray]{.9} 43.0 & \cellcolor[gray]{.9} 13.0 & 76.6 & 48.2 & (13.7) & 49.6 & (4.3) & 2.1 & (10.6)\\
&A & \cellcolor[gray]{.9} 52.4 & \cellcolor[gray]{.9} 21.3 & \cellcolor[gray]{.9} 52.6 & \cellcolor[gray]{.9} 21.3 & 82.1 & 32.1 & \cellcolor[gray]{.9} 42.0 & \cellcolor[gray]{.9} 13.5 & \cellcolor[gray]{.9} 44.9 & \cellcolor[gray]{.9} 14.9 & \cellcolor[gray]{.9} 43.4 & \cellcolor[gray]{.9} 13.4 & 56.5 & 21.2 & \cellcolor[gray]{.9} 41.9 & \cellcolor[gray]{.9} 13.5 & 74.3 & 48.5 & (13.7) & 50.0 & (3.5) & 2.8 & (12.2)\\\hline
 \multirow{3}{*}{\rotatebox[origin=c]{90}{SF}} & R18 & 85.9 & 18.0 & \cellcolor[gray]{.9} 46.1 & \cellcolor[gray]{.9} 14.4 & \cellcolor[gray]{.9} 55.3 & \cellcolor[gray]{.9} 23.0 & \cellcolor[gray]{.9} 43.3 & \cellcolor[gray]{.9} 14.3 & 45.4 & 15.1 & \cellcolor[gray]{.9} 44.0 & \cellcolor[gray]{.9} 13.2 & \cellcolor[gray]{.9} 51.9 & \cellcolor[gray]{.9} 20.2 & \cellcolor[gray]{.9} 42.8 & \cellcolor[gray]{.9} 13.2 & 34.8 & 53.1 & (13.9) & 38.9 & (8.6) & 44.5 & (31.7)\\
&R50 & \cellcolor[gray]{.9} 50.0 & \cellcolor[gray]{.9} 17.7 & 85.3 & 15.1 & \cellcolor[gray]{.9} 56.3 & \cellcolor[gray]{.9} 22.6 & \cellcolor[gray]{.9} 44.4 & \cellcolor[gray]{.9} 14.4 & \cellcolor[gray]{.9} 44.9 & \cellcolor[gray]{.9} 15.2 & 44.7 & 13.5 & \cellcolor[gray]{.9} 52.7 & \cellcolor[gray]{.9} 21.2 & \cellcolor[gray]{.9} 42.7 & \cellcolor[gray]{.9} 13.0 & 42.1 & 52.9 & (13.9) & 38.4 & (8.7) & 45.1 & (32.0)\\
&A & \cellcolor[gray]{.9} 46.4 & \cellcolor[gray]{.9} 16.3 & \cellcolor[gray]{.9} 44.2 & \cellcolor[gray]{.9} 14.0 & 85.9 & 23.0 & \cellcolor[gray]{.9} 41.7 & \cellcolor[gray]{.9} 13.6 & \cellcolor[gray]{.9} 45.5 & \cellcolor[gray]{.9} 15.1 & \cellcolor[gray]{.9} 43.7 & \cellcolor[gray]{.9} 13.5 & 53.6 & 20.4 & \cellcolor[gray]{.9} 41.8 & \cellcolor[gray]{.9} 13.3 & \textbf{22.4} & 51.4 & (14.0) & 40.7 & (7.7) & 28.5 & (24.9)\\
\hline
\hline
 \multirow{3}{*}{\rotatebox[origin=c]{90}{U-FG}} & R18 & 92.4 & 86.1 & \cellcolor[gray]{.9} 86.6 & \cellcolor[gray]{.9} 55.9 & \cellcolor[gray]{.9} 67.8 & \cellcolor[gray]{.9} 34.3 & \cellcolor[gray]{.9} 85.5 & \cellcolor[gray]{.9} 53.5 & 86.1 & 59.7 & \cellcolor[gray]{.9} 67.0 & \cellcolor[gray]{.9} 32.6 & \cellcolor[gray]{.9} 62.5 & \cellcolor[gray]{.9} 29.4 & \cellcolor[gray]{.9} 67.1 & \cellcolor[gray]{.9} 33.4 & 58.1 & 45.6 & (12.5) & 32.0 & (0.9) & 26.1 & (8.3)\\
&R50 & \cellcolor[gray]{.9} 85.9 & \cellcolor[gray]{.9} 54.4 & 91.7 & 84.3 & \cellcolor[gray]{.9} 65.0 & \cellcolor[gray]{.9} 31.0 & \cellcolor[gray]{.9} 86.1 & \cellcolor[gray]{.9} 53.3 & \cellcolor[gray]{.9} 69.6 & \cellcolor[gray]{.9} 34.1 & 80.3 & 47.4 & \cellcolor[gray]{.9} 61.5 & \cellcolor[gray]{.9} 27.0 & \cellcolor[gray]{.9} 67.2 & \cellcolor[gray]{.9} 32.4 & 52.1 & 45.3 & (12.8) & 32.5 & (0.9) & 24.1 & (8.3)\\
&A & \cellcolor[gray]{.9} 72.3 & \cellcolor[gray]{.9} 40.2 & \cellcolor[gray]{.9} 67.0 & \cellcolor[gray]{.9} 32.9 & 93.8 & 88.7 & \cellcolor[gray]{.9} 63.5 & \cellcolor[gray]{.9} 28.6 & \cellcolor[gray]{.9} 66.4 & \cellcolor[gray]{.9} 31.8 & \cellcolor[gray]{.9} 60.8 & \cellcolor[gray]{.9} 27.2 & 91.2 & 73.8 & \cellcolor[gray]{.9} 58.0 & \cellcolor[gray]{.9} 24.5 & 47.2 & \textbf{29.7} & (17.1) & 29.9 & (0.8) & 34.0 & (9.2)\\\hline
 \multirow{3}{*}{\rotatebox[origin=c]{90}{R-FG}} & R18 & 72.3 & 40.2 & \cellcolor[gray]{.9} 67.0 & \cellcolor[gray]{.9} 32.9 & \cellcolor[gray]{.9} 93.8 & \cellcolor[gray]{.9} 88.7 & \cellcolor[gray]{.9} 63.5 & \cellcolor[gray]{.9} 28.6 & 69.9 & 40.6 & \cellcolor[gray]{.9} 53.3 & \cellcolor[gray]{.9} 21.2 & \cellcolor[gray]{.9} 56.5 & \cellcolor[gray]{.9} 25.6 & \cellcolor[gray]{.9} 50.2 & \cellcolor[gray]{.9} 19.5 & 94.3 & 45.9 & (12.7) & 33.1 & (1.0) & 20.7 & (8.3)\\
&R50 & \cellcolor[gray]{.9} 63.1 & \cellcolor[gray]{.9} 32.7 & 99.9 & 97.0 & \cellcolor[gray]{.9} 59.3 & \cellcolor[gray]{.9} 27.2 & \cellcolor[gray]{.9} 59.3 & \cellcolor[gray]{.9} 27.4 & \cellcolor[gray]{.9} 55.3 & \cellcolor[gray]{.9} 23.0 & 63.4 & 31.4 & \cellcolor[gray]{.9} 56.0 & \cellcolor[gray]{.9} 24.2 & \cellcolor[gray]{.9} 49.9 & \cellcolor[gray]{.9} 18.8 & 95.8 & 44.9 & (13.0) & 33.4 & (1.0) & 20.2 & (8.3)\\
&A & \cellcolor[gray]{.9} 53.9 & \cellcolor[gray]{.9} 23.1 & \cellcolor[gray]{.9} 52.3 & \cellcolor[gray]{.9} 19.9 & 99.8 & 96.4 & \cellcolor[gray]{.9} 49.6 & \cellcolor[gray]{.9} 17.8 & \cellcolor[gray]{.9} 52.1 & \cellcolor[gray]{.9} 20.0 & \cellcolor[gray]{.9} 50.1 & \cellcolor[gray]{.9} 19.0 & 74.0 & 46.5 & \cellcolor[gray]{.9} 47.6 & \cellcolor[gray]{.9} 17.0 & 96.7 & 34.7 & (16.5) & 32.4 & (1.0) & 19.3 & (8.2)\\\hline
 \multirow{3}{*}{\rotatebox[origin=c]{90}{L-FG}} & R18 & \textbf{100.0} & \textbf{99.8} & \cellcolor[gray]{.9} 64.4 & \cellcolor[gray]{.9} 34.4 & \cellcolor[gray]{.9} 62.0 & \cellcolor[gray]{.9} 30.1 & \cellcolor[gray]{.9} 61.5 & \cellcolor[gray]{.9} 32.0 & 74.1 & 45.3 & \cellcolor[gray]{.9} 54.5 & \cellcolor[gray]{.9} 23.4 & \cellcolor[gray]{.9} 58.4 & \cellcolor[gray]{.9} 27.1 & \cellcolor[gray]{.9} 52.1 & \cellcolor[gray]{.9} 21.3 & 99.3 & 46.3 & (12.6) & 32.7 & (1.0) & 19.2 & (8.2)\\
&R50 & \cellcolor[gray]{.9} 68.0 & \cellcolor[gray]{.9} 38.8 & \textbf{100.0} & \textbf{99.7} & \cellcolor[gray]{.9} 60.3 & \cellcolor[gray]{.9} 28.1 & \cellcolor[gray]{.9} 62.4 & \cellcolor[gray]{.9} 32.7 & \cellcolor[gray]{.9} 57.0 & \cellcolor[gray]{.9} 25.0 & 66.3 & 34.6 & \cellcolor[gray]{.9} 57.2 & \cellcolor[gray]{.9} 25.2 & \cellcolor[gray]{.9} 52.9 & \cellcolor[gray]{.9} 22.2 & 99.9 & 45.1 & (13.1) & 32.9 & (0.9) & 18.7 & (8.3)\\
&A & \cellcolor[gray]{.9} 56.9 & \cellcolor[gray]{.9} 25.8 & \cellcolor[gray]{.9} 53.4 & \cellcolor[gray]{.9} 21.7 & \textbf{100.0} & \textbf{99.7} & \cellcolor[gray]{.9} 51.3 & \cellcolor[gray]{.9} 19.5 & \cellcolor[gray]{.9} 54.0 & \cellcolor[gray]{.9} 23.2 & \cellcolor[gray]{.9} 51.6 & \cellcolor[gray]{.9} 19.9 & 79.0 & 53.4 & \cellcolor[gray]{.9} 49.5 & \cellcolor[gray]{.9} 18.4 & 99.6 & 34.9 & (16.5) & 32.0 & (0.8) & 19.6 & (8.3)\\\hline
\multirow{3}{*}{\rotatebox[origin=c]{90}{P-FG}} & R18 & \textbf{100.0} & 97.4 & \cellcolor[gray]{.9} 60.4 & \cellcolor[gray]{.9} 29.0 & \cellcolor[gray]{.9} 60.7 & \cellcolor[gray]{.9} 28.3 & \cellcolor[gray]{.9} 57.0 & \cellcolor[gray]{.9} 26.9 & 71.2 & 39.1 & \cellcolor[gray]{.9} 53.5 & \cellcolor[gray]{.9} 20.7 & \cellcolor[gray]{.9} 57.3 & \cellcolor[gray]{.9} 25.4 & \cellcolor[gray]{.9} 51.3 & \cellcolor[gray]{.9} 19.2 & 93.6 & 45.9 & (12.7) & 33.2 & (1.0) & 22.4 & (8.4)\\
&R50 & \cellcolor[gray]{.9} 63.8 & \cellcolor[gray]{.9} 33.0 & 99.8 & 97.0 & \cellcolor[gray]{.9} 59.5 & \cellcolor[gray]{.9} 26.9 & \cellcolor[gray]{.9} 58.6 & \cellcolor[gray]{.9} 27.3 & \cellcolor[gray]{.9} 55.0 & \cellcolor[gray]{.9} 23.1 & 63.2 & 30.7 & \cellcolor[gray]{.9} 56.6 & \cellcolor[gray]{.9} 24.7 & \cellcolor[gray]{.9} 51.5 & \cellcolor[gray]{.9} 19.3 & 96.2 & 44.9 & (13.0) & 33.3 & (1.0) & 21.4 & (8.5)\\
&A & \cellcolor[gray]{.9} 55.1 & \cellcolor[gray]{.9} 23.4 & \cellcolor[gray]{.9} 52.9 & \cellcolor[gray]{.9} 19.7 & 99.7 & 96.1 & \cellcolor[gray]{.9} 49.6 & \cellcolor[gray]{.9} 18.5 & \cellcolor[gray]{.9} 52.5 & \cellcolor[gray]{.9} 20.6 & \cellcolor[gray]{.9} 50.2 & \cellcolor[gray]{.9} 19.0 & 75.1 & 47.0 & \cellcolor[gray]{.9} 47.8 & \cellcolor[gray]{.9} 17.4 & 95.4 & 34.8 & (16.4) & 32.4 & (0.9) & 24.4 & (9.0)\\ \hline
%%%%%%%%%%%%%%%%
\multirow{3}{*}{\rotatebox[origin=c]{90}{EOT}} & R18 &  94.0 &	72.0&	\cellcolor[gray]{.9}48.3&	\cellcolor[gray]{.9}17.5&	\cellcolor[gray]{.9}54.2&	\cellcolor[gray]{.9}22.5&	\cellcolor[gray]{.9}46.1&	\cellcolor[gray]{.9}15.8& 56.8&	23.0&\cellcolor[gray]{.9}	46.9&\cellcolor[gray]{.9}	15.5&\cellcolor[gray]{.9}	52.8&\cellcolor[gray]{.9}	21.6&	\cellcolor[gray]{.9}44.0&	\cellcolor[gray]{.9}14.4 & 93.4& 45.3 & (13.6) & 42.3 & (2.1)  & 3.2 & (4.0)\\
& R50 & \cellcolor[gray]{.9}50.9&	\cellcolor[gray]{.9}20.6&	91.3&	67.4&\cellcolor[gray]{.9}	\cellcolor[gray]{.9}54.0&\cellcolor[gray]{.9}	22.2&\cellcolor[gray]{.9}	47.1&\cellcolor[gray]{.9}	16.9&\cellcolor[gray]{.9} 48.6&\cellcolor[gray]{.9}	17.9&	52.3&	19.8&\cellcolor[gray]{.9}	52.9&	\cellcolor[gray]{.9}21.2&	\cellcolor[gray]{.9}44.1&\cellcolor[gray]{.9}	14.8 & 93.2&45.3 & (13.6)& 42.8 & (1.9) & 3.4 & (4.2)\\
& A &  \cellcolor[gray]{.9}46.9&\cellcolor[gray]{.9}	16.9&\cellcolor[gray]{.9}	45.3&\cellcolor[gray]{.9}	14.2&	90.8&	72.7&\cellcolor[gray]{.9}	42.3&\cellcolor[gray]{.9}	13.7&\cellcolor[gray]{.9} 45.2&\cellcolor[gray]{.9}	15.3&\cellcolor[gray]{.9}	44.3&\cellcolor[gray]{.9}	13.7&	62.2&	31.6&	\cellcolor[gray]{.9}42.3&\cellcolor[gray]{.9}	13.1 & 93.9& 43.8 & (14.4)& 43.3& (2.0) & 3.6 & (4.3) \\ \hline
\multirow{7}{*}{\rotatebox[origin=c]{90}{DI-FG (T)}} 
 & R18 & \textbf{100.0} & 98.9 &\cellcolor[gray]{.9} 71.1 & \cellcolor[gray]{.9} 43.1 & \cellcolor[gray]{.9} 71.5  & \cellcolor[gray]{.9} 42.3 & \cellcolor[gray]{.9} 68.4 & \cellcolor[gray]{.9}39.6 & 82.7 & 56.2 & \cellcolor[gray]{.9}61.5 & \cellcolor[gray]{.9}30.1 & \cellcolor[gray]{.9}60.1 & \cellcolor[gray]{.9}29.2 & \cellcolor[gray]{.9}58.2 & \cellcolor[gray]{.9}26.9 & 82.8 	& 43.6 & (13.5) & 32.8 & (0.7) & 	26.2 & (8.8)\\
 & R50 & \cellcolor[gray]{.9} 80.8 &\cellcolor[gray]{.9}	54.5 & \textbf{100.0} & 98.3 & \cellcolor[gray]{.9} 64.5  & \cellcolor[gray]{.9} 32.6 &\cellcolor[gray]{.9} 76.6 & \cellcolor[gray]{.9}49.6 & \cellcolor[gray]{.9}66.4 & \cellcolor[gray]{.9}36.1 & 74.8 & 44.7 &  \cellcolor[gray]{.9}59.6 & \cellcolor[gray]{.9}28.5 & \cellcolor[gray]{.9} 61.8 & \cellcolor[gray]{.9} 30.5 & 86.7 &43.6 & (13.5) & 33.0 & (0.7) & 25.3 &	(8.9)\\
 & A   & \cellcolor[gray]{.9} 60.5 &\cellcolor[gray]{.9}	28.0 & \cellcolor[gray]{.9}54.8	 & \cellcolor[gray]{.9}23.4 & \textbf{100.0}	& 97.9& \cellcolor[gray]{.9} 52.5 & \cellcolor[gray]{.9} 20.4  & \cellcolor[gray]{.9} 56.9 & \cellcolor[gray]{.9} 25.7 & \cellcolor[gray]{.9} 53.9 & \cellcolor[gray]{.9} 22.1 & 85.2 & 64.0  & \cellcolor[gray]{.9} 50.9 & \cellcolor[gray]{.9}18.8 & 86.6	& 36.6 & (16.1) & 32.6 & (0.6) & 26.5 &	(9.1)\\
 & R18+R50 & \textbf{100.0} & 99.1 & \textbf{100.0} & 98.8 & \cellcolor[gray]{.9} 68.4 & \cellcolor[gray]{.9} 37.1 & \cellcolor[gray]{.9} 86.5 & \cellcolor[gray]{.9} 62.8 & 83.8 & 59.9 & 77.0 & 49.2 & \cellcolor[gray]{.9} 63.2 &  \cellcolor[gray]{.9}31.8 & \cellcolor[gray]{.9} 68.7 & \cellcolor[gray]{.9} 38.9   & 85.0 & 43.5 & (13.5) & 32.7 & (0.7) & 27.6 &	(8.9) \\ 
 & R18+A & \textbf{100.0} & 98.6 & \cellcolor[gray]{.9}71.5 & \cellcolor[gray]{.9}42.3 & \textbf{100.0} & 98.3 & \cellcolor[gray]{.9}67.4 & \cellcolor[gray]{.9}37.8 & 79.9 & 54.9 & \cellcolor[gray]{.9}63.1 & \cellcolor[gray]{.9} 31.8 & 83.6 & 62.0 & \cellcolor[gray]{.9}58.9 & \cellcolor[gray]{.9}28.7     & 87.8 	& 39.6 & (15.0) & 32.2 & (0.6) & 28.2 & (9.0)\\
 & R50+A & \cellcolor[gray]{.9} 83.2 & \cellcolor[gray]{.9} 59.6 & \textbf{100.0} & 98.2 & \textbf{100.0} & 98.4 & \cellcolor[gray]{.9} 75.2 & \cellcolor[gray]{.9} 48.1 & \cellcolor[gray]{.9} 71.4 &  \cellcolor[gray]{.9} 42.4 & 74.1 & 45.2 & 82.8 & 61.4 & \cellcolor[gray]{.9} 62.8 & \cellcolor[gray]{.9} 32.5     & 92.2 & 39.8 & (14.9) & 32.3 & (0.6)  & 27.5 	& (9.1)\\
 & R18+R50+A & \textbf{100.0} & 98.5 & \textbf{100.0} & 98.4 & 99.9 & 98.2 &\cellcolor[gray]{.9} 87.1 & \cellcolor[gray]{.9} 65.0 & 83.2 & 59.0 & 76.7 & 48.5 & 83.3 & 61.6 & \cellcolor[gray]{.9} 69.4 & \cellcolor[gray]{.9} 40.1 & 89.0 	& 40.5 & (14.7) & 32.0 & (0.7)  & 	29.1 &	(9.1)
\\ \hline
 \multirow{7}{*}{\rotatebox[origin=c]{90}{DI-FG (U)}} & R18 & 92.4 & 86.0 & \cellcolor[gray]{.9} 89.5 & \cellcolor[gray]{.9} 67.6 & \cellcolor[gray]{.9} 73.5 & \cellcolor[gray]{.9} 41.3 & \cellcolor[gray]{.9} 89.1 & \cellcolor[gray]{.9} 64.6 & 89.7 & 71.7 & \cellcolor[gray]{.9} 74.6 & \cellcolor[gray]{.9} 43.4 & \cellcolor[gray]{.9} 67.6 & \cellcolor[gray]{.9} 34.8 & \cellcolor[gray]{.9} 73.9 & \cellcolor[gray]{.9} 42.1 & 53.2 & 43.4 & (13.5) & 32.9 & (0.7) & 25.8 & (8.5)\\
&R50 & \cellcolor[gray]{.9} 89.7 & \cellcolor[gray]{.9} 68.2 & 91.8 & 84.5 & \cellcolor[gray]{.9} 70.6 & \cellcolor[gray]{.9} 37.7 & \cellcolor[gray]{.9} 90.0 & \cellcolor[gray]{.9} 67.0 & \cellcolor[gray]{.9} 77.5 & \cellcolor[gray]{.9} 46.6 & 87.0 & 61.5 & \cellcolor[gray]{.9} 65.9 & \cellcolor[gray]{.9} 31.3 & \cellcolor[gray]{.9} 76.7 & \cellcolor[gray]{.9} 42.0 & 46.6 & 43.8 & (13.5) & 33.2 & (0.7) & 24.4 & (8.5)\\
&A & \cellcolor[gray]{.9} 74.3 & \cellcolor[gray]{.9} 42.8 & \cellcolor[gray]{.9} 66.5 & \cellcolor[gray]{.9} 32.7 & 94.3 & 89.6 & \cellcolor[gray]{.9} 62.6 & \cellcolor[gray]{.9} 28.3 & \cellcolor[gray]{.9} 70.2 & \cellcolor[gray]{.9} 36.7 & \cellcolor[gray]{.9} 64.0 & \cellcolor[gray]{.9} 29.9 & 92.9 & 78.5 & \cellcolor[gray]{.9} 59.4 & \cellcolor[gray]{.9} 26.4 & 46.8 & 34.9 & (16.6) & 32.2 & (0.6) & 27.5 & (9.0)\\
&R18+R50 & 93.2 & 87.2 & 93.1 & 86.2 & \cellcolor[gray]{.9} 78.0 & \cellcolor[gray]{.9} 46.3 & \cellcolor[gray]{.9} 92.2 & \cellcolor[gray]{.9} 76.1 & 90.4 & 72.9 & 88.0 & 64.0 & \cellcolor[gray]{.9} 72.0 & \cellcolor[gray]{.9} 37.7 & \cellcolor[gray]{.9} 81.7 & \cellcolor[gray]{.9} 51.5 & 42.7 & 43.5 & (13.5) & 32.7 & (0.7) & 26.4 & (8.6)\\
&R18+A & 92.2 & 86.1 & \cellcolor[gray]{.9} 89.6 & \cellcolor[gray]{.9} 67.3 & 94.0 & 87.8 & \cellcolor[gray]{.9} 88.4 & \cellcolor[gray]{.9} 64.4 & 89.4 & 70.5 & \cellcolor[gray]{.9} 76.8 & \cellcolor[gray]{.9} 46.1 & 90.3 & 72.5 & \cellcolor[gray]{.9} 76.6 & \cellcolor[gray]{.9} 44.5 & 51.1 & 40.0 & (14.4) & 32.3 & (0.7) & 27.6 & (8.7)\\
&R50+A & \cellcolor[gray]{.9} 88.9 & \cellcolor[gray]{.9} 68.3 & 91.2 & 83.4 & 93.6 & 87.6 & \cellcolor[gray]{.9} 87.9 & \cellcolor[gray]{.9} 64.7 & \cellcolor[gray]{.9} 79.8 & \cellcolor[gray]{.9} 50.6 & 84.7 & 59.3 & 90.2 & 72.1 & \cellcolor[gray]{.9} 76.8 & \cellcolor[gray]{.9} 43.7 & 42.0 & 40.0 & (14.4) & 32.5 & (0.7) & 26.4 & (8.6)\\
&R18+R50+A & 93.6 & 87.4 & 93.6 & 86.4 & 94.3 & 87.4 & \cellcolor[gray]{.9} 92.4 & \cellcolor[gray]{.9} 76.0 & 90.7 & 72.9 & 87.5 & 64.3 & 89.1 & 70.4 & \cellcolor[gray]{.9} 83.4 & \cellcolor[gray]{.9} 53.7 & 41.0 & 41.1 & (14.0) & 32.3 & (0.7) & 27.6 & (8.7)\\\hline
\multirow{7}{*}{\rotatebox[origin=c]{90}{E-FG}} & R18 & \textbf{100.0} & 96.4 & \cellcolor[gray]{.9} 51.9 & \cellcolor[gray]{.9} 20.2 & \cellcolor[gray]{.9} 56.2 & \cellcolor[gray]{.9} 23.7 & \cellcolor[gray]{.9} 48.1 & \cellcolor[gray]{.9} 18.6 & 64.5 & 32.8 & \cellcolor[gray]{.9} 48.5 & \cellcolor[gray]{.9} 18.0 & \cellcolor[gray]{.9} 54.1 & \cellcolor[gray]{.9} 22.1 & \cellcolor[gray]{.9} 46.2 & \cellcolor[gray]{.9} 15.8 & 99.9 & 45.3 & (13.3) & 37.5 & (2.3) & 9.4 & (7.5)\\
&R50 & \cellcolor[gray]{.9} 55.1 & \cellcolor[gray]{.9} 24.9 & \textbf{100.0} & 95.9 & \cellcolor[gray]{.9} 55.8 & \cellcolor[gray]{.9} 23.8 & \cellcolor[gray]{.9} 50.7 & \cellcolor[gray]{.9} 19.7 & \cellcolor[gray]{.9} 51.2 & \cellcolor[gray]{.9} 20.2 & 58.8 & 27.1 & \cellcolor[gray]{.9} 53.8 & \cellcolor[gray]{.9} 21.9 & \cellcolor[gray]{.9} 46.0 & \cellcolor[gray]{.9} 16.2 & 99.9 & 44.7 & (13.4) & 37.3 & (2.3) & 9.7 & (7.6)\\
&A & \cellcolor[gray]{.9} 49.2 & \cellcolor[gray]{.9} 19.4 & \cellcolor[gray]{.9} 48.5 & \cellcolor[gray]{.9} 16.3 & \textbf{100.0} & 97.0 & \cellcolor[gray]{.9} 45.5 & \cellcolor[gray]{.9} 15.5 & \cellcolor[gray]{.9} 48.9 & \cellcolor[gray]{.9} 17.6 & \cellcolor[gray]{.9} 47.4 & \cellcolor[gray]{.9} 15.6 & 69.2 & 39.7 & \cellcolor[gray]{.9} 44.9 & \cellcolor[gray]{.9} 14.5 & 99.8 & 38.9 & (15.9) & 36.1 & (2.3) & 13.3 & (8.7)\\
&R18+R50 & \textbf{100.0} & 97.4 & \textbf{100.0} & 96.9 & \cellcolor[gray]{.9} 58.1 & \cellcolor[gray]{.9} 26.1 & \cellcolor[gray]{.9} 58.7 & \cellcolor[gray]{.9} 27.3 & 67.0 & 35.6 & 60.9 & 29.0 & \cellcolor[gray]{.9} 55.1 & \cellcolor[gray]{.9} 23.9 & \cellcolor[gray]{.9} 50.1 & \cellcolor[gray]{.9} 19.9 & 100.0 & 44.9 & (13.3) & 35.7 & (2.4) & 13.9 & (9.1)\\
&R18+A & \textbf{100.0} & 96.5 & \cellcolor[gray]{.9} 55.1 & \cellcolor[gray]{.9} 24.0 & \textbf{100.0} & 96.7 & \cellcolor[gray]{.9} 52.7 & \cellcolor[gray]{.9} 21.8 & 66.6 & 36.5 & \cellcolor[gray]{.9} 50.9 & \cellcolor[gray]{.9} 19.5 & 70.6 & 42.1 & \cellcolor[gray]{.9} 48.9 & \cellcolor[gray]{.9} 18.1 & 99.8 & 41.1 & (14.7) & 34.6 & (2.3) & 17.5 & (9.8)\\
&R50+A & \cellcolor[gray]{.9} 59.6 & \cellcolor[gray]{.9} 28.9 & \textbf{100.0} & 96.1 & \textbf{100.0} & 96.6 & \cellcolor[gray]{.9} 54.7 & \cellcolor[gray]{.9} 22.6 & \cellcolor[gray]{.9} 54.4 & \cellcolor[gray]{.9} 22.3 & 59.6 & 28.2 & 70.1 & 41.3 & \cellcolor[gray]{.9} 49.4 & \cellcolor[gray]{.9} 18.4 & 99.9 & 40.9 & (14.7) & 34.6 & (2.2) & 17.2 & (9.7)\\
&R18+R50+A & \textbf{100.0} & 97.4 & \textbf{100.0} & 96.8 & \textbf{100.0} & 97.3 & \cellcolor[gray]{.9} 62.3 & \cellcolor[gray]{.9} 31.4 & 73.5 & 43.8 & 66.2 & 35.2 & 73.9 & 46.5 & \cellcolor[gray]{.9} 57.0 & \cellcolor[gray]{.9} 24.7 & 99.9 & 41.8 & (14.2) & 33.8 & (2.3) & 19.8 & (9.6)\\\hline
\multirow{7}{*}{\rotatebox[origin=c]{90}{{\bf RP-FG (T)}}} & R18 & \textbf{100.0} & 98.4 & \cellcolor[gray]{.9} 62.6 & \cellcolor[gray]{.9} 31.8 & \cellcolor[gray]{.9} 62.3 & \cellcolor[gray]{.9} 29.7 & \cellcolor[gray]{.9} 60.4 & \cellcolor[gray]{.9} 30.1 & 94.4 & 80.1 & \cellcolor[gray]{.9} 58.6 & \cellcolor[gray]{.9} 27.9 & \cellcolor[gray]{.9} 62.0 & \cellcolor[gray]{.9} 30.6 & \cellcolor[gray]{.9} 56.0 & \cellcolor[gray]{.9} 25.1 & 67.0 & 44.8 & (13.2) & 33.6 & (0.6) & 23.0 & (9.0)\\
&R50 & \cellcolor[gray]{.9} 68.2 & \cellcolor[gray]{.9} 38.6 & \textbf{100.0} & 97.8 & \cellcolor[gray]{.9} 61.0 & \cellcolor[gray]{.9} 28.9 & \cellcolor[gray]{.9} 62.8 & \cellcolor[gray]{.9} 31.8 & \cellcolor[gray]{.9} 62.5 & \cellcolor[gray]{.9} 31.5 & 85.7 & 59.6 & \cellcolor[gray]{.9} 61.0 & \cellcolor[gray]{.9} 29.1 & \cellcolor[gray]{.9} 57.4 & \cellcolor[gray]{.9} 27.6 & 79.0 & 44.2 & (13.4) & 33.8 & (0.5) & 22.2 & (9.0)\\
&A & \cellcolor[gray]{.9} 54.6 & \cellcolor[gray]{.9} 23.8 & \cellcolor[gray]{.9} 51.5 & \cellcolor[gray]{.9} 20.4 & \textbf{100.0} & 98.1 & \cellcolor[gray]{.9} 49.4 & \cellcolor[gray]{.9} 18.3 & \cellcolor[gray]{.9} 53.0 & \cellcolor[gray]{.9} 22.6 & \cellcolor[gray]{.9} 50.8 & \cellcolor[gray]{.9} 20.0 & 91.0 & 74.2 & \cellcolor[gray]{.9} 48.6 & \cellcolor[gray]{.9} 18.0 & 64.1 & 37.7 & (15.7) & 33.2 & (0.6) & 23.4 & (9.1)\\
&R18+R50 & \textbf{100.0} & 98.9 & \textbf{100.0} & 98.7 & \cellcolor[gray]{.9} 68.2 & \cellcolor[gray]{.9} 39.1 & \cellcolor[gray]{.9} 84.4 & \cellcolor[gray]{.9} 59.6 & 96.5 & 85.5 & 93.7 & 80.1 & \cellcolor[gray]{.9} 65.4 & \cellcolor[gray]{.9} 35.0 & \cellcolor[gray]{.9} 75.7 & \cellcolor[gray]{.9} 49.0 & 58.6 & 30.6 & (7.0) & 31.0 & (0.5) & 34.0 & (10.9)\\
&R18+A & \textbf{100.0} & 98.7 & \cellcolor[gray]{.9} 71.4 & \cellcolor[gray]{.9} 43.5 & \textbf{100.0} & 98.2 & \cellcolor[gray]{.9} 67.2 & \cellcolor[gray]{.9} 39.8 & 93.8 & 78.2 & \cellcolor[gray]{.9} 66.5 & \cellcolor[gray]{.9} 36.5 & 91.9 & 75.8 & \cellcolor[gray]{.9} 63.0 & \cellcolor[gray]{.9} 33.3 & 65.5 & 30.6 & (7.6) & 30.8 & (0.5) & 33.9 & (10.7)\\
&R50+A & \cellcolor[gray]{.9} 79.9 & \cellcolor[gray]{.9} 53.2 & 99.9 & 98.6 & \textbf{100.0} & 98.6 & \cellcolor[gray]{.9} 70.5 & \cellcolor[gray]{.9} 42.3 & \cellcolor[gray]{.9} 72.8 & \cellcolor[gray]{.9} 44.6 & 90.3 & 72.4 & 91.7 & 75.8 & \cellcolor[gray]{.9} 65.8 & \cellcolor[gray]{.9} 35.8 & 72.9 & 30.1 & (7.5) & 30.8 & (0.5) & 33.6 & (10.8)\\
&R18+R50+A & \textbf{100.0} & 99.1 & \textbf{100.0} & 98.6 & \textbf{100.0} & 98.2 & \cellcolor[gray]{.9} 86.5 & \cellcolor[gray]{.9} 64.6 & \textbf{96.8} & 88.0 & 93.3 & 79.3 & 91.7 & 77.7 & \cellcolor[gray]{.9} 79.4 & \cellcolor[gray]{.9} 55.1 & 55.9 & 33.4 & (7.4) & 29.6 & (0.4) & 39.8 & (11.5)\\\hline
 \multirow{7}{*}{\rotatebox[origin=c]{90}{{\bf RP-FG (U)}}} & R18 & 93.5 & 88.5 & \cellcolor[gray]{.9} 88.3 & \cellcolor[gray]{.9} 62.6 & \cellcolor[gray]{.9} 70.1 & \cellcolor[gray]{.9} 37.7 & \cellcolor[gray]{.9} 86.4 & \cellcolor[gray]{.9} 59.0 & 93.2 & 83.9 & \cellcolor[gray]{.9} 80.7 & \cellcolor[gray]{.9} 53.3 & \cellcolor[gray]{.9} 67.1 & \cellcolor[gray]{.9} 34.3 & \cellcolor[gray]{.9} 80.9 & \cellcolor[gray]{.9} 50.0 & 50.6 & 44.6 & (13.1) & 33.3 & (0.6) & 23.7 & (8.8)\\
&R50 & \cellcolor[gray]{.9} 88.3 & \cellcolor[gray]{.9} 63.1 & 92.9 & 86.9 & \cellcolor[gray]{.9} 67.6 & \cellcolor[gray]{.9} 34.9 & \cellcolor[gray]{.9} 88.1 & \cellcolor[gray]{.9} 60.5 & \cellcolor[gray]{.9} 83.6 & \cellcolor[gray]{.9} 55.2 & 92.3 & 79.3 & \cellcolor[gray]{.9} 65.4 & \cellcolor[gray]{.9} 32.1 & \cellcolor[gray]{.9} 81.7 & \cellcolor[gray]{.9} 52.7 & 42.9 & 44.2 & (13.4) & 33.6 & (0.6) & 22.5 & (8.8)\\
&A & \cellcolor[gray]{.9} 67.6 & \cellcolor[gray]{.9} 34.4 & \cellcolor[gray]{.9} 60.1 & \cellcolor[gray]{.9} 26.7 & 95.6 & 91.1 & \cellcolor[gray]{.9} 56.8 & \cellcolor[gray]{.9} 22.8 & \cellcolor[gray]{.9} 66.0 & \cellcolor[gray]{.9} 33.1 & \cellcolor[gray]{.9} 59.6 & \cellcolor[gray]{.9} 26.1 & \textbf{94.9} & 86.6 & \cellcolor[gray]{.9} 56.7 & \cellcolor[gray]{.9} 22.7 & 47.3 & 36.1 & (15.9) & 32.4 & (0.5) & 26.1 & (9.2)\\
&R18+R50 & 95.1 & 90.6 & 94.4 & 88.6 & \cellcolor[gray]{.9} 79.6 & \cellcolor[gray]{.9} 50.2 & \cellcolor[gray]{.9} 93.5 & \cellcolor[gray]{.9} 78.5 & 93.6 & 83.3 & 93.8 & 84.5 & \cellcolor[gray]{.9} 76.2 & \cellcolor[gray]{.9} 45.9 & \cellcolor[gray]{.9} 89.8 & \cellcolor[gray]{.9} 67.5 & 41.7 & 45.3 & (12.5) & 30.9 & (0.5) & 34.1 & (10.8)\\
&R18+A & 94.7 & 89.6 & \cellcolor[gray]{.9} 90.9 & \cellcolor[gray]{.9} 69.3 & 95.3 & 90.3 & \cellcolor[gray]{.9} 89.2 & \cellcolor[gray]{.9} 66.2 & 92.8 & 79.3 & \cellcolor[gray]{.9} 81.4 & \cellcolor[gray]{.9} 52.8 & \textbf{94.9} & 88.1 & \cellcolor[gray]{.9} 80.8 & \cellcolor[gray]{.9} 51.7 & 57.6 & 40.4 & (13.9) & 30.5 & (0.5) & 34.8 & (10.6)\\
&R50+A & \cellcolor[gray]{.9} 91.6 & \cellcolor[gray]{.9} 72.5 & 94.0 & 87.8 & 95.2 & 90.3 & \cellcolor[gray]{.9} 90.6 & \cellcolor[gray]{.9} 68.0 & \cellcolor[gray]{.9} 85.4 & \cellcolor[gray]{.9} 59.8 & 90.1 & 70.3 & 94.7 & \textbf{88.4} & \cellcolor[gray]{.9} 82.0 & \cellcolor[gray]{.9} 50.7 & 49.0 & 40.0 & (13.9) & 30.6 & (0.5) & 34.2 & (10.6)\\
&R18+R50+A & 97.0 & 93.8 & 95.9 & 91.0 & 95.7 & 91.0 & \cellcolor[gray]{.9} \textbf{95.2} & \cellcolor[gray]{.9} \textbf{82.6} & {96.5} & \textbf{91.7} & \textbf{95.4} & \textbf{87.7} & \textbf{94.9} & 84.4 & \cellcolor[gray]{.9} \textbf{93.2} & \cellcolor[gray]{.9} \textbf{77.0} & 45.3 & 41.5 & (13.1) & 29.4 & (0.4) & 40.5 & (11.3)\\
\Xhline{3\arrayrulewidth}
\end{tabular}
}
\end{table*}
%===================================================

%
%
The top row of Figure~\ref{fig:properties} shows the trade-off between misleading seen (unseen) classifiers with and without defenses. The proposed attack outperforms the other attacks in these two plots (i.e.~{is} near{er} to the top-right corner of the plot).
The second row of Figure~\ref{fig:properties} shows the relation{ship} between image quality and undetectability. We can observe that all attacks obtain similar image quality, whereas attacks can be divided into two groups {for the undetectability}, {those below} 40\% (CW, DeepFool,  R-FGSM, L-FGSM, P-FGSM, EOT, E-FGSM, and DI-FGSM targeted), and those above 40\% (JSMA, SparseFool,  U-FGSM, and DI-FGSM untargeted). The proposed attack, RP-FGSM, in both targeted and untargeted versions, is within the group of less detectable attacks.

Table~\ref{tab:All-in-one} summarizes all performance measures.

%====================================================================
\begin{table*}[t]
\caption{Example of adversarial attacks and top-5 predictions with different classifiers. Attacks shown are the best-performing with respect to the set of privacy-preserving performance measures.
{CW, DeepFool, SparseFool, U-FGSM, P-FGSM, and EOT} use ResNet50; {DI-FGSM, E-FGSM} and RP-FGSM use a combination of ResNet18, ResNet50 and AlexNet; DI-FGSM and RP-FGSM are presented with their untargeted versions.
First and sixth column: original or adversarial image; other columns: top-5 predicted classes and corresponding probability value; blue shading: true class; ${\scriptsize \bigtriangleup}$: probability higher than 99.95; ${\scriptsize \bigtriangledown}$: probability lower than 0.05.}
\label{fig:adv_imgs}
\centering
\setlength\tabcolsep{1.5pt}
\resizebox{\textwidth}{!}{%
\begin{tabular}{llrlrlrlr|llrlrlrlr}
\Xhline{3\arrayrulewidth}
& \multicolumn{2}{c}{ResNet18} & \multicolumn{2}{c}{ResNet50} & \multicolumn{2}{c}{AlexNet} & \multicolumn{2}{c|}{DenseNet161} &  & \multicolumn{2}{c}{ResNet18} & \multicolumn{2}{c}{ResNet50} & \multicolumn{2}{c}{AlexNet} & \multicolumn{2}{c}{DenseNet161} \\ \hline
\multicolumn{1}{c}{Original} &  &  &  &  &  &  &  &  & \multicolumn{1}{c}{P-FGSM} &  &  &  &  &  &  &  &  \\
\multirow{5}{*}{\includegraphics[width=0.085\textwidth]{tikz/image/O/Places365_val_00024874.png}} & \cellcolor{blue!20}jacuzzi & 87.2 & \cellcolor{blue!20}jacuzzi & 87.4 & \cellcolor{blue!20}jacuzzi & 83.5 & \cellcolor{blue!20}jacuzzi & 90.4 & \multirow{5}{*}{\includegraphics[width=0.085\textwidth]{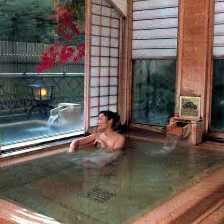}} & \cellcolor{blue!20}jacuzzi & 77.3 & garage ind. & $^{\scriptscriptstyle\bigtriangleup}100.0$ & \cellcolor{blue!20}jacuzzi & 62.2 & \cellcolor{blue!20}jacuzzi & 77.2 \\
 & swim pool ind. & 7.1 & swim pool ind. & 6.7 & swim pool ind. & 6.7 & hot spring & 6.4 &  & swim pool ind. & 19.3 & parking & $^{\scriptscriptstyle\bigtriangledown}0.0$ & fishpond & 6.2 & swim pool ind. & 19.3 \\
 & hot spring & 2.5 & hot spring & 3.5 & patio & 1.5 & swim pool ind. & 3.0 &  & fountain & 0.5 & garage out. & $^{\scriptscriptstyle\bigtriangledown}0.0$ & patio & 5.4 & lobby & 0.6 \\
 & fountain & 0.9 & sauna & 0.6 & hot spring & 1.2 & water park & 0.1 &  & hot spring & 0.5 & bus station & $^{\scriptscriptstyle\bigtriangledown}0.0$ & swim pool ind. & 4.6 & fountain & 0.4 \\
 & water park & 0.7 & water park & 0.5 & porch & 1.1 & sauna & $^{\scriptscriptstyle\bigtriangledown}0.0$ &  & water park & 0.3 & subway st. & $^{\scriptscriptstyle\bigtriangledown}0.0$ & loading dock & 2.6 & hot spring & 0.2 \\ \hline
\multicolumn{1}{c}{CW} &  &  &  &  &  &  &  &  & \multicolumn{1}{c}{EOT} &  &  &  &  &  &  &  &  \\
\multirow{5}{*}{\includegraphics[width=0.085\textwidth]{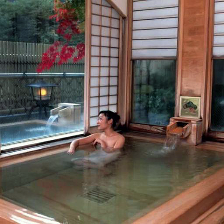}} & swim pool ind. & 52.0 & swim pool ind. & 52.0 & \cellcolor{blue!20}jacuzzi & 82.3 & \cellcolor{blue!20}jacuzzi & 86.8 & \multirow{5}{*}{\includegraphics[width=0.085\textwidth]{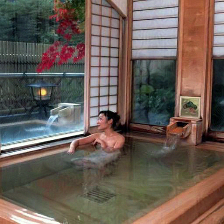}} & \cellcolor{blue!20}jacuzzi & 84.1 & greenho. ind. & 52.5 & \cellcolor{blue!20}jacuzzi & 81.8 & \cellcolor{blue!20}jacuzzi & 93.6 \\
 & \cellcolor{blue!20}jacuzzi & 31.6 & \cellcolor{blue!20}jacuzzi & 31.6 & swim pool ind. & 7.2 & hot spring & 7.8 &  & swim pool ind. & 4.9 & greenho. out. & 44.9 & swim pool ind. & 5.1 & swim pool ind. & 2.7 \\
 & hot spring & 4.7 & hot spring & 4.7 & patio & 1.5 & swim pool ind. & 5.0 &  & fountain & 3.7 & pet shop & 1.8 & fishpond & 3.2 & hot spring & 1.8 \\
 & water park & 3.4 & water park & 3.4 & hot spring & 1.2 & water park & 0.1 &  & hot spring & 1.7 & roof garden & 0.5 & patio & 1.7 & fountain & 0.3 \\
 & swim pool out. & 2.7 & swim pool out. & 2.7 & porch & 1.1 & sauna & 0.1 &  & water park & 1.7 & aquarium & 0.1 & hot spring & 1.3 & fishpond & 0.3 \\ \hline
\multicolumn{1}{c}{DeepFool} &  &  &  &  &  &  &  &  & \multicolumn{1}{c}{DI-FGSM} &  &  &  &  &  &  &  &  \\
\multirow{5}{*}{\includegraphics[width=0.085\textwidth]{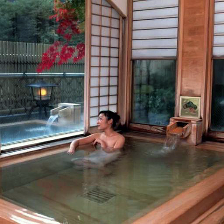}} & \cellcolor{blue!20}jacuzzi & 82.9 & hot spring & 64.4 & \cellcolor{blue!20}jacuzzi & 82.7 & \cellcolor{blue!20}jacuzzi & 83.9 & \multirow{5}{*}{\includegraphics[width=0.085\textwidth]{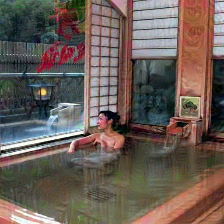}} & amus. park & 98.9 & fastfood rest. & 91.7 & lock chamb & 56.7 & rest. patio & 17.6 \\
 & swim pool ind. & 8.9 & \cellcolor{blue!20}jacuzzi & 24.0 & swim pool ind. & 6.8 & hot spring & 12.3 &  & carrousel & 0.9 & food court & 3.2 & canal & 21.9 & diner & 14.7 \\
 & hot spring & 3.7 & swim pool ind. & 4.8 & patio & 1.6 & swim pool ind. & 3.3 &  & water park & 0.2 & icecream parlor & 1.1 & boathouse & 14.9 & food court & 13.4 \\
 & water park & 1.3 & water park & 1.6 & hot spring & 1.3 & water park & 0.1 &  & playground & $^{\scriptscriptstyle\bigtriangledown}0.0$ & coffee shop & 0.9 & lake & 1.4 & amus. park & 8.9 \\
 & fountain & 1.1 & sauna & 1.0 & porch & 1.1 & sauna & 0.1 &  & ticket booth & $^{\scriptscriptstyle\bigtriangledown}0.0$ & cafeteria & 0.8 & bridge & 1.1 & coffee shop & 8.5 \\ \hline
\multicolumn{1}{c}{SparseFool} &  &  &  &  &  &  &  &  & \multicolumn{1}{c}{E-FGSM} &  &  &  &  &  &  &  &  \\
\multirow{5}{*}{\includegraphics[width=0.085\textwidth]{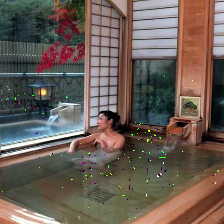}} & \cellcolor{blue!20}jacuzzi & 88.5 & hot spring & 46.1 & \cellcolor{blue!20}jacuzzi & 80.9 & \cellcolor{blue!20}jacuzzi & 85.6 & \multirow{5}{*}{\includegraphics[width=0.085\textwidth]{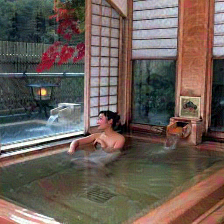}} & foot. field & 99.5 & foot. field & 99.0 & foot. field & 99.7 & \cellcolor{blue!20}jacuzzi & 74.6 \\
 & swim pool ind. & 6.6 & \cellcolor{blue!20}jacuzzi & 45.7 & swim pool ind. & 6.5 & hot spring & 11.0 &  & stadium & 0.4 & stadium foot. & 0.9 & stadium & 0.2 & swim pool ind. & 6.4 \\
 & hot spring & 2.3 & swim pool ind. & 5.2 & patio & 2.1 & swim pool ind. & 3.0 &  & soc. field & $^{\scriptscriptstyle\bigtriangledown}0.0$ & stadium base. & $^{\scriptscriptstyle\bigtriangledown}0.0$ & soc. field & 0.1 & hot spring & 5.6 \\
 & fountain & 0.9 & sauna & 0.9 & fishpond & 1.6 & water park & 0.1 &  & athl. field & $^{\scriptscriptstyle\bigtriangledown}0.0$ & athl. field & $^{\scriptscriptstyle\bigtriangledown}0.0$ & basket. court & $^{\scriptscriptstyle\bigtriangledown}0.0$ & fountain & 5.2 \\
 & water park & 0.8 & water park & 0.5 & hot spring & 1.3 & sauna & 0.1 &  & stadium & $^{\scriptscriptstyle\bigtriangledown}0.0$ & soc. field & $^{\scriptscriptstyle\bigtriangledown}0.0$ & stadium & $^{\scriptscriptstyle\bigtriangledown}0.0$ & water park & 1.6 \\ \hline
\multicolumn{1}{c}{U-FGSM} &  &  &  &  &  &  &  &  & \multicolumn{1}{c}{RP-FGSM} &  &  &  &  &  &  &  &  \\
\multirow{5}{*}{\includegraphics[width=0.085\textwidth]{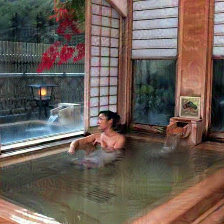}} & hot spring & 88.3 & hot spring & $^{\scriptscriptstyle\bigtriangleup}100.0$ & \cellcolor{blue!20}jacuzzi & 83.5 & hot spring & 95.8 & \multirow{5}{*}{\includegraphics[width=0.085\textwidth]{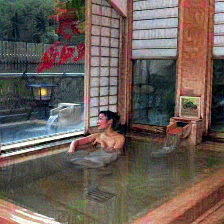}} & boathouse & 96.0 & fastfood rest. & 40.8 & lock chamb & 99.9 & amus. park & 27.3 \\
 & \cellcolor{blue!20}jacuzzi & 5.2 & volcano & $^{\scriptscriptstyle\bigtriangledown}0.0$ & swim pool ind. & 6.7 & \cellcolor{blue!20}jacuzzi & 3.3 &  & lock chamb & 3.3 & icecream parlor & 12.7 & boathouse & $^{\scriptscriptstyle\bigtriangledown}0.0$ & boathouse & 9.7 \\
 & water park & 3.1 & fountain & $^{\scriptscriptstyle\bigtriangledown}0.0$ & patio & 1.5 & swim pool ind. & 0.5 &  & pier & 0.2 & coffee shop & 12.6 & bridge & $^{\scriptscriptstyle\bigtriangledown}0.0$ & lock chamb & 7.2 \\
 & swim pool ind. & 1.3 & waterfall & $^{\scriptscriptstyle\bigtriangledown}0.0$ & hot spring & 1.2 & fountain & 0.2 &  & harbor & 0.2 & pizzeria & 12.4 & canal urban & $^{\scriptscriptstyle\bigtriangledown}0.0$ & pet shop & 4.0 \\
 & fountain & 0.9 & mountain & $^{\scriptscriptstyle\bigtriangledown}0.0$ & porch & 1.1 & water park & 0.1 &  & canal & 0.1 & diner & 9.7 & canal natural & $^{\scriptscriptstyle\bigtriangledown}0.0$ & rail. track & 3.7 \\
 \Xhline{3\arrayrulewidth}
\end{tabular}%
}
\end{table*}
%====================================================================

%MISLEADING SEEN
For \emph{seen classifiers without defenses}, JSMA, CW, DeepFool and SparseFool  have the lowest misleading rate  (lower than 32.1\% in top-5) as they craft their perturbation towards the closest class to the predicted class {and} the resulting class of the adversarial image could be the true class that we aim to protect (see Table~\ref{tab:accuracy}). Instead, targeted (untargeted) FGSM-based attacks have higher misleading rates because they move the prediction of the adversarial image closer {to} (farther {from}) the target (predicted) class at each iteration.
The misleading rates of EOT are only between 67.4\% and 72.7\% in top-5, showing a limited ability to mislead the classifier when no transformation is applied.
The misleading rates of untargeted FGSM-based attacks{,} such as U-FGSM and DI-FGSM{,} are above 83.4\% in top-5, whereas those of RP-FGSM are above 86.9\% when attacking a single classifier.
Targeted FGSM-based attacks such as L-FGSM and P-FGSM obtain the highest misleading rates (above 95\% in top-5). Similar results are obtained by attacks that use an ensemble of classifiers (E-FGSM and DI-FGSM).  
The target (untargeted) version of RP-FGSM obtains comparable results to {the other} FGSM-based attacks, i.e.~above 97.8\% (86.9\%) in the top-5.

{For} an \emph{unseen classifier}, JSMA, CW, DeepFool, SparseFool and EOT {obtain} low misleading rates (under 21.6\% in the top-5).
{Similarly to the above results for seen classifiers}, targeted {FGSM} attacks such as R-FGSM, L-FGSM, P-FGSM and E-FGSM (attacking {a} single classifier) obtain a higher performance, with misleading rates between 15.5\% and 88.7\% in the top-5. The highest misleading rates are provided by DI-FGSM and RP-FGSM when using three classifiers for generating the attack. In their targeted versions, they obtain misleading rates of 65.0\% and 64.6\% in the top-5, respectively, while the untargeted versions reach  76.0\% and 82.6\%, respectively. 
{Adversarial attacks that combine classifiers obtain the highest misleading rates in unseen classifiers}. Targeted DI-FGSM and RP-FGSM have comparable results (0.4 percentage points difference), while untargeted RP-FGSM outperforms untargeted DI-FGSM by more than 6 percentage points, thus indicating that the random selection of classifiers and transformations is an effective strategy for privacy protection.

{For {\em seen classifiers with}} {\em defenses}, JSMA, CW, DeepFool and SparseFool have low misleading rates (less than 21.2\% in the top-5). 
Targeted attacks such as R-FGSM, L-FGSM, P-FGSM, EOT and E-FGSM (attacking one classifier) obtain much lower misleading rates compared to when no defenses are employed (19.8\% - 53.4\% in the top-5).
Also in this case, untargeted FGSM-based attacks perform the best in the top-5 (above 47.4\% for U-FGSM and above 59.3\% for DI-FGSM). RP-FGSM outperforms all attacks when three classifiers are employed for crafting the adversarial images, with misleading rates in the top-5 between 59.6\% and 88.0\% when targeted, and between 70.3\% and 91.7\% when untargeted. 
When attacking three classifiers, the misleading rate of E-FGSM drops by 61.6 percentage points (35.2\% from 96.8\%) when defenses are applied. Instead, RP-FGSM untargeted only drops by 3.3 percentage points (87.7\% from 91.0\%).

%

%===============================
\begin{table*}[!t]
\centering
\setlength\tabcolsep{0.5pt}
\caption{Impact on misleading and detectability of using a random transformation in crafting adversarial perturbations (values are the variation of the score compared to the score obtained when no transformation is used).
Green shading represents an increase in misleading rate or a decrease in detectability.
Red shading represents a decrease in misleading rate or an increase in detectability.
KEY-- Class. comb.: classifier combination; N/A: Not applicable; 
R18: ResNet18; R50: ResNet50; A: AlexNet; DN: DenseNet161; T1: top-1 misleading rate; T5: top-5 misleading rate.}
\resizebox{\textwidth}{!}{
\begin{tabular}{c|l|rr|rr|rr|rr|rr|rr|rr|rr|r||rr|rr|rr|rr|rr|rr|rr|rr|r}
 \Xhline{3\arrayrulewidth}
& & \multicolumn{17}{c||}{Targeted} & \multicolumn{17}{c}{Untargeted} \\
 \Xhline{3\arrayrulewidth}
\multirow{2}{*}{Class.} & \multicolumn{1}{c|}{\multirow{3}{*}{Classifier}} & \multicolumn{8}{c|}{Misleading~$\uparrow$} & \multicolumn{8}{c|}{Misleading with defense~$\uparrow$} & \multirow{3}{*}{Det.~$\downarrow$} & \multicolumn{8}{c|}{Misleading~$\uparrow$} & \multicolumn{8}{c|}{Misleading with defense~$\uparrow$} & \multirow{3}{*}{Det.~$\downarrow$} \\ \cline{3-18}
 \cline{20-35}
\multirow{2}{*}{comb.} &  & \multicolumn{2}{c|}{R18} & \multicolumn{2}{c|}{R50} & \multicolumn{2}{c|}{A} & \multicolumn{2}{c|}{DN} & \multicolumn{2}{c|}{R18} & \multicolumn{2}{c|}{R50} & \multicolumn{2}{c|}{A} & \multicolumn{2}{c|}{DN} &  
& \multicolumn{2}{c|}{R18} & \multicolumn{2}{c|}{R50} & \multicolumn{2}{c|}{A} & \multicolumn{2}{c|}{DN} & \multicolumn{2}{c|}{R18} & \multicolumn{2}{c|}{R50} & \multicolumn{2}{c|}{A} & \multicolumn{2}{c|}{DN} &  \\
  & & T1 & T5 & T1 & T5 & T1 & T5 & T1 & T5  & T1 & T5 & T1 & T5 & T1 & T5 & T1 & T5  & & T1 & T5 & T1 & T5 & T1 & T5 & T1 & T5 & T1 & T5 & T1 & T5 & T1 & T5 & T1 & T5 & \\ 
 \Xhline{3\arrayrulewidth}
\multirow{4}{*}{N/A } & R18 & 0.0 &\cellcolor{green!20} 0.1 &\cellcolor{green!20} 2.6 &\cellcolor{green!20} 2.7 &\cellcolor{green!20} 1.4 &\cellcolor{green!20} 2.6 &\cellcolor{green!20} 2.1 &\cellcolor{green!20} 2.6 &\cellcolor{red!20} -0.3 &\cellcolor{red!20} -3.3 &\cellcolor{red!20} -0.7 &\cellcolor{red!20} -2.2 &\cellcolor{red!20} -2.1 &\cellcolor{red!20} -0.5 &\cellcolor{red!20} -1.3 &\cellcolor{red!20} -0.9 &\cellcolor{green!20} -23.0 
& \cellcolor{green!20} 1.9 &\cellcolor{green!20} 2.2 &\cellcolor{green!20} 1.5 &\cellcolor{green!20} 6.8 &\cellcolor{green!20} 2.5 &\cellcolor{green!20} 3.5 &\cellcolor{green!20} 1.1 &\cellcolor{green!20} 5.6 &\cellcolor{green!20} 26.2 &\cellcolor{green!20} 52.7 &\cellcolor{red!20} -5.3 &\cellcolor{green!20} 23.4 &\cellcolor{green!20} 4.9 &\cellcolor{green!20} 4.9 &\cellcolor{green!20} 13.7 &\cellcolor{green!20} 16.1 &\cellcolor{green!20} -7.4 \\
& R50  & \cellcolor{green!20} 4.6 &\cellcolor{green!20} 5.6 &0.0 &\cellcolor{green!20} 0.1 &\cellcolor{green!20} 2.5 &\cellcolor{green!20} 2.1 &\cellcolor{green!20} 4.2 &\cellcolor{green!20} 4.2 &\cellcolor{green!20} 6.6 &\cellcolor{green!20} 5.7 &\cellcolor{green!20} 22.5 &\cellcolor{green!20} 30.2 &\cellcolor{green!20} 2.1 &\cellcolor{green!20} 3.3 &\cellcolor{green!20} 7.6 &\cellcolor{green!20} 6.7 &\cellcolor{green!20} -17.8 
& \cellcolor{green!20} 2.1 &\cellcolor{green!20} 9.0 &\cellcolor{green!20} 1.3 &\cellcolor{green!20} 2.4 &\cellcolor{green!20} 2.6 &\cellcolor{green!20} 4.1 &\cellcolor{green!20} 2.2 &\cellcolor{green!20} 8.3 &\cellcolor{green!20} 10.9 &\cellcolor{green!20} 19.0 &\cellcolor{green!20} 11.8 &\cellcolor{green!20} 36.4 &\cellcolor{green!20} 4.2 &\cellcolor{green!20} 5.0 &\cellcolor{green!20} 14.5 &\cellcolor{green!20} 20.3 &\cellcolor{green!20} -8.8 \\
& A  & \cellcolor{green!20} 1.1 &\cellcolor{green!20} 0.5 &\cellcolor{red!20} -0.2 &\cellcolor{green!20} 0.1 &0.0 &0.0 &\cellcolor{red!20} -0.3 &\cellcolor{red!20} -0.1 &\cellcolor{green!20} 1.8 &\cellcolor{green!20} 2.1 &\cellcolor{green!20} 1.2 &\cellcolor{green!20} 0.8 &\cellcolor{green!20} 14.7 &\cellcolor{green!20} 24.4 &\cellcolor{green!20} 1.1 &\cellcolor{green!20} 0.2 &\cellcolor{green!20} -17.3  
& \cellcolor{red!20} -4.9 &\cellcolor{red!20} -5.8 &\cellcolor{red!20} -7.0 &\cellcolor{red!20} -6.2 &\cellcolor{green!20} 1.8 &\cellcolor{green!20} 2.3 &\cellcolor{red!20} -6.5 &\cellcolor{red!20} -5.9 &\cellcolor{red!20} -0.4 &\cellcolor{green!20} 1.5 &\cellcolor{red!20} -1.0 &\cellcolor{red!20} -1.4 &\cellcolor{green!20} 3.5 &\cellcolor{green!20} 13.0 &\cellcolor{red!20} -1.5 &\cellcolor{red!20} -1.4 &\cellcolor{green!20} -0.2 
\tabularnewline 
\cline{2-36}
& \multicolumn{1}{l|}{Average} &
\cellcolor{green!20} 1.9 & \cellcolor{green!20}2.1 &\cellcolor{green!20}0.8 &\cellcolor{green!20}1.0& \cellcolor{green!20}1.3& \cellcolor{green!20}1.6& \cellcolor{green!20}2.0& \cellcolor{green!20}2.2& \cellcolor{green!20}2.7& \cellcolor{green!20}1.5&\cellcolor{green!20} 7.7&\cellcolor{green!20} 9.6&\cellcolor{green!20} 4.9&\cellcolor{green!20} 9.1&\cellcolor{green!20} 2.5&\cellcolor{green!20} 2.0&\cellcolor{green!20} -19.4&\cellcolor{red!20}
-0.3 & \cellcolor{green!20}1.8 & \cellcolor{red!20}-1.4 & \cellcolor{green!20}1.0& \cellcolor{green!20}2.3& \cellcolor{green!20}3.3& \cellcolor{red!20}-1.1& \cellcolor{green!20}2.7& \cellcolor{green!20}12.2& \cellcolor{green!20}24.4&\cellcolor{green!20} 1.8&\cellcolor{green!20} 19.5&\cellcolor{green!20} 4.2&\cellcolor{green!20} 7.6&\cellcolor{green!20} 8.9&\cellcolor{green!20} 11.7& \cellcolor{green!20}-5.5
%\Xhline{3\arrayrulewidth}
\\ \hline \hline  
%\Xcline{3\arrayrulewidth}{3-1}
%\Xhline{3\arrayrulewidth}
\multirow{5}{*}{Ensemble} & \multirow{1}{*}{R18+R50}  & 0.0 &\cellcolor{red!20} -0.4 &0.0 &\cellcolor{red!20} -0.7 &\cellcolor{green!20} 1.7 &\cellcolor{green!20} 2.1 &\cellcolor{green!20} 3.4 &\cellcolor{green!20} 4.9 &\cellcolor{green!20} 18.6 &\cellcolor{green!20} 29.2 &\cellcolor{green!20} 10.5 &\cellcolor{green!20} 16.5 &\cellcolor{green!20} 3.3 &\cellcolor{green!20} 2.4 &\cellcolor{green!20} 10.3 &\cellcolor{green!20} 12.7 &\cellcolor{green!20} -22.7 
& \cellcolor{green!20} 0.6 &\cellcolor{green!20} 1.3 &\cellcolor{green!20} 0.8 &\cellcolor{green!20} 1.5 &\cellcolor{green!20} 2.3 &\cellcolor{green!20} 5.8 &\cellcolor{green!20} 0.4 &\cellcolor{green!20} 1.3 &\cellcolor{green!20} 2.9 &\cellcolor{green!20} 12.8 &\cellcolor{green!20} 5.0 &\cellcolor{green!20} 19.0 &\cellcolor{green!20} 5.6 &\cellcolor{green!20} 10.4 &\cellcolor{green!20} 7.9 &\cellcolor{green!20} 18.1 &\cellcolor{red!20} 9.0 \\
 &  \multirow{1}{*}{R18+A}  & 0.0 &0.0 &\cellcolor{green!20} 1.1 &\cellcolor{green!20} 2.6 &0.0 &\cellcolor{red!20} -0.6 &\cellcolor{green!20} 1.5 &\cellcolor{green!20} 2.3 &\cellcolor{green!20} 19.1 &\cellcolor{green!20} 28.5 &\cellcolor{green!20} 5.6 &\cellcolor{green!20} 6.2 &\cellcolor{green!20} 8.5 &\cellcolor{green!20} 18.2 &\cellcolor{green!20} 28.1 &\cellcolor{red!20} -0.8 &\cellcolor{green!20} -18.7 
& \cellcolor{green!20} 1.0 &\cellcolor{green!20} 1.7 &\cellcolor{green!20} 0.5 &\cellcolor{green!20} 3.4 &\cellcolor{green!20} 0.9 &\cellcolor{green!20} 2.5 &\cellcolor{green!20} 0.5 &\cellcolor{green!20} 2.7 &\cellcolor{green!20} 3.6 &\cellcolor{green!20} 14.0 &\cellcolor{green!20} 8.9 &\cellcolor{green!20} 15.1 &\cellcolor{green!20} 3.7 &\cellcolor{green!20} 12.0 &\cellcolor{green!20} 8.6 &\cellcolor{green!20} 15.2 &\cellcolor{red!20} 8.7 \\
 & \multirow{1}{*}{R50+A}  & \cellcolor{green!20} 4.2 &\cellcolor{green!20} 5.3 &\cellcolor{red!20} -0.1 &\cellcolor{red!20} -0.9 &0.0 &\cellcolor{red!20} -1.0 &\cellcolor{green!20} 1.9 &\cellcolor{green!20} 3.5 &\cellcolor{green!20} 8.8 &\cellcolor{green!20} 10.0 &\cellcolor{green!20} 19.1 &\cellcolor{green!20} 26.2 &\cellcolor{green!20} 11.5 &\cellcolor{green!20} 17.5 &\cellcolor{green!20} 7.5 &\cellcolor{red!20} -18.8 &\cellcolor{green!20} -8.8 
& \cellcolor{green!20} 2.2 &\cellcolor{green!20} 4.3 &\cellcolor{green!20} 1.3 &\cellcolor{green!20} 2.3 &\cellcolor{green!20} 1.0 &\cellcolor{green!20} 2.5 &\cellcolor{green!20} 1.5 &\cellcolor{green!20} 2.8 &\cellcolor{green!20} 8.0 &\cellcolor{green!20} 14.8 &\cellcolor{green!20} 7.0 &\cellcolor{green!20} 20.4 &\cellcolor{green!20} 3.9 &\cellcolor{green!20} 12.0 &\cellcolor{green!20} 8.9 &\cellcolor{green!20} 15.2 &\cellcolor{red!20} 10.1 \\
 &  \multirow{1}{*}{R18+R50+A}  & 0.0 &\cellcolor{red!20} -0.7 &\cellcolor{red!20} -0.1 &\cellcolor{red!20} -1.0 &\cellcolor{red!20} -0.1 &\cellcolor{red!20} -1.7 &\cellcolor{green!20} 3.1 &\cellcolor{green!20} 3.6 &\cellcolor{green!20} 24.4 &\cellcolor{green!20} 35.9 &\cellcolor{green!20} 12.0 &\cellcolor{green!20} 18.4 &\cellcolor{green!20} 9.4 &\cellcolor{green!20} 13.7 &\cellcolor{green!20} 10.3 &\cellcolor{green!20} 13.9 &\cellcolor{green!20} -12.6 
& \cellcolor{green!20} 1.9 &\cellcolor{green!20} 4.3 &\cellcolor{green!20} 2.2 &\cellcolor{green!20} 4.6 &\cellcolor{green!20} 1.6 &\cellcolor{green!20} 5.2 &\cellcolor{green!20} 2.8 &\cellcolor{green!20} 8.1 &\cellcolor{green!20} 4.7 &\cellcolor{green!20} 17.7 &\cellcolor{green!20} 8.1 &\cellcolor{green!20} 24.9 &\cellcolor{green!20} 6.6 &\cellcolor{green!20} 17.8 &\cellcolor{green!20} 10.9 &\cellcolor{green!20} 27.5 &\cellcolor{red!20} 9.8 \\
\cline{2-36}
 & \multicolumn{1}{l|}{Average} &
 \cellcolor{green!20}1.1 &\cellcolor{green!20} 1.1 &\cellcolor{green!20} 0.2 &\cellcolor{green!20} 0.0 &\cellcolor{green!20} 0.4 &\cellcolor{red!20} -0.3&\cellcolor{green!20} 2.5&\cellcolor{green!20} 3.6&\cellcolor{green!20} 17.7&\cellcolor{green!20} 25.9&\cellcolor{green!20} 11.8&\cellcolor{green!20} 16.8&\cellcolor{green!20} 8.2&\cellcolor{green!20} 13.0&\cellcolor{green!20} 14.1&\cellcolor{green!20} 1.8&\cellcolor{green!20} -15.7 &\cellcolor{green!20}1.7 &\cellcolor{green!20}3.5
 &\cellcolor{green!20} 1.4
 &\cellcolor{green!20} 3.4&\cellcolor{green!20} 1.2&\cellcolor{green!20} 3.4&\cellcolor{green!20} 1.6&\cellcolor{green!20} 4.5&\cellcolor{green!20} 5.4&\cellcolor{green!20} 15.5&\cellcolor{green!20} 8.0&\cellcolor{green!20} 20.1&\cellcolor{green!20} 4.7&\cellcolor{green!20} 13.9&\cellcolor{green!20} 9.5&\cellcolor{green!20} 19.3&\cellcolor{red!20}9.5
%\Xhline{3\arrayrulewidth}
\\ \hline \hline  
\multirow{5}{*}{Random} & \multirow{1}{*}{R18+R50} & \cellcolor{red!20} -0.4 &\cellcolor{red!20} -2.1 &\cellcolor{red!20} -0.2 &\cellcolor{red!20} -0.8 &\cellcolor{green!20} 0.4 &\cellcolor{green!20} 0.7 &\cellcolor{red!20} -1.5 &\cellcolor{red!20} -0.7 &\cellcolor{green!20} 9.8 &\cellcolor{green!20} 16.5 &\cellcolor{green!20} 14.9 &\cellcolor{green!20} 18.7 &\cellcolor{green!20} 1.5 &\cellcolor{green!20} 1.8 &\cellcolor{green!20} 6.6 &\cellcolor{green!20} 5.7 &\cellcolor{green!20} -14.0  
& \cellcolor{green!20} 0.9 &\cellcolor{green!20} 1.8 &\cellcolor{green!20} 1.0 &\cellcolor{green!20} 1.6 &\cellcolor{green!20} 4.0 &\cellcolor{green!20} 6.9 &\cellcolor{green!20} 1.3 &\cellcolor{green!20} 6.8 &\cellcolor{green!20} 9.0 &\cellcolor{green!20} 27.4 &\cellcolor{green!20} 4.6 &\cellcolor{green!20} 17.8 &\cellcolor{green!20} 6.1 &\cellcolor{green!20} 9.5 &\cellcolor{green!20} 11.9 &\cellcolor{green!20} 24.1 &\cellcolor{green!20} -9.9 \\
 & \multirow{1}{*}{R18+A} & \cellcolor{red!20} -0.6 &\cellcolor{red!20} -3.8 &\cellcolor{red!20} -1.9 &\cellcolor{red!20} -2.0 &\cellcolor{red!20} -0.2 &\cellcolor{green!20} 0.3 &\cellcolor{red!20} -1.5 &\cellcolor{red!20} -0.9 &\cellcolor{green!20} 8.9 &\cellcolor{green!20} 11.0 &\cellcolor{green!20} 2.6 &\cellcolor{green!20} 2.5 &\cellcolor{green!20} 11.5 &\cellcolor{green!20} 17.6 &\cellcolor{green!20} 2.4 &\cellcolor{green!20} 1.9 &\cellcolor{green!20} -5.8  
& \cellcolor{green!20} 1.1 &\cellcolor{green!20} 2.0 &\cellcolor{green!20} 2.1 &\cellcolor{green!20} 5.1 &\cellcolor{green!20} 1.2 &\cellcolor{green!20} 1.1 &\cellcolor{green!20} 1.9 &\cellcolor{green!20} 5.3 &\cellcolor{green!20} 4.9 &\cellcolor{green!20} 15.3 &\cellcolor{green!20} 7.3 &\cellcolor{green!20} 11.2 &\cellcolor{green!20} 4.3 &\cellcolor{green!20} 13.2 &\cellcolor{green!20} 7.2 &\cellcolor{green!20} 11.9 &\cellcolor{green!20} -2.8 \\

 &\multirow{1}{*}{R50+A} & \cellcolor{red!20} -7.0 &\cellcolor{red!20} -8.2 &\cellcolor{red!20} -2.8 &\cellcolor{red!20} -9.4 &\cellcolor{red!20} -0.0 &\cellcolor{green!20} 0.1 &\cellcolor{red!20} -5.8 &\cellcolor{red!20} -16.1 &\cellcolor{green!20} 3.5 &\cellcolor{green!20} 5.5 &\cellcolor{green!20} 7.4 &\cellcolor{green!20} 7.7 &\cellcolor{green!20} 11.7 &\cellcolor{green!20} 17.9 &\cellcolor{green!20} 2.5 &\cellcolor{green!20} 2.9 &\cellcolor{red!20} 0.1  
& \cellcolor{green!20} 2.5 &\cellcolor{green!20} 6.9 &\cellcolor{green!20} 1.1 &\cellcolor{green!20} 1.6 &\cellcolor{green!20} 1.0 &\cellcolor{green!20} 0.8 &\cellcolor{green!20} 2.8 &\cellcolor{green!20} 8.2 &\cellcolor{green!20} 7.7 &\cellcolor{green!20} 12.9 &\cellcolor{green!20} 8.2 &\cellcolor{green!20} 18.4 &\cellcolor{green!20} 3.8 &\cellcolor{green!20} 15.8 &\cellcolor{green!20} 8.3 &\cellcolor{green!20} 11.6 &\cellcolor{green!20} -2.9 \\

 & \multirow{1}{*}{R18+R50+A}  & \cellcolor{red!20} -0.3 &\cellcolor{red!20} -2.3 &\cellcolor{red!20} -2.3 &\cellcolor{red!20} -8.3 &\cellcolor{red!20} -1.3 &\cellcolor{red!20} -2.3 &\cellcolor{red!20} -2.9 &\cellcolor{red!20} -2.9 &\cellcolor{green!20} 1.9 &\cellcolor{green!20} 3.7 &\cellcolor{green!20} 19.3 &\cellcolor{green!20} 27.8 &\cellcolor{green!20} 9.5 &\cellcolor{green!20} 11.3 &\cellcolor{green!20} 4.0 &\cellcolor{green!20} 5.0 &\cellcolor{green!20} -7.9 
& \cellcolor{green!20} 2.4 &\cellcolor{green!20} 4.5 &\cellcolor{green!20} 2.1 &\cellcolor{green!20} 3.8 &\cellcolor{red!20} -0.4 &\cellcolor{red!20} -0.6 &\cellcolor{green!20} 2.5 &\cellcolor{green!20} 11.4 &\cellcolor{green!20} 5.1 &\cellcolor{green!20} 27.1 &\cellcolor{green!20} 8.8 &\cellcolor{green!20} 31.6 &\cellcolor{green!20} 3.1 &\cellcolor{green!20} 13.9 &\cellcolor{green!20} 12.4 &\cellcolor{green!20} 30.7 &\cellcolor{green!20} -3.1 \\
\cline{2-36}
 & \multicolumn{1}{l|}{Average} 
%\Xhline{3\arrayrulewidth}
&\cellcolor{red!20}
-2.1&\cellcolor{red!20}	-4.1&\cellcolor{red!20}	-1.8&\cellcolor{red!20}	-5.1&\cellcolor{red!20}	-0.3&\cellcolor{red!20}	-0.3&\cellcolor{red!20}	-2.9&\cellcolor{red!20}	-5.2&\cellcolor{green!20}	6.0	&\cellcolor{green!20}9.2	&\cellcolor{green!20}11.1&\cellcolor{green!20}	14.2&\cellcolor{green!20} 8.6&\cellcolor{green!20}12.2& \cellcolor{green!20}3.9 & \cellcolor{green!20} 3.9	& \cellcolor{green!20}-6.9
& \cellcolor{green!20}2.0& \cellcolor{green!20}	4.5& \cellcolor{green!20}	1.8& \cellcolor{green!20}	3.5& \cellcolor{green!20}	0.6& \cellcolor{green!20}	0.4& \cellcolor{green!20}	2.4& \cellcolor{green!20}	8.3& \cellcolor{green!20}	5.9& \cellcolor{green!20}	18.4& \cellcolor{green!20}	8.1& \cellcolor{green!20}	20.4& \cellcolor{green!20}	3.7& \cellcolor{green!20}	14.3& \cellcolor{green!20}	9.3	& \cellcolor{green!20}18.1& \cellcolor{green!20}	-2.9
\\
\Xhline{3\arrayrulewidth}
%\Xhline{3\arrayrulewidth} \multicolumn{2}{l|}{Overall change}  & \cellcolor{green!20} 0.1 & \cellcolor{red!20} -0.5 &\cellcolor{red!20} -0.4 &\cellcolor{red!20} -1.6 & \cellcolor{green!20} 0.4 & \cellcolor{green!20} 0.2 & \cellcolor{green!20} 0.4 & 0.0 & \cellcolor{green!20} 9.4 &\cellcolor{green!20} 13.2 & \cellcolor{green!20} 10.4 & \cellcolor{green!20} 13.9 &\cellcolor{green!20} 7.4 & \cellcolor{green!20} 11.6 & \cellcolor{green!20} 7.2 & \cellcolor{green!20} 2.6 &\cellcolor{green!20} -13.5& \cellcolor{green!20} 1.1 & \cellcolor{green!20} 2.9 & \cellcolor{green!20} 0.6 &\cellcolor{green!20} 2.4 & \cellcolor{green!20} 1.7 &\cellcolor{green!20} 3.1 & \cellcolor{green!20} 1.0 &\cellcolor{green!20} 5.0 &\cellcolor{green!20} 7.5 & \cellcolor{green!20} 19.6 & \cellcolor{green!20} 5.8 & \cellcolor{green!20}19.7 &\cellcolor{green!20} 4.5 & \cellcolor{green!20} 11.6& \cellcolor{green!20} 9.3&\cellcolor{green!20} 17.2 &\cellcolor{red!20} 0.2\\  \Xhline{3\arrayrulewidth}
\end{tabular}}
\label{tab:ablationTransf}
\end{table*}  
%===========================================

%MISLEADING W DEFENSE UNSEEN
When evaluating \emph{misleading unseen classifiers with defenses}, JSMA, CW, DeepFool and SparseFool still obtain the lowest misleading rates (lower than 21.2\% in top-5). Similarly, FGSM-based attacks such as R-FGSM, L-FGSM, P-FGSM, EOT and E-FGSM obtain very {low} misleading rates (under 27.1\% in the top-5){, while} 
DI-FGSM, when attacking three classifiers, obtains higher misleading rates with values of 40.1\% (targeted) and 53.7\% (untargeted).
Also in this case, the proposed RP-FGSM outperforms all other attacks, obtaining a misleading rate in the top-5 of 55.1\% (targeted) and of 77.0\% (untargeted).
This confirms that untargeted attacks exhibit a higher degree of privacy protection with unseen classifiers, both with and without defenses.

Table~\ref{fig:adv_imgs} compares predicted probabilities of sample adversarial images. 
We observe that most of the adversarial attacks are unsuitable for privacy protection as the true class might be within the top-5.
DI-FGSM and RP-FGSM are the only attacks capable of removing the true class {from} the top-5 for all classifiers in both seen (ResNet18, ResNet50 and AlexNet) and unseen (DenseNet161) settings. {This is mainly due to the fact that the attacks do not overfit the perturbation{s} to a specific classifier and that the crafted perturbations generate adversarial predicted classes that are far {from the predicted class} in the decision space.}

Targeted attacks are in general more detectable than their untargeted versions{,} as reaching the target class often means crafting a larger  perturbation.
R-FGSM, L-FGSM, P-FGSM, EOT and E-FGSM are highly \emph{detectable}, with values above 93.2\%, while U-FGSM obtains detection rates below 58.1\%. The detectability of targeted DI-FGSM is between 82.8\% and 92.2\%, and that of targeted RP-FGSM between 55.9\% and 79.0\% (RP-FGSM is less detectable than DI-FGSM in this setting). Untargeted DI-FGSM obtains detectability between 41.0\% and 53.2\%, while untargeted RP-FGSM between 41.7\% and 57.6\%, thus DI-FGSM is slightly less detectable than RP-FGSM.
JSMA and SparseFool have the lowest detection rates, with values under 42.1\%.

Attacks with a sparse perturbation, such as JSMA and SparseFool, have the highest {\em image quality} results (PSNR values above 38.4 dB). However, sparse perturbations are often of high magnitude (Figure~\ref{fig:noises}) and, therefore, highly noticeable (Table~\ref{fig:adv_imgs}). 
This effect is reflected {by the} {poor} BRISQUE {score}, {above 51.0 points, for} JSMA and SparseFool.
Attacks with dense perturbations obtain a lower PSNR than sparse ones, whose  accumulated variation of intensity across the whole image is smaller. 
However, dense perturbations might be more desirable for privacy protection, as they are less visible {and} their per-pixel intensity variation is lower than that of sparse attacks. 
FGSM-based attacks obtain consistently better BRISQUE scores than non-FGSM based attacks. FGSM-based attacks obtain around 44 BRISQUE score and non-FGSM based attacks obtain around 5 points more (the lower the better).
EOT, which has an image quality regularization in its modeling {(see Eq.~\ref{eq:eot})}, is the FGSM-based attack with the highest PSNR results (above 42.3~dB) and the lowest MAD score (below 3.6 points). However, its BRISQUE scores are higher (i.e.~worse) than those of other FGSM-based attacks, thus indicating a lower image naturalness.
When the number of classifiers {used} to create the adversarial images {increases}, {a higher level of perturbations is required, which results in a decrease of image quality}. For instance, {we see a decrease of} about 3-4~dBs when {DI-FGSM, E-FGSM and RP-FGSM employ} three classifiers instead of only one. However, {FGSM-based attacks can obtain the desired trade-off between misleading rate and image quality by tuning the  parameter~$\epsilon$}
(Eq.~\ref{eq:clipping}), a desirable feature for applications such as privacy protection (see {the} sensitivity analysis in Sec.~\ref{sec:eval:study}).
RP-FGSM obtains quality scores comparable to other FGSM-based attacks in terms of BRISQUE and PSNR. For instance, when attacking three classifiers, E-FGSM obtains BRISQUE score 41.8 and PSNR 33.8~dB, whereas RP-FGSM obtains BRISQUE score 33.4 (the lower the better) and PSNR 29.6~dB (the higher the better).
However, among the FGSM methods for three classifiers, E-FGSM obtains the best MAD score 19.8, whereas RP-FGSM obtains the worst MAD score 40.5 (the lower the better).

%%%%%%%%%%%%%%%%%%%%%%%%%%%
\subsection{Ablation and sensitivity analysis}
\label{sec:eval:study}

We present two analyses of {RP-FGSM}, namely an ablation study to validate the effectiveness of the random selection of classifier and transformation, and of different number of iterations; and a sensitivity analysis on the parameter $\epsilon$.

The ablation study compares the effect of using a randomly selected transformation with respect to not using any transformations{,} and the effect of using randomly selected classifiers with respect to using an ensemble of classifiers {at each iteration}~\cite{tramer2017ensemble}.
Furthermore, we {evaluate} the effect of the number of iterations{, $N$,} when attacking multiple classifiers. {For a fair comparison, all these experiments follow the same sequence of transformations, classifiers and target class, whenever applicable, by fixing the random seed.} {We report the variation of performance measures, namely misleading rates with and without defense, and detectability, for both targeted and untargeted RP-FGSM.}
%
%The analysis is performed for both targeted and untargeted RP-FGSM.
%are summarized in Tables~\ref{tab:ablationT} and~\ref{tab:ablationU}, respectively.

%===========================================+
\begin{table*}[!t]
\centering
\setlength\tabcolsep{1pt}
\caption{Impact on misleading and detectability of using a randomly selected classifier in crafting adversarial perturbations (values are the variation of the score compared to the score obtained when combining classifiers as an ensemble~\cite{tramer2017ensemble}).
Green shading represents an increase in misleading rate or a decrease in detectability.
Red shading represents a decrease in misleading rate or an increase in detectability.
KEY -- Trans.: Transformation; N/A: not applicable;
R18: ResNet18; R50: ResNet50; A: AlexNet; DN: DenseNet161; T1: top-1 misleading rate; T5: top-5 misleading rate.
}
\label{tab:ablationClass_random_VS_ensemble}
\resizebox{\textwidth}{!}{
\begin{tabular}{c|l|rr|rr|rr|rr|rr|rr|rr|rr|r||rr|rr|rr|rr|rr|rr|rr|rr|r}
 \Xhline{3\arrayrulewidth}
& & \multicolumn{17}{c||}{Targeted} & \multicolumn{17}{c}{Untargeted} \\
 \Xhline{3\arrayrulewidth}
 \multirow{3}{*}{Trans.}& \multicolumn{1}{c|}{\multirow{3}{*}{Classifier}} & \multicolumn{8}{c|}{Misleading~$\uparrow$} & \multicolumn{8}{c|}{Misleading with defense~$\uparrow$} & \multirow{3}{*}{Det.~$\downarrow$} & \multicolumn{8}{c|}{Misleading~$\uparrow$} & \multicolumn{8}{c|}{Misleading with defense~$\uparrow$} & \multirow{3}{*}{Det.~$\downarrow$} \\ \cline{3-18}
 \cline{20-35}
& & \multicolumn{2}{c|}{R18} & \multicolumn{2}{c|}{R50} & \multicolumn{2}{c|}{A} & \multicolumn{2}{c|}{DN} & \multicolumn{2}{c|}{R18} & \multicolumn{2}{c|}{R50} & \multicolumn{2}{c|}{A} & \multicolumn{2}{c|}{DN} &  
& \multicolumn{2}{c|}{R18} & \multicolumn{2}{c|}{R50} & \multicolumn{2}{c|}{A} & \multicolumn{2}{c|}{DN} & \multicolumn{2}{c|}{R18} & \multicolumn{2}{c|}{R50} & \multicolumn{2}{c|}{A} & \multicolumn{2}{c|}{DN} &  \\
& & T1 & T5 & T1 & T5 & T1 & T5 & T1 & T5  & T1 & T5 & T1 & T5 & T1 & T5 & T1 & T5  & & T1 & T5 & T1 & T5 & T1 & T5 & T1 & T5 & T1 & T5 & T1 & T5 & T1 & T5 & T1 & T5 & \\ 
\Xhline{3\arrayrulewidth}
\multirow{5}{*}{N/A  } &  \multirow{1}{*}{R18+R50} & 0.0 &\cellcolor{red!20} -0.2 &0.0 &\cellcolor{red!20} -0.8 &\cellcolor{red!20} -0.5 &\cellcolor{red!20} -0.7 &\cellcolor{red!20} -2.0 &\cellcolor{red!20} -2.4 &\cellcolor{red!20} -1.2 &\cellcolor{red!20} -2.0 &\cellcolor{red!20} -10.1 &\cellcolor{red!20} -10.8 &\cellcolor{red!20} -0.3 &\cellcolor{red!20} -1.0 &\cellcolor{red!20} -1.6 &\cellcolor{red!20} -0.9 &\cellcolor{red!20} 0.6
& \cellcolor{green!20} 1.1 &\cellcolor{green!20} 1.6 &\cellcolor{green!20} 0.4 &\cellcolor{green!20} 0.8 &\cellcolor{red!20} -2.5 &\cellcolor{red!20} -3.0 &\cellcolor{red!20} -0.1 &\cellcolor{red!20} -4.3 &\cellcolor{red!20} -5.8 &\cellcolor{red!20} -17.2 &\cellcolor{green!20} 1.2 &\cellcolor{green!20} 2.7 &\cellcolor{red!20} -1.8 &\cellcolor{red!20} -1.0 &\cellcolor{red!20} -3.9 &\cellcolor{red!20} -8.0 &\cellcolor{red!20} 8.9 \\
& \multirow{1}{*}{R18+A} & 0.0 &0.0 &\cellcolor{red!20} -2.1 &\cellcolor{red!20} -1.1 &0.0 &\cellcolor{red!20} -0.6 &\cellcolor{red!20} -1.7 &\cellcolor{red!20} -1.5 &\cellcolor{red!20} -2.1 &\cellcolor{red!20} -1.9 &\cellcolor{red!20} -1.1 &\cellcolor{red!20} -1.3 &\cellcolor{red!20} -1.8 &\cellcolor{red!20} -1.9 &\cellcolor{green!20} 21.9 &\cellcolor{red!20} -7.5 &\cellcolor{red!20} 0.7
& \cellcolor{green!20} 1.3 &\cellcolor{green!20} 1.6 &\cellcolor{red!20} -0.8 &\cellcolor{red!20} -2.9 &\cellcolor{green!20} 0.2 &\cellcolor{green!20} 1.2 &\cellcolor{red!20} -1.0 &\cellcolor{red!20} -3.4 &\cellcolor{red!20} -1.5 &\cellcolor{red!20} -6.6 &\cellcolor{red!20} -2.6 &\cellcolor{red!20} -4.5 &\cellcolor{green!20} 0.4 &0.0 &\cellcolor{red!20} -2.8 &\cellcolor{red!20} -4.8 &\cellcolor{red!20} 9.8 \\
& \multirow{1}{*}{R50+A} & \cellcolor{green!20} 4.2 &\cellcolor{green!20} 5.3 &0.0 &\cellcolor{red!20} -1.1 &\cellcolor{red!20} -0.1 &\cellcolor{red!20} -0.8 &\cellcolor{green!20} 1.9 &\cellcolor{green!20} 13.0 &\cellcolor{red!20} -1.5 &\cellcolor{red!20} -1.2 &\cellcolor{red!20} -1.5 &\cellcolor{red!20} -1.2 &\cellcolor{red!20} -1.3 &\cellcolor{red!20} -2.5 &\cellcolor{red!20} -1.5 &\cellcolor{red!20} -26.9 &\cellcolor{green!20} -8.0
& \cellcolor{green!20} 0.4 &\cellcolor{red!20} -2.6 &\cellcolor{green!20} 1.7 &\cellcolor{green!20} 2.8 &\cellcolor{green!20} 0.6 &\cellcolor{green!20} 1.7 &\cellcolor{green!20} 0.2 &\cellcolor{red!20} -4.8 &\cellcolor{red!20} -2.1 &\cellcolor{red!20} -4.0 &\cellcolor{red!20} -2.8 &\cellcolor{red!20} -7.4 &\cellcolor{green!20} 0.7 &\cellcolor{green!20} 0.2 &\cellcolor{red!20} -3.4 &\cellcolor{red!20} -4.4 &\cellcolor{red!20} 9.8 \\
& \multirow{1}{*}{R18+R50+A} & 0.0 &\cellcolor{red!20} -0.2 &\cellcolor{red!20} -0.1 &\cellcolor{red!20} -0.9 &\cellcolor{red!20} -0.2 &\cellcolor{red!20} -3.1 &\cellcolor{red!20} -2.4 &\cellcolor{red!20} -3.7 &\cellcolor{green!20} 4.5 &\cellcolor{green!20} 5.0 &\cellcolor{red!20} -8.1 &\cellcolor{red!20} -10.1 &\cellcolor{red!20} -2.7 &\cellcolor{red!20} -3.8 &\cellcolor{red!20} -2.6 &\cellcolor{red!20} -2.2 &\cellcolor{red!20} 0.2
& \cellcolor{green!20} 1.2 &\cellcolor{green!20} 2.1 &\cellcolor{green!20} 0.3 &\cellcolor{green!20} 0.8 &\cellcolor{green!20} 1.8 &\cellcolor{green!20} 4.3 &\cellcolor{green!20} 0.2 &\cellcolor{red!20} -4.4 &\cellcolor{green!20} 0.7 &\cellcolor{red!20} -8.2 &\cellcolor{red!20} -1.0 &\cellcolor{red!20} -10.1 &\cellcolor{green!20} 2.8 &\cellcolor{green!20} 0.2 &\cellcolor{red!20} -2.5 &\cellcolor{red!20} -7.4 &\cellcolor{red!20} 7.0 \\
\cline{2-36}
 & \multicolumn{1}{l|}{Average} 
&\cellcolor{green!20}1.1&\cellcolor{green!20}	1.2&\cellcolor{red!20}	-0.5&\cellcolor{red!20}	-1.0 &\cellcolor{red!20}	-0.2 &\cellcolor{red!20}	-1.3 &\cellcolor{red!20}	-1.1&\cellcolor{green!20}	1.4&\cellcolor{red!20}	-0.1 &	0.0 &\cellcolor{red!20}	-5.2&\cellcolor{red!20}	-5.9&\cellcolor{red!20}		-1.5	&\cellcolor{red!20}	-2.3&\cellcolor{green!20}		4.1&\cellcolor{red!20}		-9.4&\cellcolor{green!20}		-1.6&\cellcolor{red!20}	1.0	&\cellcolor{green!20}0.7	&\cellcolor{green!20}0.4	&\cellcolor{green!20}0.4	& 0.0	&\cellcolor{green!20} 1.1 &\cellcolor{red!20}	-0.2	&\cellcolor{red!20} -4.2 &\cellcolor{red!20}	-2.2	&\cellcolor{red!20} -9.0&\cellcolor{red!20}	-1.3 &\cellcolor{red!20}	-4.8	&\cellcolor{green!20} 0.5 &\cellcolor{red!20}	-0.2 &\cellcolor{red!20}	-3.2 &\cellcolor{red!20}	-6.2 &\cellcolor{red!20}	8.9
%\Xhline{3\arrayrulewidth}
\\ \hline \hline  

\multirow{5}{*}{Random} & \multirow{1}{*}{R18+R50} & 0.0 &\cellcolor{green!20} 0.7 &0.0 &\cellcolor{green!20} 0.7 &\cellcolor{green!20} 36.6 &\cellcolor{green!20} 65.4 &\cellcolor{green!20} 8.7 &\cellcolor{green!20} 11.1 &\cellcolor{green!20} 0.6 &\cellcolor{green!20} 1.8 &\cellcolor{green!20} 7.4 &\cellcolor{green!20} 11.0 &\cellcolor{green!20} 1.3 &\cellcolor{green!20} 5.4 &\cellcolor{green!20} 9.7 &\cellcolor{green!20} 11.6 &\cellcolor{green!20} -7.9
&  \cellcolor{green!20} 1.3 &\cellcolor{green!20} 2.1 &\cellcolor{green!20} 0.6 &\cellcolor{green!20} 0.9 &\cellcolor{red!20} -0.8 &\cellcolor{red!20} -1.9 &\cellcolor{green!20} 0.8 &\cellcolor{green!20} 1.2 &\cellcolor{green!20} 0.3 &\cellcolor{red!20} -2.6 &\cellcolor{green!20} 0.8 &\cellcolor{green!20} 1.5 &\cellcolor{red!20} -1.3 &\cellcolor{red!20} -1.9 &\cellcolor{green!20} 0.1 &\cellcolor{red!20} -2.0 &\cellcolor{green!20} -10.0 \\
 & \multirow{1}{*}{R18+A} & 0.0 &\cellcolor{green!20} 0.6 &\cellcolor{green!20} 7.4 &\cellcolor{green!20} 8.6 &0.0 &\cellcolor{green!20} 0.9 &\cellcolor{green!20} 7.2 &\cellcolor{green!20} 8.8 &\cellcolor{red!20} -3.3 &\cellcolor{red!20} -5.1 &\cellcolor{green!20} 5.6 &\cellcolor{green!20} 6.2 &\cellcolor{green!20} 12.2 &\cellcolor{green!20} 19.0 &\cellcolor{green!20} 5.9 &\cellcolor{green!20} 6.8 &\cellcolor{green!20} -3.4
&  \cellcolor{green!20} 1.4 &\cellcolor{green!20} 1.9 &\cellcolor{green!20} 0.8 &\cellcolor{red!20} -1.3 &\cellcolor{green!20} 0.5 &\cellcolor{red!20} -0.2 &\cellcolor{green!20} 0.4 &\cellcolor{red!20} -0.8 &\cellcolor{red!20} -0.2 &\cellcolor{red!20} -5.3 &\cellcolor{red!20} -4.2 &\cellcolor{red!20} -8.4 &\cellcolor{green!20} 1.0 &\cellcolor{green!20} 1.2 &\cellcolor{red!20} -4.2 &\cellcolor{red!20} -8.1 &\cellcolor{green!20} -1.7 \\
 & \multirow{1}{*}{R50+A} &  \cellcolor{green!20} 6.7 &\cellcolor{green!20} 10.0 &\cellcolor{green!20} 0.1 &\cellcolor{green!20} 1.4 &0.0 &\cellcolor{green!20} 0.5 &\cellcolor{green!20} 8.7 &\cellcolor{green!20} 10.1 &\cellcolor{green!20} 5.8 &\cellcolor{green!20} 8.3 &\cellcolor{red!20} -3.3 &\cellcolor{red!20} -5.0 &\cellcolor{green!20} 8.9 &\cellcolor{green!20} 18.8 &\cellcolor{green!20} 5.1 &\cellcolor{green!20} 7.2 &\cellcolor{green!20} -9.8
&  \cellcolor{green!20} 0.6 &0.0 &\cellcolor{green!20} 1.5 &\cellcolor{green!20} 2.2 &\cellcolor{green!20} 0.5 &\cellcolor{green!20} 0.1 &\cellcolor{green!20} 1.5 &\cellcolor{green!20} 0.6 &\cellcolor{red!20} -2.4 &\cellcolor{red!20} -5.9 &\cellcolor{red!20} -1.6 &\cellcolor{red!20} -9.4 &\cellcolor{green!20} 0.6 &\cellcolor{green!20} 4.0 &\cellcolor{red!20} -4.0 &\cellcolor{red!20} -8.0 &\cellcolor{green!20} -3.2 \\
 &\multirow{1}{*}{R18+R50+A} & 0.0 &\cellcolor{green!20} 1.0 &\cellcolor{green!20} 0.1 &\cellcolor{green!20} 1.6 &0.0 &\cellcolor{green!20} 1.1 &\cellcolor{green!20} 11.4 &\cellcolor{green!20} 16.4 &\cellcolor{green!20} 3.9 &\cellcolor{green!20} 9.6 &\cellcolor{green!20} 8.5 &\cellcolor{green!20} 19.1 &\cellcolor{green!20} 6.2 &\cellcolor{green!20} 12.0 &\cellcolor{green!20} 8.9 &\cellcolor{green!20} 12.4 &\cellcolor{green!20} -28.7
&  \cellcolor{green!20} 1.7 &\cellcolor{green!20} 2.2 &\cellcolor{green!20} 0.1 &0.0 &\cellcolor{red!20} -0.2 &\cellcolor{red!20} -1.5 &0.0 &\cellcolor{red!20} -1.1 &\cellcolor{green!20} 1.1 &\cellcolor{green!20} 1.2 &\cellcolor{red!20} -0.3 &\cellcolor{red!20} -3.4 &\cellcolor{red!20} -0.7 &\cellcolor{red!20} -3.7 &\cellcolor{red!20} -1.0 &\cellcolor{red!20} -4.2 &\cellcolor{green!20} -5.9 

\\
\cline{2-36}
 & \multicolumn{1}{l|}{Average} 
&\cellcolor{green!20}1.7	&\cellcolor{green!20}3.1&\cellcolor{green!20}	1.9&\cellcolor{green!20}	3.1&\cellcolor{green!20}	9.2	&\cellcolor{green!20} 17.0	&\cellcolor{green!20} 9.0 &\cellcolor{green!20}	11.6	&\cellcolor{green!20}1.8	&\cellcolor{green!20}3.7&\cellcolor{green!20}	4.6&\cellcolor{green!20}	7.8&\cellcolor{green!20}	7.2&\cellcolor{green!20}	13.8&\cellcolor{green!20}	7.4	&\cellcolor{green!20} 9.5&\cellcolor{green!20}	-12.5&\cellcolor{green!20}	1.3&\cellcolor{green!20}	1.5&\cellcolor{green!20}	0.8&\cellcolor{green!20}	0.4&\cellcolor{green!20}	0.0&\cellcolor{red!20}	-0.9	&\cellcolor{green!20}0.7	&\cellcolor{green!20} 0.0&\cellcolor{red!20}	-0.3&\cellcolor{red!20}	-3.2&\cellcolor{red!20}	-1.3&\cellcolor{red!20}	-4.9	&\cellcolor{red!20}-0.1	&\cellcolor{red!20}-0.1&\cellcolor{red!20}	-2.3&\cellcolor{red!20}	-5.6&\cellcolor{green!20}	-5.2
\\
\Xhline{3\arrayrulewidth}
\end{tabular}
}
\end{table*}
%===========================================+

%===========================================+
\begin{table*}[!t]
\centering
\setlength\tabcolsep{2pt}
\caption{
Impact on misleading and detectability of using a random selection of classifiers with varying number of iterations in crafting adversarial perturbations (values are the variation of the score compared to the score obtained when using an ensemble of classifiers~\cite{tramer2017ensemble}).
%Note that to use the classifiers the same number of times, the random selection attack needs to be performed for double (triple) iterations of the ensemble, when using two (three) classifiers.
Green shading represents an increase in misleading rate or a decrease in detectability.
Red shading represents a decrease in misleading rate or an increase in detectability.
KEY -- R18: ResNet18; R50: ResNet50; A: AlexNet; DN: DenseNet161; T1: top-1 misleading rate; T5: top-5 misleading rate.
%KEY -- \# iterations Ensemble: number of iterations when using an ensemble of classifiers at each iteration; \# iterations Random: number of iterations when randomly selecting a classifier at each iteration.
%(a) Ensemble of classifiers run for 20 iterations. Random selection of classifiers run for \# iter, which allows the proposed attack to use the classifiers the same number of times as an ensemble of classifiers.
%
%(b) Both random selection and ensemble of classifiers run for 60 iterations (using three classifiers).
%Note that in this setting, ensemble uses a classifier three times more, on average, than the random selection. 
}
\label{tab:ablationIterationNumber}
\begin{tabular}{c}
    \resizebox{0.75\textwidth}{!}
    {
      \begin{tabular}{cc|l|rr|rr|rr|rr|rr|rr|rr|rr|r}
        \Xhline{3\arrayrulewidth}
        & & & \multicolumn{8}{c|}{Misleading~$\uparrow$} & \multicolumn{8}{c|}{Misleading with defense~$\uparrow$} & \multirow{3}{*}{Det.~$\downarrow$}  \\ \cline{1-19}
        \multirow{2}{*}{\begin{tabular}[c]{@{}c@{}}\# iterations\\   Ensemble\end{tabular}} &
        \multirow{2}{*}{\begin{tabular}[c]{@{}c@{}}\# iterations\\  Random \end{tabular}} & \multirow{2}{*}{Classifier}&
        \multicolumn{2}{c|}{R18} & \multicolumn{2}{c|}{R50} & \multicolumn{2}{c|}{A} & \multicolumn{2}{c|}{DN} & \multicolumn{2}{c|}{R18} & \multicolumn{2}{c|}{R50} & \multicolumn{2}{c|}{A} & \multicolumn{2}{c|}{DN} &  \\ 
        & &    &   T1 & T5 & T1 & T5 & T1 & T5 & T1 & T5 & T1 & T5 & T1 & T5 & T1 & T5 & T1 & T5 & \\ 
        \Xhline{3\arrayrulewidth}
        \multirow{4}{*}{$20$} &
        \multirow{4}{*}{$20$}
        &   R18+R50 &\cellcolor{red!20} -0.4 &\cellcolor{red!20} -1.9 &\cellcolor{red!20} -0.2 &\cellcolor{red!20} -0.9 &\cellcolor{red!20} -1.8 &\cellcolor{red!20} -2.1 &\cellcolor{red!20} -6.9 &\cellcolor{red!20} -8.0 &\cellcolor{red!20} -10.0 &\cellcolor{red!20} -14.7 &\cellcolor{red!20} -5.7 &\cellcolor{red!20} -8.6 &\cellcolor{red!20} -2.1 &\cellcolor{red!20} -1.6 &\cellcolor{red!20} -5.3 &\cellcolor{red!20} -7.9 &\cellcolor{red!20} 9.3\\
        & &   R18+A &\cellcolor{red!20} -0.6 &\cellcolor{red!20} -3.8 &\cellcolor{red!20} -5.1 &\cellcolor{red!20} -5.7 &\cellcolor{red!20} -0.2 &\cellcolor{green!20} 0.3 &\cellcolor{red!20} -4.7 &\cellcolor{red!20} -4.7 &\cellcolor{red!20} -12.3 &\cellcolor{red!20} -19.4 &\cellcolor{red!20} -4.1 &\cellcolor{red!20} -5.0 &\cellcolor{green!20} 1.2 &\cellcolor{red!20} -2.5 &\cellcolor{red!20} -3.8 &\cellcolor{red!20} -4.8 &\cellcolor{red!20} 13.6\\
        & &   R50+A &\cellcolor{red!20} -7.0 &\cellcolor{red!20} -8.2 &\cellcolor{red!20} -2.7 &\cellcolor{red!20} -9.6 &\cellcolor{red!20} -0.1 &\cellcolor{green!20} 0.3 &\cellcolor{red!20} -5.8 &\cellcolor{red!20} -6.6 &\cellcolor{red!20} -6.8 &\cellcolor{red!20} -5.7 &\cellcolor{red!20} -13.2 &\cellcolor{red!20} -19.7 &\cellcolor{red!20} -1.1 &\cellcolor{red!20} -2.1 &\cellcolor{red!20} -6.5 &\cellcolor{red!20} -5.2 &\cellcolor{red!20} 0.9\\
        & &   R18+R50+A &\cellcolor{red!20} -0.3 &\cellcolor{red!20} -1.8 &\cellcolor{red!20} -2.3 &\cellcolor{red!20} -8.2 &\cellcolor{red!20} -1.4 &\cellcolor{red!20} -3.7 &\cellcolor{red!20} -8.4 &\cellcolor{red!20} -10.2 &\cellcolor{red!20} -18.0 &\cellcolor{red!20} -27.2 &\cellcolor{red!20} -0.8 &\cellcolor{red!20} -0.7 &\cellcolor{red!20} -2.6 &\cellcolor{red!20} -6.2 &\cellcolor{red!20} -8.9 &\cellcolor{red!20} -11.1 &\cellcolor{red!20} 4.9\\ \hline
        %%%%%
        \multirow{3}{*}{$20$} &
        \multirow{3}{*}{$40$}&   R18+R50 &0.0 &\cellcolor{green!20} 0.7 &0.0 &\cellcolor{green!20} 0.7 &\cellcolor{green!20} 36.6 &\cellcolor{green!20} 65.4 &\cellcolor{green!20} 8.7 &\cellcolor{green!20} 11.1 &\cellcolor{green!20} 0.6 &\cellcolor{green!20} 1.8 &\cellcolor{green!20} 7.4 &\cellcolor{green!20} 11.0 &\cellcolor{green!20} 1.3 &\cellcolor{green!20} 5.4 &\cellcolor{green!20} 9.7 &\cellcolor{green!20} 11.6 &\cellcolor{green!20} -7.9\\
        & &   R18+A &0.0 &\cellcolor{green!20} 0.6 &\cellcolor{green!20} 7.4 &\cellcolor{green!20} 8.6 &0.0 &\cellcolor{green!20} 0.9 &\cellcolor{green!20} 7.2 &\cellcolor{green!20} 8.8 &\cellcolor{red!20} -3.3 &\cellcolor{red!20} -5.1 &\cellcolor{green!20} 5.6 &\cellcolor{green!20} 6.2 &\cellcolor{green!20} 12.2 &\cellcolor{green!20} 19.0 &\cellcolor{green!20} 5.9 &\cellcolor{green!20} 6.8 &\cellcolor{green!20} -3.4\\
        & &   R50+A & \cellcolor{green!20} 6.7 &\cellcolor{green!20} 10.0 &\cellcolor{green!20} 0.1 &\cellcolor{green!20} 1.4 &0.0 &\cellcolor{green!20} 0.5 &\cellcolor{green!20} 8.7 &\cellcolor{green!20} 10.1 &\cellcolor{green!20} 5.8 &\cellcolor{green!20} 8.3 &\cellcolor{red!20} -3.3 &\cellcolor{red!20} -5.0 &\cellcolor{green!20} 8.9 &\cellcolor{green!20} 18.8 &\cellcolor{green!20} 5.1 &\cellcolor{green!20} 7.2 &\cellcolor{green!20} -9.8\\ \hline 
        $20$ &
        $60$ & R18+R50+A &0.0 &\cellcolor{green!20} 1.0 &\cellcolor{green!20} 0.1 &\cellcolor{green!20} 1.6 &0.0 &\cellcolor{green!20} 1.1 &\cellcolor{green!20} 11.4 &\cellcolor{green!20} 16.4 &\cellcolor{green!20} 3.9 &\cellcolor{green!20} 9.6 &\cellcolor{green!20} 8.5 &\cellcolor{green!20} 19.1 &\cellcolor{green!20} 6.2 &\cellcolor{green!20} 12.0 &\cellcolor{green!20} 8.9 &\cellcolor{green!20} 12.4 &\cellcolor{green!20} -28.7\\  \hline \hline
        $60$ &
        $60$ & R18+R50+A &  0.0	&  \cellcolor{green!20} 0.2& 	0.0	& 0.0	& \cellcolor{red!20} -0.1	& \cellcolor{red!20} -0.4& \cellcolor{green!20} 6.6	& \cellcolor{green!20} 10.7& 	\cellcolor{green!20} 15.2& 	\cellcolor{green!20} 28.1	& \cellcolor{green!20} 21.2	& \cellcolor{green!20} 37.0	& \cellcolor{green!20} 10.4	& \cellcolor{green!20} 18.7	& \cellcolor{green!20}13.1	& \cellcolor{green!20} 18.3& 	\cellcolor{green!20}-28.9 \\ 
        \Xhline{3\arrayrulewidth}
        \end{tabular}
        }
      \end{tabular}
\end{table*}
%%%%%%%%%%%%%%%%%%%%%%%%%%%%%%%%%

{Table~\ref{tab:ablationTransf} shows the effect of using random transformations with respect to not using any transformation. When employing a single classifier, using transformations generally improves the misleading and decreases the detectability. In particular, the detectability decreases by an average of 19.4 (5.5) percentage points in the targeted (untargeted) version. When using an ensemble of classifiers, the misleading generally increases for both targeted and untargeted. For the untargeted version, the detectability increases by an average of 9.5 percentage points, whereas for the targeted one, increases by an average of 15.7 percentage points. We argue that using an ensemble of classifiers generates overfitted perturbations that successfully mislead the classifier, but are more detectable. When using a random classifier at each iteration, as we propose, the misleading with defense increases in both targeted and untargeted versions. Moreover, the detectability in the untargeted attack decreases, contrary to when using an ensemble of classifiers. This experiment shows that introducing random transformations improves the performance, especially when a random classifier is employed at each iteration.}

{Table~\ref{tab:ablationClass_random_VS_ensemble} compares the effect of using a randomly selected classifier with respect to using an ensemble of classifiers, at each iteration. We report the results when not using any transformation, for reference, and also with a random selection of transformations. When random transformations are employed in the generation, the use of a randomly selected classifier at each iteration increases, in general, the misleading rate in the targeted version by a maximum of 4.2 percentage points in top-1 with an unseen classifier (DenseNet) . Also, detectability decreases by 12.5 (5.2) percentage points in the targeted (untargeted) version. Some instances of misleading rate decreases are observed, especially in the untargeted version.}
%

%====================================================================
\pgfplotstableread{tikz/epsilon/misleading.txt}\epsMisleading
\pgfplotstableread{tikz/epsilon/misleading_defense.txt}\epsMisleadingDefense
\pgfplotstableread{tikz/epsilon/quality_PSNR.txt}\epsQuality
\pgfplotstableread{tikz/epsilon/detectability.txt}\epsDetectability
\pgfplotsset{compat=1.3}
\begin{figure}[!t]
 \centering
  \setlength{\tabcolsep}{0pt}
  \begin{tabular}{ll}
    \begin{tikzpicture}
    \begin{axis}[
    height=4cm,
    width=4.5cm,
    xmode=log,
    ylabel={MS},
    xmin=0.00392, xmax=0.26,
    ymin=40, ymax=100,
    label style={font=\small},
    tick label style={font=\small},
    ylabel near ticks,
    xlabel near ticks,
    xtick={0.00392,0.00784,0.01569,0.03137,0.06275,0.12549,0.25098,0.50196,1},
    xticklabels={},
    ylabel shift = -8 pt,
    ]
    \addplot+[mark options={fill=cR18, scale=0.6}, cR18] table[x=epsilon, y=T1_R18]{\epsMisleading};
    \addplot+[mark options={fill=cR50, scale=0.6}, cR50] table[x=epsilon, y=T1_R50]{\epsMisleading};
    \addplot+[mark options={fill=cA, scale=0.6}, cA] table[x=epsilon, y=T1_A]{\epsMisleading};
    \addplot+[mark options={fill=c3m, scale=0.6}, c3m] table[x=epsilon, y=T1_R18_R50_A]{\epsMisleading};
    \end{axis}
    \end{tikzpicture}
    &
    \begin{tikzpicture}
    \begin{axis}[
    height=4cm,
    width=4.5cm,
    xmode=log,
    ylabel={MU},
    xmin=0.00392, xmax=0.26,
    ymin=40, ymax=100,
    label style={font=\small},
    tick label style={font=\small},
    ylabel near ticks,
    xlabel near ticks,
    xtick={0.00392,0.00784,0.01569,0.03137,0.06275,0.12549,0.25098},
    xticklabels={},
    ylabel shift = -8 pt,
    ]
    \addplot+[mark options={fill=cR18, scale=0.6}, cR18] table[x=epsilon, y=bT1_R18]{\epsMisleading};
    \addplot+[mark options={fill=cR50, scale=0.6}, cR50] table[x=epsilon, y=bT1_R50]{\epsMisleading};
    \addplot+[mark options={fill=cA, scale=0.6}, cA] table[x=epsilon, y=bT1_A]{\epsMisleading};
    \addplot+[mark options={fill=c3m, scale=0.6}, c3m] table[x=epsilon, y=bT1_R18_R50_A]{\epsMisleading};
    \end{axis}
    \end{tikzpicture}
    \\
    \begin{tikzpicture}
    \begin{axis}[
    height=4cm,
    width=4.5cm,
    xmode=log,
    ylabel={MSwD},
    xmin=0.00392, xmax=0.26,
    ymin=40, ymax=100,
    label style={font=\small},
    tick label style={font=\small},
    ylabel near ticks,
    xlabel near ticks,
    xtick={0.00392,0.00784,0.01569,0.03137,0.06275,0.12549,0.25098,0.50196,1},
    xticklabels={},
    ylabel shift = -8 pt,
    ]
    \addplot+[mark options={fill=cR18, scale=0.6}, cR18] table[x=epsilon, y=T1_R18]{\epsMisleadingDefense};
    \addplot+[mark options={fill=cR50, scale=0.6}, cR50] table[x=epsilon, y=T1_R50]{\epsMisleadingDefense};
    \addplot+[mark options={fill=cA, scale=0.6}, cA] table[x=epsilon, y=T1_A]{\epsMisleadingDefense};
    \addplot+[mark options={fill=c3m, scale=0.6}, c3m] table[x=epsilon, y=T1_R18_R50_A]{\epsMisleadingDefense};
    \end{axis}
    \end{tikzpicture}
    &
    \begin{tikzpicture}
    \begin{axis}[
    height=4cm,
    width=4.5cm,
    xmode=log,
    ylabel={MUwD},
    xmin=0.00392, xmax=0.26,
    ymin=40, ymax=100,
    label style={font=\small},
    tick label style={font=\small},
    ylabel near ticks,
    xlabel near ticks,
    xtick={0.00392,0.00784,0.01569,0.03137,0.06275,0.12549,0.25098},
    xticklabels={},
    ylabel shift = -8 pt,
    ]
    \addplot+[mark options={fill=cR18, scale=0.6}, cR18] table[x=epsilon, y=bT1_R18]{\epsMisleadingDefense};
    \addplot+[mark options={fill=cR50, scale=0.6}, cR50] table[x=epsilon, y=bT1_R50]{\epsMisleadingDefense};
    \addplot+[mark options={fill=cA, scale=0.6}, cA] table[x=epsilon, y=bT1_A]{\epsMisleadingDefense};
    \addplot+[mark options={fill=c3m, scale=0.6}, c3m] table[x=epsilon, y=bT1_R18_R50_A]{\epsMisleadingDefense};
    \end{axis}
    \end{tikzpicture}
    \\
    \begin{tikzpicture}
    \begin{axis}[
    height=4cm,
    width=4.5cm,
    xmode=log,
    ylabel={Detectability},
    xlabel={$\epsilon$},
    xmin=1, xmax=64,
    ymin=0, ymax=100,
    label style={font=\small},
    tick label style={font=\small},
    ylabel near ticks,
    xlabel near ticks,
    xtick={1,2,4,8,16,32,64},
    xticklabels={1,2,4,8,16,32,64},
    ylabel shift = -8 pt,
    ]
    \addplot+[mark options={fill=cR18, scale=0.6}, cR18] table[x=epsilon, y=R18]{\epsDetectability};
    \addplot+[mark options={fill=cR50, scale=0.6}, cR50] table[x=epsilon, y=R50]{\epsDetectability};
    \addplot+[mark options={fill=cA, scale=0.6}, cA] table[x=epsilon, y=A]{\epsDetectability};
    \addplot+[mark options={fill=c3m, scale=0.6}, c3m] table[x=epsilon, y=R18_R50_A]{\epsDetectability};
    \end{axis}
    \end{tikzpicture}
    &
    \begin{tikzpicture}
    \begin{axis}[
    height=4cm,
    width=4.5cm,
    xmode=log,
    ylabel={PSNR},
    xlabel={$\epsilon$},
    xmin=0.00392, xmax=0.26,
    ymin=25, ymax=50,
    ytick={25,30,35,40,45,50},
    yticklabels={25,30,35,40,45,50},
    label style={font=\footnotesize},
    tick label style={font=\footnotesize},
    ylabel near ticks,
    xlabel near ticks,
    xtick={0.00392,0.00784,0.01569,0.03137,0.06275,0.12549,0.25098},
    xticklabels={1,2,4,8,16,32,64},
    ]
    \addplot+[mark options={fill=cR18, scale=0.6}, cR18] table[x=epsilon, y=R18_PSNR]{\epsQuality};
    \addplot+[mark options={fill=cR50, scale=0.6}, cR50] table[x=epsilon, y=R50_PSNR]{\epsQuality};
    \addplot+[mark options={fill=cA, scale=0.6}, cA] table[x=epsilon, y=A_PSNR]{\epsQuality};
    \addplot+[mark options={fill=c3m, scale=0.6}, c3m] table[x=epsilon, y=R18R50A_PSNR]{\epsQuality};
    \end{axis}
    \end{tikzpicture}
\end{tabular}
    \caption{Sensitivity analysis of {targeted} RP-FGSM when~$\epsilon$ varies. The misleading rate is evaluated on top-1 on the test dataset.
    When using three classifiers to craft the adversarial images, they are evaluated with the most accurate classifier, ResNet50 (seen classifier) and DenseNet161 (unseen classifier).
    KEY -- 
    {\protect\raisebox{2pt}{\protect\tikz \protect\draw[cR18,line width=1] (0,0) -- (0.3,0);}}:
    ResNet18;
    {\protect\raisebox{2pt}{\protect\tikz \protect\draw[cR50,line width=1] (0,0) -- (0.3,0);}}: ResNet50,
    {\protect\raisebox{2pt}{\protect\tikz \protect\draw[cA,line width=1] (0,0) -- (0.3,0);}}:~AlexNet;
    {\protect\raisebox{2pt}{\protect\tikz \protect\draw[c3m,line width=1] (0,0) -- (0.3,0);}}: {the} three classifiers;
    MS: misleading a seen classifier;
    MU: misleading an unseen classifier;
    MSwD: MS with defense;
    MUwD: MU with defense.}
    \label{fig:sensitivity}
\end{figure}
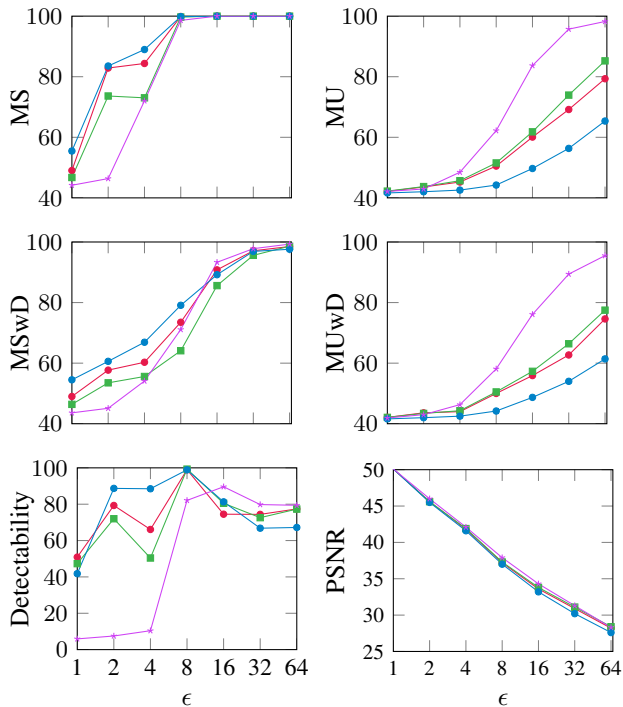
%====================================================================

%
%The above confirms that by combining a random selection of classifier and transformation generally improves the privacy properties of an adversarial attack.
Table~\ref{tab:ablationIterationNumber} reports how results vary using a random selection of the classifier with respect to an ensemble of classifiers, with varying numbers of iterations.
We report the performance measures with the proposed number of iterations (i.e.~40, with two classifiers; and 60, with three classifiers), which allows the proposed attack to perform the same number of forward and backward passes on the classifiers compared to an ensemble FGSM (e.g. E-FGSM). We also study the effect of the number of iterations on the performance measures by considering fewer iterations (i.e.~20).
For an equal number of uses of a classifier, the random selection of classifier improves the misleading of unseen classifiers by 16.4 percentage points in the top-5 (when using three classifiers), and decreases the detectability by 28.7 percentage points. Moreover, when the ensemble uses the classifiers three times more than RP-FGSM, RP-FGSM still outperforms the use of ensemble by an average of 15.0 percentage points in terms of misleading rate for classifiers with defense and by 28.9 percentage points in detectability (Table~\ref{tab:ablationIterationNumber}, last row).
This ablation study confirms that randomly selecting classifiers and transformations can improve the misleading rate, with and without defenses, when attacks are evaluated on the same number of forward/backward passes of a classifier.

%\subsection{Sensitivity analysis}
%\label{sec:eval:sensitivity}

%Now, we analyze the effect of varying the parameter~$\epsilon$ in the proposed RP-FGSM with respect to the proposed private properties. 
Figure~\ref{fig:sensitivity} reports the effect of varying the parameter~{$\epsilon \in \{1,2,4,8,16,32,64\}$} (larger values produce lower quality images~\cite{kurakin2016adversarial}).
Image quality is reported as PSNR, and we attack single classifiers, i.e. ResNet18, ResNet50 and AlexNet, and also their combination.
With seen classifiers, misleading rates above 95\% occur with $\epsilon$ larger than 8 (with defenses) and 32 (without defenses). 
With unseen classifiers, the proposed attack performs similarly with and without defenses, thus supporting the use of transformations to make the adversarial images robust to defenses. 
The proposed strategy of randomly choosing a classifier at each iteration, when multiple classifiers are considered, increases the misleading rate with unseen classifiers as shown in the first two plots on the right column. 
The detectability rate does not show a clear relation with the parameter $\epsilon$, as {it} {remains} between 40\% and 80\% for any $\epsilon$ when one classifier is attacked. When three classifiers are attacked the detection rate increases for $\epsilon \in \{1,2,4,8\}$ and it establishes at around 80\% for larger~$\epsilon$.
{I}mage quality monotonically decreases {when} $\epsilon$ increases.
{However, when multiple classifiers are used the PSNR does not significantly change. }

\subsection{Runtime analysis}
\label{sec:eval:time}

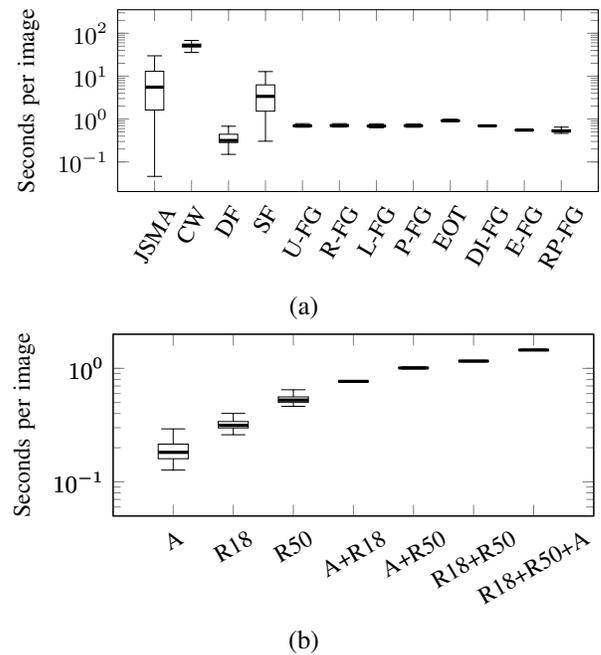
\begin{figure}[!t]
 \pgfplotsset{
    every non boxed x axis/.style={},
    boxplot/every box/.style={solid,ultra thin,black},
    boxplot/every whisker/.style={solid,ultra thin,black},
    boxplot/every median/.style={solid,very thick, black},
}
\pgfplotsset{ every non boxed x axis/.append style={x axis line style=-},
     every non boxed y axis/.append style={y axis line style=-}}
%\pgfplotsset{every tick label/.append style={font=\tiny}}
\pgfplotstableread{tikz/runtime/overall_time.txt}\data
\pgfplotstableread{tikz/runtime/overall_time_new.txt}\EFGSM
\setlength\tabcolsep{5pt}
\centering
\begin{tabular}{c}
\begin{tikzpicture}
 \begin{axis}
    [ymode=log,
    width=0.9\columnwidth,
    height=4cm,
    %axis x line=bottom,
    xmin=0, xmax=13,
    ymajorgrids=false,
    boxplot/draw direction=y,
    xtick={1,...,12},
    ytick={0.1,1,10,100},
    xticklabels={JSMA,CW,DF,SF,U-FG,R-FG,L-FG,P-FG,EOT,DI-FG,E-FG,RP-FG},
    ylabel={Seconds per image},
    label style={font=\small},
    tick label style={font=\small},
    xticklabel style={rotate=60}
    ]
\addplot+ [boxplot,mark=none, mark options={white,scale=0.5},boxplot/box extend=0.5] table[y expr=(\thisrow{JSMA_resnet50})]{\data};
\addplot+ [boxplot,mark=none, mark options={white,scale=0.5},boxplot/box extend=0.5] table[y expr=(\thisrow{CW_resnet50})]{\data};
\addplot+ [boxplot,mark=none, mark options={white,scale=0.5},boxplot/box extend=0.5] table[y expr=(\thisrow{DeepFool_resnet50})]{\data};
\addplot+ [boxplot,mark=none, mark options={white,scale=0.5},boxplot/box extend=0.5] table[y expr=(\thisrow{SparseFool_resnet50})]{\data};
\addplot+ [boxplot,mark=none, mark options={white,scale=0.5},boxplot/box extend=0.5] table[y expr=(\thisrow{N-FGSM_resnet50})]{\data};
\addplot+ [boxplot,mark=none, mark options={white,scale=0.5},boxplot/box extend=0.5] table[y expr=(\thisrow{R-FGSM_resnet50})]{\data};
\addplot+ [boxplot,mark=none, mark options={white,scale=0.5},boxplot/box extend=0.5] table[y expr=(\thisrow{L-FGSM_resnet50})]{\data};
\addplot+ [boxplot,mark=none, mark options={white,scale=0.5},boxplot/box extend=0.5] table[y expr=(\thisrow{P-FGSM_resnet50})]{\data};
\addplot+ [boxplot,mark=none, mark options={white,scale=0.5},boxplot/box extend=0.5] table[y expr=(\thisrow{EOT_resnet50})]{\EFGSM};
\addplot+ [boxplot,mark=none, mark options={white,scale=0.5},boxplot/box extend=0.5] table[y expr=(\thisrow{DI-FGSM_resnet50})]{\data};
\addplot+ [boxplot,mark=none, mark options={white,scale=0.5},boxplot/box extend=0.5] table[y expr=(\thisrow{E-FGSM_resnet50})]{\EFGSM};
\addplot+ [boxplot,mark=none, mark options={white,scale=0.5},boxplot/box extend=0.5] table[y expr=(\thisrow{pRMP-FGSM_resnet50})]{\data};
\end{axis}
\end{tikzpicture} \\
(a)\\
\pgfplotstableread{tikz/runtime/overall_time_new.txt}\data
\begin{tikzpicture}
\begin{axis}
    [ymode=log,
    width=0.9\columnwidth,
    height=4cm,
    only marks,
    %axis lines=left, 
    %xtick=\empty, ytick=\empty,
    %width=0.58\columnwidth,
    %axis x line=bottom,
    xmin=0, xmax=8,
    ymin=0.05, ymax=2,
    ymajorgrids=false,
    boxplot/draw direction=y,
    xtick={1,...,7},
    ytick style={draw=none},
    xtick style={draw=none},
    xticklabels={},
    ylabel={},
    label style={font=\small},
    tick label style={font=\small},
    ]
  \pgfplotsset{
    every non boxed x axis/.style={},
    boxplot/every box/.style={solid,ultra thin,black},
    boxplot/every whisker/.style={solid,ultra thin,black},
    boxplot/every median/.style={solid,very thick, black},
    boxplot width/.initial=0.2em,
}
\addplot+ [boxplot,mark=none, mark options={white,scale=0.5},boxplot/box extend=0.5] table[y expr=(\thisrow{pRMP-FGSM_alexnet})]{\data};
\addplot+ [boxplot,mark=none, mark options={white,scale=0.5},boxplot/box extend=0.5] table[y expr=(\thisrow{pRMP-FGSM_resnet18})]{\data};
\addplot+ [boxplot,mark=none, mark options={white,scale=0.5},boxplot/box extend=0.5] table[y expr=(\thisrow{pRMP-FGSM_resnet50})]{\data};
\addplot+ [boxplot,mark=none, mark options={white,scale=0.5},boxplot/box extend=0.5] table[y expr=(\thisrow{pRMP-FGSM_resnet18_alexnet})]{\data};
\addplot+ [boxplot,mark=none, mark options={white,scale=0.5},boxplot/box extend=0.5] table[y expr=(\thisrow{pRMP-FGSM_resnet50_alexnet})]{\data};
\addplot+ [boxplot,mark=none, mark options={white,scale=0.5},boxplot/box extend=0.5] table[y expr=(\thisrow{pRMP-FGSM_resnet18_resnet50})]{\data};
\addplot+ [boxplot,mark=none, mark options={white,scale=0.5},boxplot/box extend=0.5] table[y expr=(\thisrow{pRMP-FGSM_resnet18_resnet50_alexnet})]{\data};
\end{axis}
\begin{axis}
    [ymode=log,
    width=0.9\columnwidth,
    height=4cm,
    %axis x line=bottom,
    xmin=0, xmax=8,
    ymin=0.05, ymax=2,
    ymajorgrids=false,
    boxplot/draw direction=y,
    xtick={1,...,7},
    xticklabels={A, R18, R50, A+R18, A+R50, R18+R50, R18+R50+A},
    ylabel={Seconds per image},
    label style={font=\small},
    tick label style={font=\small},
    xticklabel style={rotate=30}
    ]
  \pgfplotsset{
    every non boxed x axis/.style={},
    boxplot/every box/.style={solid,ultra thin,black},
    boxplot/every whisker/.style={solid,ultra thin,black},
    boxplot/every median/.style={solid,very thick, blue},
}
\end{axis}
\end{tikzpicture}\\
(b)
\end{tabular}
\caption{Runtime analysis as average seconds per image. The test was performed on a random subset of 300 images. The horizontal line within the box shows the median; the bottom and top edges show the minimum and maximum values; and the lower and upper edges show the 25-percentile and 75-percentile, respectively. (a) Different adversarial attacks with ResNet50; and (b) proposed RP-FGSM when attacking a varying number of classifiers.
KEY --
    {
    JSMA: Jacobian-based Saliency Map Attack;
    CW: CarliniWagner;
    DF: DeepFool; 
    SF: SparseFool;
    }
    {U}-FG: Untargeted FGSM;
    R-FG: Random FGSM;
    L-FG: Least-Likely FGSM;
    P-FG: Private FGSM;
    EOT: Expectation Over Transformation; {DI-FG: Diverse Input FGSM;}
    E-FG: Ensemble FGSM;
    RP-FGSM: proposed attack;  
    A: AlexNet; R18: ResNet18; R50: ResNet50.
}
% E-FGSM~\cite{liu2018feature}~\protect\raisebox{2pt}{\protect\tikz \protect\draw[blue,line width=1] (0,0) -- (0.35,0);} and the proposed RP-FGSM~{\protect\raisebox{2pt}{\protect\tikz \protect\draw[red,line width=1] (0,0) -- (0.35,0);}}.
\label{fig:runtime}
\end{figure}

We compare the runtime of each attack in an Ubuntu 18.04.3 server equipped with an NVIDIA Tesla V100 GPU{, using} a random subset of 300 images. The implementations are in Python using the PyTorch library~\cite{pytorch}.
Figure~\ref{fig:runtime}(a) shows that JSMA, CW and SparseFool are the slowest attacks taking on average 52.14, 9.87 and 4.42~seconds per image, respectively, when attacking ResNet50.
DeepFool and FGSM-based attacks have a similar runtime performance with an average under 0.7~seconds per image.
Figure~\ref{fig:runtime}(b) shows how the runtime of RP-FGSM increases when the number of attacked classifiers increases (similarly to DI-FGSM and E-FGSM). The major factor affecting the runtime is the forward/backward {passes} on the classifier.

%====================================================================
%====================================================================

%%%%%%%%%%%%%%%%%%%
\section{Conclusion}\label{sec:concl}

We presented RP-FGSM, a Robust and Private Fast Gradient Sign Method that is designed to mislead seen and unseen classifiers with and without known defenses.  
RP-FGSM has better performance than other state-of-the-art adversarial attacks for privacy protection in the Private Places365 dataset, especially for unseen classifiers and when defenses are applied. The key for this performance is the random selection of a defense transformation and a classifier at each iteration{, which} prevent{s} the crafted perturbation {from} overfit{ting} {to a particular classifier or defense}.
As future work, we will extend the validation to tasks beyond scene classification.

\section*{Acknowledgment}
We thank the Alan Turing Institute (EP/N510129/1), which is funded by the EPSRC, for its support through the project PRIMULA.

%%%%%%%%%%%%%%%%%%%%%%%%
%\FloatBarrier
\bibliographystyle{IEEEtran}
\bibliography{main}

\newpage

\begin{IEEEbiography}[{\includegraphics[width=1in,height=1.25in,keepaspectratio]{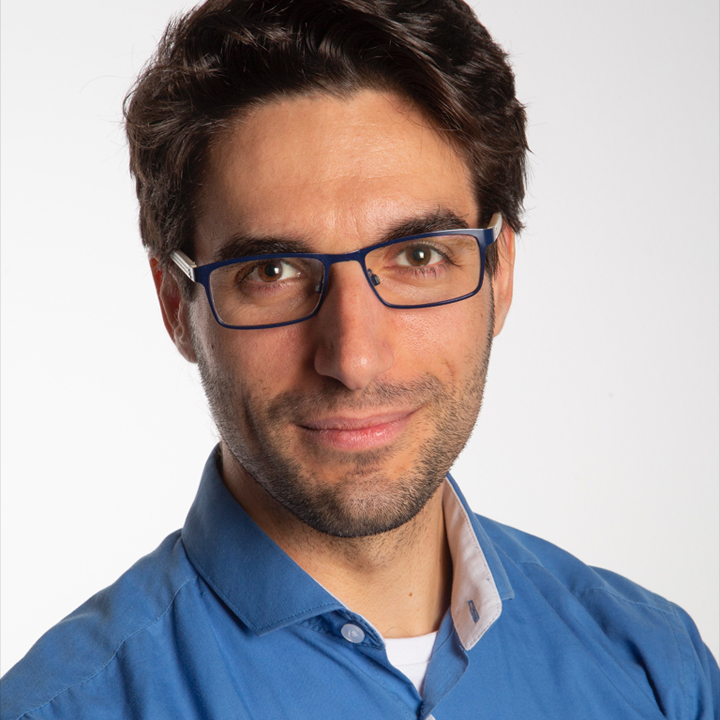}}]{Ricardo Sanchez-Matilla}
is a research assistant and a Ph.D. candidate with the Centre for Intelligent Sensing at Queen Mary University of London. He received his BSc and MSc degree in Telecommunication Engineering from the Universidad Autonoma de Madrid, Spain, in 2015. His research interests include computer vision, deep learning, and privacy.
\end{IEEEbiography}

\begin{IEEEbiography}[{\includegraphics[width=1in,height=1.25in,keepaspectratio]{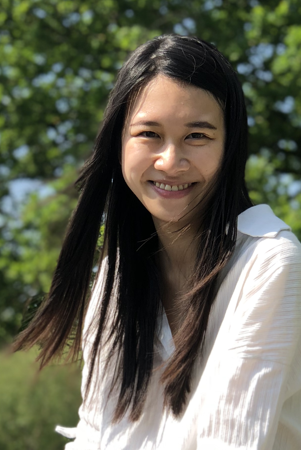}}]{Chau Yi Li} is a Ph.D. candidate in the Centre for Intelligent Sensing at Queen Mary University of London. After receiving her BSc in Mathematics and BEng in Information Engineering from The Chinese University of Hong Kong in 2015 and 2016, she received her MSc in Computer Science from Queen Mary University of London in 2017. Her research interests include underwater image processing, deep learning and privacy.
\end{IEEEbiography}

\begin{IEEEbiography}[{\includegraphics[width=1in,height=1.25in,keepaspectratio]{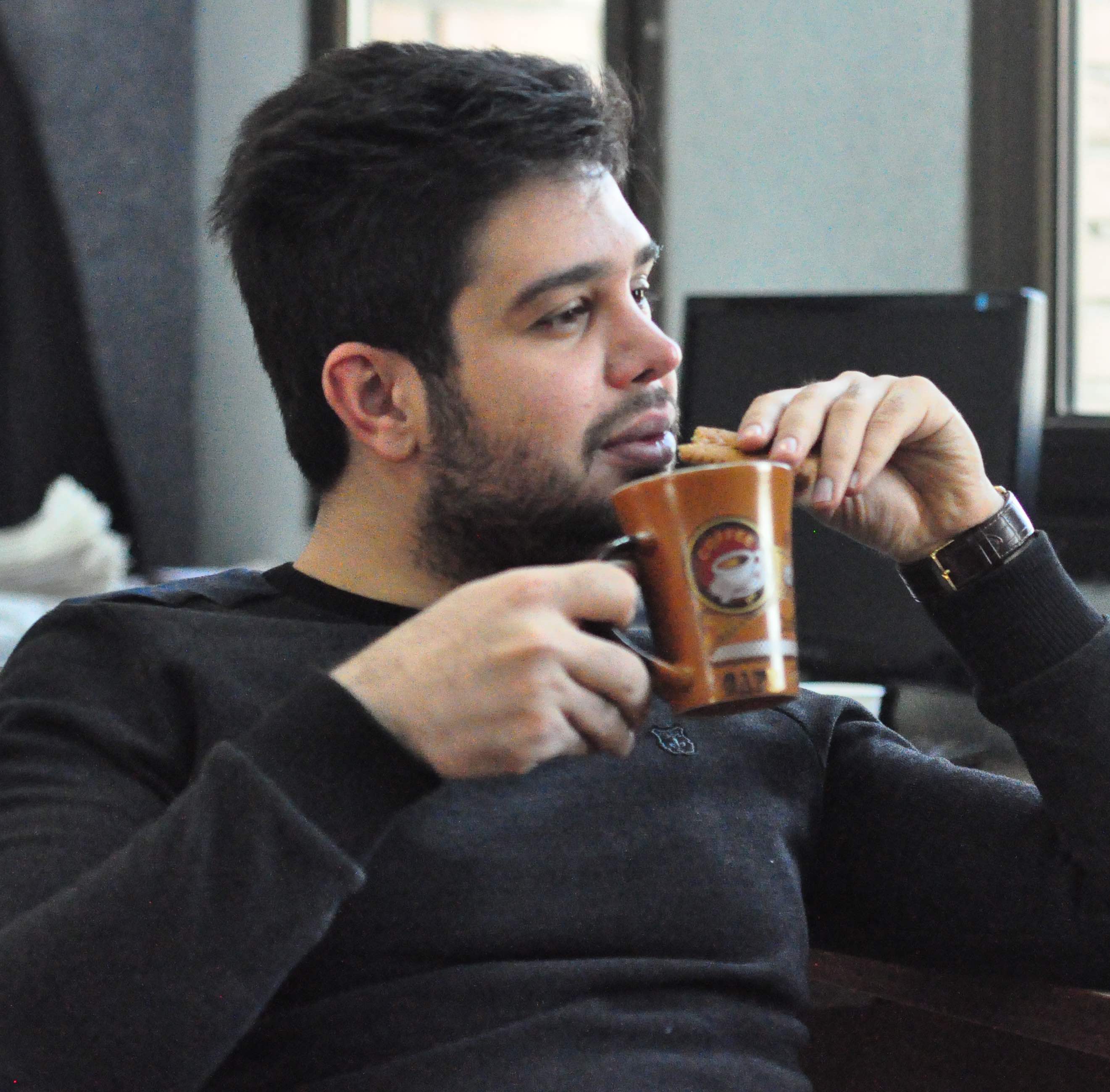}}]{Ali Shahin Shamsabadi}
is a PhD student in the Centre for Intelligent Sensing (CIS) in the school of Electronic Enginnering and Computer Science at Queen Mary University of London. His research interests are within the intersection of Machine Learning, Privacy and Image Processing. He aims to addresse the privacy risks in Machine Learning as a Service, which can be categorized in three research themes: privacy-preserving centralized learning, distributed learning and adversarial attacks for privacy protection.
\end{IEEEbiography}

%\begin{IEEEbiography}{Ali Shahin Shamsabadi}
%\end{IEEEbiography}

\begin{IEEEbiography}[{\includegraphics[width=1in,height=1.25in,keepaspectratio]{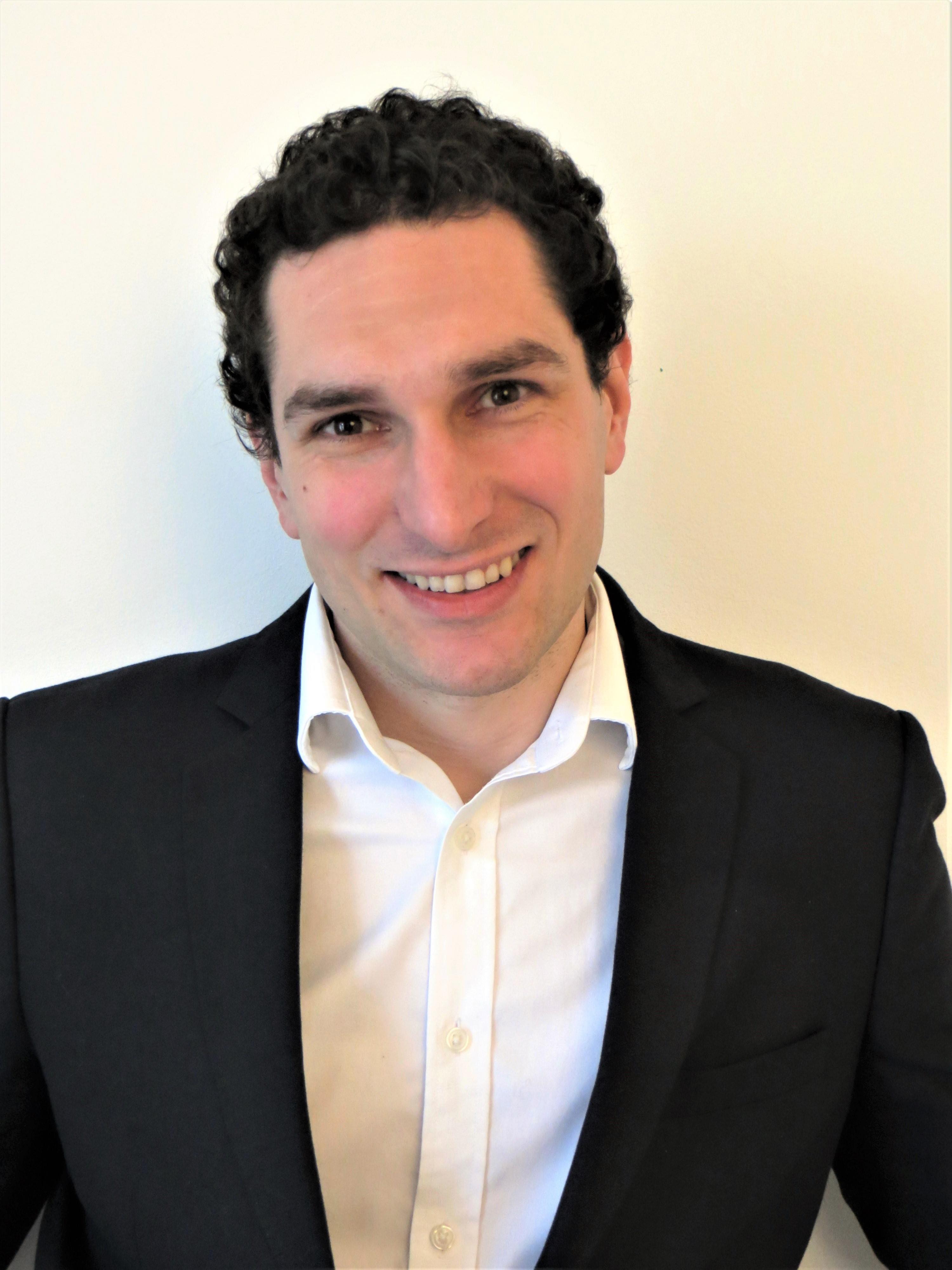}}]{Riccardo Mazzon}
received his BE in 2006 and MSc in 2009 in Computer Engineering from the University of Padova, Italy, and his PhD in Electronic Engineering from Queen Mary University of London (QMUL), UK, in 2013. Currently, he is a Research Manager at the Centre for Intelligent Sensing (CIS) and his research interests include video processing and analysis, camera networks and privacy.
\end{IEEEbiography}

\begin{IEEEbiography}[{\includegraphics[width=1in,height=1.25in,keepaspectratio]{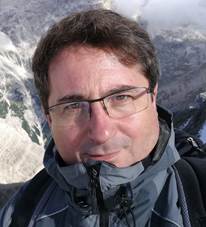}}]{Andrea Cavallaro} received the Ph.D. degree in electrical engineering from the Swiss Federal Institute of Technology, Lausanne, Switzerland, in 2002. He is Professor of Multimedia Signal Processing and the founding Director of the Centre for Intelligent Sensing at Queen Mary University of London (QMUL, UK), Turing Fellow at the Alan Turing Institute, the UK National Institute for Data Science and Artificial Intelligence, and Fellow of the International Association for Pattern Recognition.  He is Editor-in-Chief of Signal Processing: Image Communication; Senior Area Editor for the IEEE Transactions on Image Processing; Chair of the IEEE Image, Video, and Multidimensional Signal Processing Technical Committee; and an IEEE Signal Processing Society Distinguished Lecturer.

\end{IEEEbiography}

\end{document}